\documentclass{article}
\usepackage{kr}

\usepackage{microtype}

\usepackage{times}
\usepackage{soul}
\usepackage{url}
\usepackage[hidelinks]{hyperref}
\usepackage[utf8]{inputenc}
\usepackage[small]{caption}
\usepackage{graphicx}
\usepackage{amsmath}
\usepackage{mathrsfs}
\usepackage{amsthm}
\usepackage{booktabs}
\usepackage{algorithm}
\usepackage{algpseudocode}
\usepackage{scalerel}
\usepackage{enumerate}
\usepackage{amssymb}

\usepackage{pifont}

\usepackage{enumitem}

\usepackage{xspace}
\usepackage{misc}

\usepackage{stmaryrd}

\usepackage{thm-restate}

\usepackage{tikz}
\usetikzlibrary{positioning}
\usetikzlibrary{automata,arrows.meta}

\newtheorem{corollary}{Corollary}
\newtheorem{theorem}{Theorem}
\newtheorem{lemma}{Lemma}

\newtheorem{exmp}{Example}

\newcommand{\ELandELI}{\ensuremath{{\cal E\!\!\: L\mkern-1mu (\mkern-1mu I)}}\xspace}

\newcommand{\TGD}{\ensuremath{\text{TGD}}\xspace}
\newcommand{\GTGD}{\ensuremath{\text{GTGD}}\xspace}
\newcommand{\FGTGD}{\ensuremath{\text{FGTGD}}\xspace}
\newcommand{\FONETGD}{\ensuremath{\text{F1TGD}}\xspace}
\newcommand{\FTGD}{\ensuremath{\text{FullTGD}}\xspace}
\newcommand{\IND}{\ensuremath{\text{IND}}\xspace}

\newcommand{\Iel}{\ensuremath{I}\xspace}
\newcommand{\Jel}{\ensuremath{J}\xspace}
\newcommand{\Nel}{\ensuremath{N}\xspace}
\newcommand{\Pel}{\ensuremath{P}\xspace}
\newcommand{\interpretation}{instance\xspace}
\newcommand{\interpretations}{instances\xspace}

\newcommand{\lead}[1]{\medskip\noindent\textbf{#1}}

\title{Fitting Ontologies and Constraints to Relational Structures}

\author{
Simon Hosemann\textsuperscript{\rm 1}\!\and
Jean Christoph Jung\textsuperscript{\rm 2}\!\and 
Carsten Lutz\textsuperscript{\rm 1,4}\!\and
Sebastian Rudolph\textsuperscript{\rm 3,4} \\
\affiliations
\textsuperscript{\rm 1}Leipzig University
$\qquad$
\textsuperscript{\rm 2}TU Dortmund University
$\qquad$
\textsuperscript{\rm 3}TU Dresden \\
\textsuperscript{\rm 4}Center for Scalable Data Analytics and Artificial Intelligence (ScaDS.AI)
\emails 
simon.hosemann@uni-leipzig.de, 
jean.jung@tu-dortmund.de, \\[1mm]
carsten.lutz@uni-leipzig.de, 
sebastian.rudolph@tu-dresden.de 
}
\begin{document}

\maketitle

\begin{abstract}
    We study the problem of fitting ontologies and constraints to positive and negative examples that take the form of a finite relational structure. As ontology and constraint languages, we consider the description logics \EL and \ELI as well as several classes of tuple-generating dependencies (TGDs): full, guarded, frontier-guarded, frontier-one, and unrestricted TGDs as well as inclusion dependencies. We pinpoint the exact computational complexity, design algorithms, and analyze the size of fitting ontologies and TGDs. We also investigate the related problem of constructing a finite basis of concept inclusions / TGDs for a given set of finite structures. While finite bases exist for \EL, \ELI, guarded TGDs, and inclusion dependencies, they in general do not exist for
    full, frontier-guarded and frontier-one TGDs.
\end{abstract}

\section{Introduction}

In a \emph{fitting problem} as studied in this article, one is given as input a finite set of positive and negative examples, each taking the form of a logical structure, and the 
task is to produce a logical formula that satisfies all  examples. 
Problems of this form play a fundamental role in several applications. A prime
example is the query by example paradigm in the field of data management, also known as query reverse engineering~\cite{pvldb/LiCM15,icdt/Barcelo017,pods/CateDFL23}. 
In that case, the positive and negative examples are database instances and 
the formula to be constructed is a database query. In concept learning in description logics (DLs)~\cite{ml/LehmannH10,kr/JungLPW21,ijcai/CateFJL23}, the examples are ABoxes and the formula is a DL concept; note that fitting problems
are  connected to PAC learning by the fundamental theorem of computational
learning theory. Another example  is entity comparison~\cite{semweb/PetrovaSGH17,semweb/PetrovaKGH19} where the examples
are knowledge graphs and the formula is a SPARQL query.

In this article, we study fitting problems that aim to support the
construction of two kinds of artefacts: (i)~ontologies formulated in a description logic
or an existential rule language and (ii)~constraints on databases that 
take the form of tuple-generating dependencies (TGDs). Since `existential
rule' and `TGD' are two names for  the same thing, from now on
we shall only  speak of TGDs. We will consider both unrestricted TGDs
and restricted classes of TGDs: full, guarded, frontier-guarded, and frontier-one, as well as inclusion dependencies, which also form a restricted class of TGDs. In the DLs \EL and \ELI that we consider in this article, an ontology is a set of
concept inclusions that can be  translated into an equivalent TGD. For uniformity, we shall thus also refer to concept inclusions as TGDs. 

Let us be more precise about the fitting problems that we study. Examples
are finite relational structures that, as in data management, we refer to as instances. In the 
DL case, such structures only admit unary and binary relations and are 
usually called interpretations. The formulas that we seek to construct fall into two classes. We may either be
interested in a single TGD, to be used as a building block in an ontology or as a database constraint, or we may want to construct a finite set of TGDs
to be used as an ontology or as a collection of constraints. From our perspective, there is in fact no difference between an ontology and a set of constraints: any set of TGDs
can be used as an ontology when an open world semantics is adopted and
as a set of constraints under a closed world semantics. We are interested both in the construction
of fitting TGDs and ontologies, and
in deciding their existence.

The problem of fitting an ontology to a given set of examples
turns out to be closely related to a problem that has been studied
in the area of description logic and is known as finite basis
construction~\cite{phd/Distel2011,jair/GuimaraesOPS23,Kriegel_2024}. One fixes an ontology language \Lmc
and is given as input a finite instance $I$ and the task is to produce an \Lmc-ontology \Omc such that $I \models \rho$ if and only if $\Omc \models \rho$, for all \Lmc-TGDs $\rho$. We
generalize this problem to a finite set \Hsf of input instances.
It turns out that a finite basis \Omc for the set of positive examples
is a canonical fitting ontology in the sense that if any \Lmc-ontology
fits the given (positive and negative) examples, then \Omc does. 
Thus, constructing finite bases provides
an approach to constructing fitting
ontologies and to deciding their
existence. 
This approach in fact often yields decidability
and tight upper complexity bounds (`tight' meaning that we prove matching lower and upper bounds).

We first consider the DLs  \EL and \ELI as well as their extensions with
the $\bot$ concept. We reprove the existence of finite bases for \EL, 
already known from~\cite{icfca/BaaderD08,phd/Distel2011}, and
simultaneously prove that finite bases exist also for \ELI, which, to the
best of our knowledge, is a new result. In contrast to the proofs from~\cite{icfca/BaaderD08,phd/Distel2011}, our proofs are direct
in that they do not rely on the machinery of formal concept analysis. The constructed
bases are of double exponential size, but can be succinctly represented in
single exponential size by  structure sharing. We also show that these
size bounds are tight, both for \EL and for \ELI. We obtain from this an \ExpTime upper bound for the fitting
existence problem for \ELandELI-ontologies. 

We then provide a semantic characterization of fitting  \ELandELI-TGD 
existence in terms of simulations and direct products. This characterization gives
rise to an algorithm for fitting  \ELandELI-TGD existence  and opens up an alternative path to algorithms for  fitting  \ELandELI-ontology existence. It also 
enables us to prove
lower complexity bounds and we in fact show that all four problems are \ExpTime-complete. We also prove tight bounds on the size of fitting TGDs and fitting
ontologies, which are identical to the size bounds on finite bases described
above.

We next turn to TGDs. For guarded TGDs ({\GTGD}s), we implement exactly
the same program described above for \ELandELI, but obtain different
complexities. We show that finite GTGD-bases always exist and establish a tight single
exponential bound on their size.
Succinct representation does
not help to reduce the size.
We give a characterization of fitting GTGD existence and fitting GTGD-ontology existence in terms of
products and homomorphisms, show that fitting GTGD existence and fitting GTGD-ontology existence are \coNExpTime-complete, and give
a tight single exponential bound on
the size of fitting GTGDs and GTGD-ontologies. The \coNExpTime
upper bound may be obtained either
via finite bases or via the semantic characterization.

For all other considered
classes of TGDs, the approach via finite bases fails: for the frontier-guarded, frontier-one, and full cases, we prove that finite bases need not exist. For inclusion dependencies, finite bases trivially exist but approaching fitting via this route does not result in an optimal upper complexity bound. For unrestricted TGDs, the existence of finite bases is left open. 

We may, however, still approach fitting existence in a direct way or via a semantic characterization. For inclusion
dependencies ({\IND}s), we use direct arguments to
show that fitting \IND existence and
fitting \IND-ontology existence are \NPclass-complete, and that the size of fitting \IND-ontologies is polynomial.
For all other remaining cases, we 
establish semantic characterizations
and then use them to approach fitting
existence.
In this way, we prove the following. Fitting
ontology existence and fitting TGD existence are \coNExpTime-complete for TGDs that are 
 frontier-guarded or frontier-one.
 For full TGDs, fitting TGD existence
 is \coNExpTime-complete and fitting
 ontology existence is in $\Sigma^p_2$ 
 and DP-hard. In the case of unrestricted TGDs, both problems are  \coNExpTime-hard
 and we prove a \coTwoNExpTime upper bound for fitting
 ontology existence and a  \coThreeNExpTime upper bound for fitting TGD existence. We also show tight single exponential size bounds for fitting TGDs and ontologies in the case of 
  frontier-guarded and frontier-one TGDs. We do the same for fitting full TGDs, whereas if a fitting ontology of full TGDs exists, there always exists one of polynomial size.  For unrestricted
  TGDs 
  we give a single exponential lower bound and a triple (for TGDs) and double (for ontologies) exponential upper bound on the size of fittings. 

 \smallskip
Proofs are provided in the \hyperref[appendix]{Appendix}.

\paragraph{Related Work.} There are several related lines of work. One is~\cite{CateD15,sigmod/BogdanCKT11}
where the fitting problem is studied for schema mappings that take the form of a set of TGDs. While
the setup in~\cite{sigmod/BogdanCKT11} uses universal examples
and diverges from ours, the fitting problems considered in~\cite{CateD15}
are closely related to the ones studied here. For schema mappings, separate
source and target instances (in possibly different schemas) are used in every example while we only have a single instance per example. Upper bounds carry over from the schema mapping
setting to our setting because we can choose 
the source and target instance to be identical, while there seems no easy way
to carry over lower bounds. We remark that
~\cite{CateD15} consider 
unrestricted TGDs and
so-called GAV constraints, which correspond to single-head full TGDs. They assume a fixed schema (which we do not) and only study sets of TGDs/schema mappings (corresponding to
 ontologies), but not single TGDs. 

Also related is inductive logic programming (ILP), especially in its `learning from interpretations' incarnation. There, one question of interest is  fitting
existence (often called the `consistency problem') for definite first-order clauses, 
which correspond to single-head full TGDs~\cite{alt/MuggletonF90,sigart/KietzD94,aaai/Cohen94a,ilp/GottlobLS97}. However, the
ILP literature typically admits background knowledge as an additional input to the fitting problem and, moreover, adopts additional biases such as 
a constant number of variables,
determinacy conditions that pertain to the functionality of relations, or variable depth. We are not aware that results have been obtained for the unrestricted case studied in this article. 
 
There are also 
more loosely related lines of work concerning
fitting problems for conjunctive queries~\cite{CateD15,pods/CateDFL23,sigmod/CateFJL23} and for description logic concepts~\cite{ijcai/FunkJLPW19,kr/JungLPW20,ai/JungLPW22,ijcai/CateFJL23}. 
We also mention~\cite{DeRaedt_Passerini_Teso_2018}, which is  concerned with fitting propositional logic constraints to data.
Our work on finite bases is related to~\cite{phd/Rudolph2006,icfca/BaaderD08,phd/Distel2011,jair/GuimaraesOPS23,Kriegel_2024}. All those works
employ formal concept analysis to construct finite bases while
we use a direct approach.

\section{Preliminaries}
\label{sect:prelim}

\lead{Schema, (Pointed) Instance, Homomorphism.} A \emph{schema} \Smc is a
non-empty finite set of \emph{relation symbols} $R$, each
with an associated \emph{arity} $\mn{ar}(R) \geq 1$. An
\Smc-\emph{fact}  is an expression $R(a_1, \ldots, a_m)$, where $a_1,
\ldots, a_m$ are \emph{values}, $R\in\Smc$, and $\mn{ar}(R)=m$. An
\emph{\Smc-instance} is a (possibly infinite) 
set $I$ of \Smc-facts. The
\emph{active domain} of $I$, denoted $\mn{adom}(I)$, is the set of all values
that occur in facts of $I$.
A \emph{pointed \Smc-instance}  is a pair $(I,\bar{a})$ where $I$ is
an \Smc-instance and $\bar{a}$ is a tuple of values.
We refer to $\bar{a}$ as the
\emph{distinguished values} of  $(I,\bar{a})$ and define
its \emph{arity} to be $|\bar{a}|$.
The distinguished values need not  
belong to $\mn{adom}(I)$, though most of the time they do.

Given two instances $I,J$ over the same schema, a \emph{homomorphism} is a map
$h:\mn{adom}(I) \rightarrow\mn{adom}(J)$ such that $R(\bar a) \in I$ implies $R(h(\bar a)) \in J$ where $h(\bar a)$ means the component-wise application of $h$ to $\bar a$.
For pointed instances $(I,\bar a)$ and $(J,\bar b)$, we additionally
demand that $h(\bar a)=\bar b$.
To indicate that there
exists a homomorphism from $I$ to $J$,
we write $I \to J$ and likewise for pointed instances.

\lead{Conjunctive Query.} A \emph{conjunctive
  query (CQ)} over a schema \Smc is a formula $q(\bar{x})$ of the form
$
 \exists \bar{y} \, \varphi(\bar{x},\bar{y})
$
where $\bar{x}$ and $\bar{y}$ are tuples of variables and $\varphi$ is a
conjunction of relational atoms that uses only variables that occur in $\bar{x}$ or
$\bar{y}$ and
only relation symbols from \Smc. Every variable from $\bar{x}$ must occur in one
of the atoms in $\varphi$, a condition known as the \emph{safety} of the CQ. The \emph{arity} of $q$ is the number of variables in~$\bar{x}$, also called \emph{answer variables}. We use $\mn{var}(q)$ to denote the
set of all variables that occur in $q$. We say that  $q$ is \emph{guarded} if
$\varphi$ contains an atom that mentions all variables from $\bar{x}$ and
$\bar{y}$. 

There is a natural correspondence between CQs over a schema \Smc and
finite pointed \Smc-instances of the same arity.
First, the \emph{canonical
  instance} of a CQ $q(\bar{x})$ is
$(I_q,\bar{x})$, where $\mn{adom}(I_q)=\mn{var}(q)$  and the facts of $I_q$ are the atoms in~$q$.
Conversely, if $(I,\bar{a})$ satisfies $\bar{a}
\subseteq \mn{adom}(I) \neq \emptyset$ and no value appears more than once in $\bar a$, then the \emph{canonical CQ} of $(I,\bar{a})$ is
$q_{(I,\bar{a})}(\bar a)$ where $\mn{var}(q_{(I,\bar{a})})=\mn{adom}(I)$ and the atoms of $q_{(I,\bar{a})}$ are the facts of $I$.

With a homomorphism from $q(\bar x)$ to $(I,\bar a)$, we mean a homomorphism
from $(I_q,\bar x)$ to $(I,\bar a)$. For a CQ $q(\bar x)$ over some schema
\Smc and an \Smc-instance $I$, we use $q(I)$ to denote the set of all
\emph{answers} $\bar a$ to $q$ on $I$, that is, all tuples $\bar a \subseteq \mn{adom}(I)$ such that there is a homomorphism from $q$ to $(I,\bar a)$.

\lead{TGDs, Ontologies.}
A \emph{tuple-generating dependency} (TGD) over a schema \Smc is a formula of the form
$$ \rho = \forall \bar{x}\forall \bar{y} \big(\varphi (\bar{x}, \bar{y})
\rightarrow \exists\bar{z} \, \psi(\bar{x}, \bar{z})\big) $$ where
$\varphi(\bar{x}, \bar{y})$ and $\exists \bar{z} \, \psi(\bar{x}, \bar{z})$ are
conjunctions of relational atoms over \Smc called the \emph{body} and \emph{head} of
$\rho$. The variables~$\bar x$, which  occur both in the body and in the head, are called \emph{frontier variables}.
A TGD is \emph{full} or a \FTGD if the head contains no existentially quantified variables,
\emph{guarded} or a \GTGD if the body contains an atom that mentions all variables that occur in the body, 
\emph{frontier-guarded} or an \FGTGD if the body contains an atom that mentions all frontier variables, and
\emph{frontier-one} or an \FONETGD if it has at most one frontier variable. A TGD is an \emph{inclusion dependency} or an \IND if both the body and head contain only one atom. The repeated use of variables is admitted.
When writing TGDs, the universal quantifiers are
usually omitted.

An instance $I$ satisfies a TGD $\rho = \varphi (\bar{x}, \bar{y}) \rightarrow
\exists \bar{z} \, \psi(\bar{x}, \bar{z})$, denoted  $I\models\rho$, if $\exists
\bar{y} \, \varphi(\bar{x}, \bar{y})(I)\subseteq \exists \bar{z} \, \psi(\bar{x},
\bar{z})(I)$, where we view the body and head each as a CQ.  An \emph{ontology}
is a finite set of TGDs. We speak of a \GTGD-ontology if all TGDs are guarded, and likewise for \FGTGD and \FONETGD.  An
instance $I$ is a \emph{model} of an ontology \Omc if it satisfies all
TGDs in~\Omc. For an ontology \Omc and a
TGD $\rho$, we write $\Omc \models \rho$
if every model of \Omc satisfies $\rho$. For ontologies $\Omc,\Omc'$, we write
$\Omc \models \Omc'$ if $\Omc \models \rho$ for every $\rho \in \Omc'$.

\lead{Description Logics \EL, \ELI, $\EL_\bot$, $\ELI_\bot$.}
In the context of description logics, one uses schemas \Smc that contain only unary and
binary relation symbols, also referred to as \emph{concept names} and \emph{role names}. We let $A$ range over concept names of \Smc and $R$
over the role names of \Smc. The set of $\ELI_\bot$-concepts over
\Smc is given by the grammar
$$
C ::= \top \mid \bot \mid A \mid C\sqcap C \mid \exists R. C \mid \exists R^-. C.
$$
$\EL_\bot$-concepts are obtained by the same grammar without \emph{inverse roles} $R^-$, and \ELI- and \EL-concepts
are defined likewise, but do not permit the use of `$\bot$'.

The semantics of DLs is  given in terms of interpretations.
Every non-empty instance $I$  may be viewed as an
interpretation $\Imc = (\Delta^{\Imc}, \cdot^{\Imc})$ with
$\Delta^{\Imc} = \mn{adom}(I)$, $A^{\Imc} = \{d \mid
A(d)\in I\}$ for all concept names $A$, and $R^\Imc = \{(d,e)\mid
R(d,e)\in I\}$ 
for all role names $R$.  
The extension $C^\Imc$ of an $\ELI_\bot$-concept $C$ is then defined as usual~\cite{daglib/0041477}. For easier reference we set $C^I = C^\Imc$ and, for the empty instance
$J$,  $C^J =
\emptyset$ for all $\ELI_\bot$-concepts~$C$. We also write $\Delta^I$ in place
of $\mn{adom}(I)$.

Let $\Lmc \in \{ \EL, \ELI,\EL_\bot,\ELI_\bot \}$.
An \Lmc-\emph{concept inclusion} over \Smc takes the form  $C\sqsubseteq D$ where $C$ and $D$ are \Lmc-concepts over \Smc. An \emph{\Lmc-ontology} is a finite set of \Lmc-concept inclusions. An
instance $I$ satisfies a concept inclusion $C\sqsubseteq D$, written $I\models C\sqsubseteq
D$, if $C^I\subseteq D^I$. 
 An
instance $I$ is a \emph{model} of an ontology \Omc if it satisfies all
concept inclusions in \Omc.
Every \ELI-concept inclusion can be expressed as an \FONETGD in a straightforward way. From now on we may thus view \ELI-ontologies as \FONETGD-ontologies and concept inclusions as TGDs.\footnote{For uniformity, we will also take the freedom to refer to an $\ELI_\bot$-concept inclusion as a TGD, despite the fact that TGDs do not admit $\bot$ as the rule head.} 
We also remark
that by
a straightforward normalization which removes syntactic nesting in rule
bodies at the cost of introducing additional unary predicates, every \ELI-ontology \Omc can be converted into a \GTGD-ontology that is a conservative extension of \Omc. In our context, however, the change of schema matters
and results for \GTGD-ontologies do not automatically transfer to~\ELI. 

For any syntactic object $O$ such as a conjunctive query, a TGD, or an ontology, we use $||O||$ to denote the \emph{size} of~$O$, meaning the
length of $O$ when encoded as
a word over a suitable finite alphabet. We next explain what we
mean by the succinct representation of $\ELI_\bot$-concepts and ontologies. An $\ELI_\bot$-concept $C$ is a \emph{subconcept} of an  $\ELI_\bot$-concept $D$ (ontology \Omc) if $C$ occurs in $D$ (in \Omc) as a syntactic subexpression. An ontology \Omc (or concept $C$) can be represented succinctly
 by using structure sharing, that is, representing every subconcept $C$ of \Omc only once and using pointers to share that representation among different occurrences of $C$ in \Omc. This
representation can sometimes
make the size of the representation exponentially smaller. It is easy to see that computationally,
succinct representation of concepts and ontologies is almost always harmless. In particular, the 
following holds.
\begin{lemma}
  \label{lem:succinctPTime}
  Given a finite instance $I$ and an $\ELI_\bot$-ontology \Omc in succinct representation, it can be
  decided in polynomial time whether $I \models \Omc$.
\end{lemma}

\lead{Simulation.}
The expressive power of \EL and \ELI is closely linked to simulations. With an \emph{\EL-role} or  \emph{$\EL_\bot$-role}, we mean a role name. An \emph{\ELI-role} or  \emph{$\ELI_\bot$-role} is a role name or an inverse role. Let
$\Lmc \in \{ \EL,\ELI\}$. For instances
$I,J$, we call a relation $Z\subseteq \Delta^I \times \Delta^J$  an \Lmc-\emph{simulation} from $I$ to $J$ if it satisfies the following conditions:
\begin{enumerate}
  \item If $(d,d') \in Z$, then $d\in A^I$ implies $d'\in A^J$.
  \item If $(d,d')\in Z$ and $(d,e)\in R^I$ with $R$ an \Lmc-role, then there is a  $(d',e')\in R^J$ with $(e,e') \in Z$.
\end{enumerate}
For unary
pointed instances $(I,d),(J,e)$ we write $(I,d) \preceq_\Lmc (J,e)$ to denote
the existence of an \Lmc-simulation $Z$ from $I$ to $J$ with $(d,e)\in
Z$.
As a special case, we also write $(I,d) \preceq_\Lmc (J,e)$
if $d \notin \Delta^I$.

With the \emph{role depth}
of an $\ELI_\bot$-concept $C$, we mean its quantifier depth, defined in
the standard way. The \emph{outdegree} of $C$ is the maximum
number of existential restrictions in any conjunction in $C$. The following lemma is based on the classic elimination procedure for computing the maximal \Lmc-simulation between two interpretations. Since other
proofs refer back to it, we recall the proof details in
the appendix.
\begin{restatable}{lemma}{lemsimtothemax}
\label{lem:simtothemax}
  Let $\Lmc \in \{ \EL, \ELI \}$. There is a polynomial time algorithm that, given finite pointed \interpretations $(I,d)$ and $(J,e)$, decides whether $(I,d) \preceq_\Lmc (J,e)$ and, if this is not the case, outputs the succinct representation of an \Lmc-concept $C$ of role depth at most $|\Delta^I| \cdot |\Delta^J|$ and outdegree at most $|\Delta^J|$ such that $d \in C^I$ and $e \notin C^J$.
\end{restatable}

\lead{Disjoint Union, Direct Product.}
Let \Hsf be a non-empty finite set of instances with pairwise disjoint active
domains. Then the \emph{disjoint union} of the instances in \Hsf, denoted 
${\biguplus}\Hsf$, is the instance $\bigcup\Hsf$. When the domains of the 
instances in \Hsf are not pairwise disjoint, we assume that renaming is used
to achieve disjointness before forming ${\biguplus}\Hsf$.
The \emph{direct product} of two pointed instances
$(I,\bar{a})$ and $(J,\bar{b})$, with $\bar{a}=\langle a_1, \ldots a_k\rangle$
and $\bar{b}=\langle b_1, \ldots, b_k\rangle$ is the  pointed instance
$(I\times J, \bar a \times \bar b)$ where $I\times J$
consists of all facts $R((c_1,d_1), \ldots, (c_n,d_n))$ such that $R(c_1,
\ldots, c_n)$ is a fact of $I$ and $R(d_1, \ldots, d_n)$ is a fact of $J$,
and $\bar a \times \bar b = \langle (a_1,b_1), \ldots, (a_k,b_k)\rangle$.
Note that the elements of $\bar a \times \bar b$ are not necessarily in
$\mn{adom}(I\times J)$, and in fact this is precisely why we do not require all values of $\bar{a}$ to lie in $\mn{adom}(I)$ for a pointed instance $(I,\bar{a})$.
The product construction extends naturally to non-empty finite sets $S$ of pointed
instances.
We then use $\prod S$ to denote its product.

\lead{Fitting Problems.} Let \Lmc be one of the ontology 
languages introduced above, such as \ELI and \GTGD. A \emph{fitting instance} is a pair  $(\Psf, \Nsf)$ 
with \Psf and \Nsf finite and non-empty sets of finite instances. Let \Smc be the set of symbols used in \Psf and \Nsf.
An \Lmc-ontology \Omc over \Smc
\emph{fits} $(\Psf, \Nsf)$ if $I\models \Omc$ for all $I\in\Psf$ and
$J\not\models \Omc$ for all $J\in \Nsf$. 
For a single \Lmc-TGD $\rho$ over \Smc, fitting $(\Psf, \Nsf)$ is defined in exactly the same way.  Note that an
ontology~\Omc fitting~$(\Psf, \Nsf)$ does not imply
that every TGD $\rho \in \Omc$ fits~$(\Psf, \Nsf)$. 

We  consider the following computational problems:

\begin{center}
  \fbox{ \parbox{0.90\linewidth}{
      \begin{tabular}{lp{0.70\linewidth}@{}}
        \textsc{Problem}:  & Fitting \Lmc-ontology existence                    \\
        \textsc{Input}:    & Fitting instance  $(\Psf,\Nsf)$.          \\
        \textsc{Question:} & Does $(\Psf, \Nsf)$ have a fitting \Lmc-ontology?
      \end{tabular}\hspace*{-2pt}
    } }
\end{center}

\begin{center}
  \fbox{ \parbox{0.90\linewidth}{
      \begin{tabular}{lp{0.70\linewidth}@{}}
        \textsc{Problem}: & Fitting \Lmc-ontology construction         \\
        \textsc{Input}:   & Fitting instance  $(\Psf,\Nsf)$.   \\
        \textsc{Output:}  & \Lmc-ontology that fits $(\Psf, \Nsf)$ if
        existent, ``no fitting'' otherwise.
      \end{tabular}
    } }
\end{center}
Analogous problems for fitting \Lmc-TGDs in place of \Lmc-ontologies are defined
in the expected way.

\begin{exmp}
 Consider the instances
 $$
 \begin{array}{r@{\;}c@{\;}l}
   P&=& \{ R(a,b), R(b,a) \} \\[1mm]
   N &=& \{ R(a,b), R(b,c), R(c,a) \}.
 \end{array}
 $$
 Then $(\{P\},\{N\})$ has no fitting \ELI-TGD or -ontology, but it has fitting {\GTGD}s such as $R(x,y) \rightarrow R(y,x)$. Now let
 $$
 N'= N \cup \{ R(b,a), R(c,b), R(a,c) \}.
 $$
 Then $(\{P\},\{N'\})$  has no fitting \GTGD or \GTGD-ontology.
 But it has fitting {\FONETGD}s such as 
 $$
   R(x,y) \wedge R(y,z) \wedge R(z,x) \rightarrow R(x,x).
 $$
  All negative claims in this example are a consequence of the
  semantic characterizations established below.
\end{exmp}

How are fitting ontologies and fitting TGDs related?
The following is an immediate consequence
of the definition of fitting and the semantics of ontologies and TGDs.
\begin{lemma}
  \label{lem:oneneg}
  Let $(\Psf,\Nsf)$ be a fitting instance. Then there is an \Lmc-ontology
  that fits $(\Psf,\Nsf)$ if and only if  for every $N \in \Nsf$, there is an \Lmc-TGD that fits $(\Psf,\{N\})$.
\end{lemma}
This clearly implies the following.
\begin{lemma}
  Let $(\Psf,\Nsf)$ be a fitting instance. If there is an \Lmc-ontology
  that fits $(\Psf,\Nsf)$ then there is an \Lmc-ontology \Omc that  fits $(\Psf,\Nsf)$ and contains at
  most $|\Nsf|$ TGDs.
\end{lemma}

We also make an observation regarding full TGDs and the number of head
atoms. If there is a full TGD $\rho$ that fits an instance $(\Psf,\{N\})$ 
with a single negative example, then there is a fitting \FTGD with a single head atom. In fact, 
there must be a head atom falsified in the single negative
example, and we may drop
all
other head atoms. This is not the case, however, if more than one
negative example is present. 
\begin{exmp}
  Let $\Psf=\{P\}$ with $P=\{A(a),B_1(a),B_2(a)\}$ and $\Nsf = \{ N_1,N_2 \}$
  with $N_i=\{ A(a), B_i(a) \}$ for $i \in \{1,2\}$. Then the full TGD 
  $A(x) \rightarrow B_1(x) \wedge B_2(x)$ fits $(\Psf,\Nsf)$, but there
  is no fitting \FTGD with a single head atom.
\end{exmp}
By Lemma~\ref{lem:oneneg}, this implies
that if there is a \FTGD-ontology that fits an instance $(\Psf,\Nsf)$,
with \Nsf containing any number of examples, then there is a \FTGD-ontology
that fits  $(\Psf,\Nsf)$ and in which every TGD has a single head atom.

\lead{Finite Bases.}
Let \Hsf be a non-empty finite   set of finite \Smc-instances, for some schema \Smc, and let
\Lmc be one of the ontology languages introduced above.
An \Lmc-ontology \Omc
over \Smc is a \emph{finite} \Lmc-\emph{basis} of  \Hsf
if for all \Lmc-TGDs $\rho$, the following holds:
$$
\Omc \models \rho \text{ iff } I \models \rho \text{ for all }
I \in \Hsf.
$$
If $\Hsf = \{ I \}$ is a singleton, we also say that  \Omc is a \emph{finite} \Lmc-\emph{basis} of  $I$. 
We study the following computational problem:

\begin{center}
  \fbox{ \parbox{0.90\linewidth}{
      \begin{tabular}{lp{0.70\linewidth}@{}}
        \textsc{Problem}: & \Lmc-basis construction      \\
        \textsc{Input}:   & A finite
        set $\Hsf \neq \emptyset$ 
        of finite instances. \\
        \textsc{Output:}  & A finite \Lmc-basis of \Hsf.
      \end{tabular}
    } }
\end{center}

We remind the reader that the entailment $\Omc \models \rho$
is defined relative to all instances, including infinite ones. An alternative definition of
finite bases would use entailment only over finite models. However, in almost all of the
cases studied in this paper, finite and unrestricted entailment coincide. The only
exception is the case of unrestricted TGDs~\cite{icalp/BeeriV81}.

The following lemma connects finite basis construction with fitting \Lmc-ontology existence.
Informally, it states that a finite basis of the positive examples is a canonical
candidate for a fitting \Lmc-ontology.
\begin{restatable}{lemma}{lemfittingviabasis}
\label{lem:fitting-via-basis} \label{lem:basisfits} 
  Let $(\Psf,\Nsf)$ be a fitting instance  and let $\Omc_\Psf$ be a finite \Lmc-basis of \Psf.
  Then $\Omc_\Psf$ fits $(\Psf, \Nsf)$ if and only if $(\Psf,
  \Nsf)$ has a fitting \Lmc-ontology.
\end{restatable}

For some choices of \Lmc, finite bases of a set \Hsf of instances
coincide with finite bases of the single instance obtained by taking the disjoint union of all instances  in \Hsf.
It in fact follows directly from the definition of finite bases
that this is the case
if \Lmc-ontologies \Omc are  \emph{invariant under disjoint union}, that is, for all non-empty
finite sets of instances \Hsf: 
$ {\biguplus}\Hsf \models \Omc$ if and only if 
$I\models \Omc \text{ for all } I\in\Hsf$.
It is well known that $\ELI_\bot$-ontologies are invariant under
disjoint union while this is not the case for any of the classes
of TGDs that we consider.
\section{Finite Bases for \texorpdfstring{\EL}{EL} and
  \texorpdfstring{\ELI}{ELI}}
\label{sect:finitebasesELELI}

We reprove that finite bases always exist in \EL
and prove that the same is true for \ELI, both
with and without `$\bot$'. We also show that, in both
cases, finite bases can be constructed in double
exponential time, and in single exponential time
if represented succinctly. This also establishes
upper bounds on the size of finite bases and we prove
matching lower bounds. Our main result is the following.
\begin{theorem}
  \label{thm:ELfinbase}
  In \EL, \ELI, $\EL_\bot$, and $\ELI_\bot$, finite bases can be constructed 
  in double exponential time, and in single exponential time if ontologies are represented succinctly.
\end{theorem}
By Lemmas~\ref{lem:succinctPTime} and~\ref{lem:fitting-via-basis}, we
immediately get the following.
\begin{corollary}
  \label{cor:ELfitexistExp}
  In \EL, \ELI, $\EL_\bot$, and $\ELI_\bot$, fitting ontology existence is decidable
  in \ExpTime and fitting ontologies can be constructed in double exponential time,
  and in exponential time if ontologies are represented succinctly.
\end{corollary}

We now prove Theorem~\ref{thm:ELfinbase}, treating all four cases
simultaneously. Thus, for the remainder of this section,
let $\Lmc \in \{ \EL,\ELI,\EL_\bot,\ELI_\bot \}$.
Since \Lmc-ontologies are invariant under disjoint union, we may without loss of generality concentrate on
finite bases of single instances.

Let \Iel be a finite \interpretation given as an input to the \Lmc-basis construction problem. A set $X \subseteq \Delta^\Iel$ is
\emph{\Lmc-definable} if there is an \Lmc-concept $C$ such that $X=C^\Iel$.
For every \Lmc-definable~\mbox{$X
  \subseteq \Delta^\Iel$}, choose an \Lmc-concept $E^*_X$ 
such that
$(E^*_X)^\Iel = X$. Note that if $\bot$ is an \Lmc-concept, then the empty
set is  \Lmc-definable. Otherwise, this is the case if and only if
$\Delta^I$ contains no value $d$ that is \emph{\Lmc-total} in $I$, meaning that $d \in C^\Iel$ for all \Lmc-concepts $C$ over the schema \Smc of $I$.
We generally assume that
$E^*_{\Delta^\Iel} = \top$ and, provided that $\bot$ is an \Lmc-concept, $E^*_\emptyset = \bot$.
\begin{exmp}
  Let $\Smc = \{ R \}$, $R$ a binary relation symbol, and let $I_1 = \{ R(a,b) \}$ and $I_2 = \{ R(a,b), R(c,c) \}$. Then the empty set is \EL-definable in $I_1$ by the concept $\exists R. \exists R. \top$ while the empty
  set is not \ELI-definable in $I_2$. It is, however, trivially $\EL_\bot$-definable by $\bot$. 
\end{exmp}
In the following, we sometimes consider \Lmc-concepts that contain subsets $X
\subseteq \Delta^\Iel$ as subconcepts. We interpret those in \Iel by setting
$X^\Iel = X$. 

\smallskip

Define the ontology $\Omc_\Iel$ to contain the following concept inclusions,
for all \Lmc-definable sets $X,X',Y$:
\begin{enumerate}

  \item $E^*_X \sqsubseteq A$ for all concept names $A$ with $\Iel \models X \sqsubseteq A$;

  \item $E^*_X \sqsubseteq \exists R. E^*_Y$ for all \Lmc-roles $R$ with $\Iel \models X \sqsubseteq
        \exists R. Y$;

  \item $A \sqsubseteq E^*_X$ for all concept names $A$ with $\Iel \models A \sqsubseteq X$;

  \item $\exists R. E^*_X \sqsubseteq E^*_Y$ for all \Lmc-roles  $R$ with $\Iel \models \exists R. X
        \sqsubseteq Y$;

  \item $E^*_{X} \sqcap E^*_{X'} \sqsubseteq E^*_Y$ if $\Iel \models X
        \sqcap X' \sqsubseteq Y$;

\end{enumerate}
We show the following in the appendix.
\begin{restatable}{lemma}{lemelfinitebasis}
      $\Omc_I$ is a finite \Lmc-basis of $I$.
\end{restatable}

The  construction of $\Omc_I$ hinges on effectively identifying the \Lmc-definable sets $X$ and constructing the corresponding \Lmc-concepts $E^*_X$.
We use brute force enumeration over all sets $X \subseteq \Delta^I$ and then,
for each $X$, apply the algorithm that achieves the following, based
on Lemma~\ref{lem:simtothemax}.
\begin{restatable}{lemma}{lemsmallconcepts}
  \label{lem:smallconcepts}
  There is an algorithm that, given as input an instance $I$ and a set $X \subseteq \Delta^I$, in double
  exponential time outputs an \Lmc-concept $C$ with $C^I=X$ if such a concept exists, and reports `undefinable' otherwise. The algorithm can be modified to run in single exponential time and output a succinct representation of $C$.
\end{restatable}

Results in formal concept analysis imply a single 
exponential lower bound on the size of finite \EL-bases, even when no
role names are present~\cite{Kuznetsov04,Kriegel_2024}.
The same is also known
when the schema contains a single role name and no concept names~\cite{jair/GuimaraesOPS23}. We establish a
tight double exponential lower bound that also applies to \ELI and a tight
single exponential lower bound  when the basis is represented succinctly. In fact, the following is a
consequence of Lemma~\ref{lem:fitting-via-basis} and Theorem~\ref{thm:size-lower-el-fittings} below.
\begin{theorem}\label{thm:size-lower-el-bases} Let $\Lmc\in \{\EL,\ELI, \EL_\bot, \ELI_\bot\}$.
  For every $n\geq 1$, there is a finite  instance $I_n$ such that
  \begin{enumerate}
    \item the size of $I_n$ is bounded by $p(n)$, $p$ a polynomial;
    \item the smallest finite \Lmc-basis $\Omc$ of $I_n$ is of size $2^{2^n}$,
          and of size $2^n$ when represented succinctly.
  \end{enumerate}
\end{theorem}

\section{Fitting Existence for \texorpdfstring{\EL}{EL} and \texorpdfstring{\ELI}{ELI}}
\label{sect:fittingELELI}

We  consider  fitting TGD existence and fitting ontology existence  in  \EL, \ELI, $\EL_\bot$, and $\ELI_\bot$,  proceeding in parallel. We establish 
semantic characterizations in terms of products and simulations which suggest an alternative algorithm for fitting existence
and construction with optimal complexity, reproving Corollary~\ref{cor:ELfitexistExp}. They also support
proving lower complexity bounds. 
\begin{exmp}
  Having $\bot$ or not makes a difference. Let $\Smc = \{ R \}$ with $R$ binary, $P=\{R(a,b)\}$, and $N=\{R(a,a)\}$. Then $\exists R. \exists R. \top \sqsubseteq \bot$
  fits $(\{P\},\{N\})$, but it follows from the characterizations below that there is no fitting \ELI-ontology.
\end{exmp}

We next give a model-theoretic characterization for fitting 
TGD existence. Via Lemma~\ref{lem:oneneg},
it also provides a characterization for 
fitting ontology existence.

\begin{restatable}{theorem}{thmmtcELfittingTGD}
  \label{thm:mtc:ELfittingTGD} Let $\Lmc \in \{\EL,\ELI\}$. Let $(\Psf, \Nsf)$
  be a fitting instance where $\Nsf = \{N_1,\dots,N_k\}$ and let
  $P={\biguplus}\Psf$. Then no \Lmc-TGD fits $(\Psf,\Nsf)$ if and only if for
  every $\bar d =(d_1,\dots,d_k)\in \Delta^{\prod \Nsf}$ such that $d_i$ is
  non-\Lmc-total in $N_i$ for all $i \in [k]$, the following conditions are satisfied:
  \begin{enumerate}
    \item the set $S_{\bar{d}}=\{(P,e)\mid (\prod \Nsf, \bar d)\preceq_\Lmc (P, e)\}$ is non-empty;

    \item  $\prod S_{\bar{d}} \preceq_\Lmc (N_i,d_i)$ for some $i \in [k]$.

  \end{enumerate}
  The same is true for $\Lmc_\bot$ when the condition `$d_i$ non-\Lmc-total in $N_i$ for all $i\in[k]$' is dropped.
\end{restatable}
Theorem~\ref{thm:mtc:ELfittingTGD} provides us with an
alternative algorithm for fitting TGD existence and fitting ontology existence, by checking the conditions given there.  This reproves the upper bounds in fitting ontology existence from Corollary~\ref{cor:ELfitexistExp} and gives the same upper bounds for fitting TGD existence. The proofs are 
constructive in the sense that they also yield algorithms for fitting
TGD and ontology construction. Unlike in the approach via
finite bases, the constructed ontologies contain only
one TGD per negative example.
\begin{restatable}{theorem}{thmELIFittingInExp}
  In \EL, \ELI, $\EL_\bot$, and $\ELI_\bot$, fitting TGD existence is decidable
  in \ExpTime and fitting TGD construction is possible in double exponential time,
  and in exponential time if TGDs are represented succinctly. The same is 
  true for fitting ontology existence and construction.
\end{restatable}

We next show that the obtained upper bounds for fitting existence are optimal. This is done by a non-trivial 
reduction from
the product simulation problem, which is known to be 
\ExpTime-hard~\cite{iandc/HarelKV02}.
\begin{restatable}{theorem}{thmelfittingexistence}
\label{thm:el-fitting-existence} In \EL, \ELI, $\EL_\bot$, and
  $\ELI_\bot$, fitting TGD existence and fitting ontology existence are
  \ExpTime-hard.
\end{restatable}

We finally establish tight lower bounds on the size of fitting TGDs and ontologies. This builds on lower bounds on the size
of fitting \EL-concepts in concept learning, obtained in~\cite{Funk19}.
\begin{restatable}{theorem}{thmsizelowerelfittings}
\label{thm:size-lower-el-fittings} Let $\Lmc\in \{\EL,\ELI,\EL_\bot,\ELI_\bot\}$.
  For every $n\geq 1$, there is a fitting instance $(\Psf_n,\Nsf_n)$ such that
  \begin{enumerate}
    \item the size of $(\Psf_n,\Nsf_n)$ is bounded by $p(n)$, $p$ a polynomial;
    \item $(\Psf_n,\Nsf_n)$ admits a fitting \Lmc-TGD and a fitting \Lmc-ontology;
    \item the smallest \Lmc-TGD that fits $(\Psf_n,\Nsf_n)$ is of size at least $2^{2^n}$ and of size $2^n$ when represented succinctly, and the same 
    is true for the smallest fitting \Lmc-ontology.
  \end{enumerate}
\end{restatable}

\section{Finite Bases for TGDs}
\label{sect:finitebasesTGD}
We show that finite bases for guarded TGDs always exist
and that basis
construction is possible in single exponential time. 
This also yields a single exponential upper bound on the size of bases, and we establish a matching lower bound as well. For inclusion
dependencies, the existence of finite bases is trivial as there are
only finitely many IDs over a fixed schema~\Smc. For frontier-one TGDs, frontier-guarded TGDs, and full TGDs, we prove that finite bases do not always exist. Finite basis existence remains
open for unrestricted TGDs.

We remark that we have defined TGDs so that `$\bot$' is not admitted as
a head. The results in this section and the subsequent one can be adapted
to that case, in analogy with the differences between \ELI and $\ELI_\bot$
in the previous sections. We start with the following positive result.
\begin{theorem}
  \label{thm:GTDGfinbase}
  \GTGD-basis construction is possible
  in single exponential time. 
\end{theorem}
By Lemma~\ref{lem:fitting-via-basis}, we
immediately get the following.
\begin{corollary}
  \label{cor:GTGDfitexistExp}
 Fitting \GTGD-ontology existence is decidable
  in \coNExpTime and fitting GTGD-ontology construction is possible in single exponential time.
\end{corollary}

Since guarded TGDs are not invariant under disjoint union, we directly
construct finite bases of \emph{sets} of instances. We start by introducing
a preliminary notion.
Let $(I,\bar{a})$ be a pointed instance with $\bar{a}=\langle a_1,\dots,a_k
\rangle$. We use $(I^{*}, \bar{a}^{*})$ to denote the \emph{diversification} of $(I,\bar{a})$, that is, the pointed instance obtained
from $(I,\bar{a})$ by introducing fresh and distinct values $\bar{a}^{*} =
\langle a_1^{*}, \dots, a_k^{*} \rangle$ and adding to $I$ each $a_i^{*}$ as a
`clone' of $a_i$. More precisely, $I^{*}$ consists of all facts that can be
obtained from a fact in $I$ by replacing, for every~$a_i$, zero or more
occurrences of $a_i$ with~$a_i^{*}$. Note that $(I^{*}, \bar{a}^{*})
\rightarrow (I,\bar{a})$ while the converse does not hold in general,
since $\bar a$ may contain
repeated values whereas $\bar a^*$ does not (which is, in fact, the aim of diversification).

Let \Hsf be a non-empty finite set of \Smc-instances, for some schema \Smc. We construct
an ontology $\Omc_\Hsf$ that contains one guarded TGD for each of the (finitely
many) guarded CQs over~\Smc. Let $q(\bar{x})=\exists \bar{y} \,
\varphi(\bar{x},\bar{y})$ be such a
CQ.\@
We distinguish two cases:
\begin{enumerate}

  \item $q(I) = \emptyset$ for all $I\in\Hsf$.

        Then $\Omc_\Hsf$ contains the rule $\varphi \rightarrow \psi$ where
        $\psi$ is the conjunction of all possible atoms that use only the
        variables from $\bar{x}$ and the relation symbols from \Smc;

  \item $q(I) \neq \emptyset$ for some $I\in\Hsf$. Set
        $$
        S = \{(I,\bar{a})\mid I\in \Hsf, \bar{a}\in q(I)\},
        $$
         let $(P,\bar{b})=\prod S$ and let $q_{(P^*,\bar{b}^*)}=\exists \bar{z} \, \psi$
        be the canonical CQ  
        of $(P^*,\bar{b}^*)$. Then $\Omc_\Hsf$ contains the rule \mbox{$\varphi
          \rightarrow \exists \bar{z} \, \psi'$} where $\psi'$ is obtained from
        $\psi$ by renaming the answer variables of
        $q_{(P^*,\bar{b}^*)}$ to $\bar{x}$. Note that the renaming relies on
        $\bar{b}^*$ not containing repeated values.
\end{enumerate}

\begin{exmp}
  Let \Hsf consist of the single \Smc-instance $I= \{ R(a,a) \}$ where $\Smc = \{ R \}$ with $R$ a binary relation. The guarded CQ $q(x)=\exists y \,
  R(x,y)$ induces in $\Omc_\Hsf$ the  TGD
  $$
  R(x,y) \rightarrow \exists z \, \big(R(x,x) \wedge R(x,z) \wedge
  R(z,x) \wedge R(z,z)\big).
  $$
  Now consider the guarded CQ $q(x,y)=R(x,y)$. It induces in $\Omc_\Hsf$ the
  TGD
  \begin{align*}
    R(x,y) \rightarrow \exists z \, \big(
     & R(x,x) \wedge R(x,y) \wedge R(x,z) \wedge   \\
     & R(y,x) \wedge R(y, y) \wedge  R(y,z) \wedge \\
     & R(z,x) \wedge R(z,y) \wedge R(z,z) \big).
  \end{align*}
  Note that this may be viewed as a weakening of the equality-generating dependency
  $
  R(x,y) \rightarrow x=y
  $, 
  which is true in $I$, but not  expressible as a guarded TGD.
\end{exmp}

We show the following in the appendix.
\begin{restatable}{lemma}{lemFiniteBasesGTGDCorrect}\label{lem:GTGDbasis}
  $\Omc_\Hsf$ is a finite GTGD-basis of \Hsf.
\end{restatable}
The basis $\Omc_\Hsf$ constructed above is of double exponential size: there are double exponentially many possible guarded CQs which are all  used as the body of a TGD in~$\Omc_\Hsf$. 
Each single TGD, however, is only of single exponential size.
To reduce the size of $\Omc_\Hsf$ to single exponential,
it is thus enough to show that we can select single exponentially many TGDs from $\Omc_\Hsf$ and still obtain a finite basis of \Hsf. To achieve
this, we take inspiration from the case of $\Lmc \in \{\EL,\ELI\}$ where we were concerned only with subsets of instances that are \Lmc-definable. 
Analogously, in Point~2 of the definition of $\Omc_\Hsf$, we should only
be interested in sets $S$ that are definable by a guarded CQ $q$.
In contrast to the case of \EL and
\ELI, however, it does not suffice to choose one guarded CQ for every definable
set $S$. Instead, we need to consider all guarded CQs $q$ that define $S$ and are minimal in the sense that any CQ obtained from
$q$ by dropping an atom no longer defines~$S$.
This gives rise to the following.
\begin{restatable}{theorem}{thmsmallgtgdbasis}
Let $\Omc_\Hsf'$ be the subset of $\Omc_\Hsf$ that contains all TGDs from $\Omc_\Hsf$ whose body has at most $n:=||\Hsf||+1$ atoms. Then $\Omc_\Hsf'$ is a GTGD-basis of \Hsf with $O(S^n\cdot\ell^{\ell n})$ TGDs where $S$ is the number of relation symbols in the schema \Smc of \Hsf, and $\ell$ is the maximal arity of a symbol in~\Smc. 
\end{restatable}
We now consider lower bounds on the size of GTGD-bases.
As a consequence of Lemma~\ref{lem:fitting-via-basis} and Theorem~\ref{thm:TGDfitsizelower} below, we may obtain a single
exponential lower bound. That bound,
however, does not apply to singleton sets \Hsf. Here, we show
the same result also for this case.
\begin{restatable}{theorem}{thmsizelowergtgdontologiesimproved}
  \label{thm:size-lower-gtgd-ontologies-improved}
  For every $n\geq 1$, there is an instance $I_n$ (over a schema with only unary and binary relations) such that
  \begin{enumerate}
    \item the size of $I_n$ is bounded by $p(n)$, $p$ a polynomial;
    \item the smallest finite GTGD-basis $\Omc$ of $I_n$ is of size $2^n$.
  \end{enumerate}
\end{restatable}
\noindent\begin{proof}\
Let $P$ and $R$ be unary and binary symbols, respectively.
For $m\geq 1$, let $L_m$ denote the ``lasso'' instance, with values 
$a_0^m, \ldots, a_{2m-1}^m$
and facts
$$
\{R(a_i^m, a_{i+1}^m)\mid i<2m-1\}\cup {}  \{R(a_{2m-1}^m,a_m^m), P(a_m^m)\}
$$
Then define, for $n\geq 1$, 
$I_n=\bigcup_{i=1}^n \left(L_{p_i}\cup \{A(a_0^{p_i})\}\right)$
where $p_i$ is the $i$-th prime number. We show in the appendix
that $I_n$ is as required.
\end{proof}

Let us now discuss finite bases for inclusion dependencies.
Over any schema \Smc with relations of maximum arity $k$, there are at most 
$|\Smc|^2(2k)^{2k}$ many inclusion dependencies.
Therefore, a finite basis for a non-empty set \Hsf of finite instances always exists:
we can simply use  the set of all 
INDs that are true in~\Hsf.  If the arity of relation symbols is bounded by a constant, then this basis is only of polynomial size. 
Otherwise, exponential size cannot be avoided.
\begin{restatable}{theorem}{thmsizelowerfulltgdbases}
\label{thm:size-lower-fulltgd-bases}   
There are finite instances $I_1,I_2,\dots$ such that for all $n \geq 1$, 
the size of $I_n$ is polynomial in $n$, but all finite $\IND$-bases of 
$I_n$ contain at least $2^n$ {\IND}s.
\end{restatable}

Via Lemma~\ref{lem:fitting-via-basis} and the fact that {\IND}s can be evaluated in polynomial time, we 
also obtain decidability of fitting \IND-ontology
existence in \ExpTime. However, we will see in the
subsequent section that this complexity is
not optimal and thus do not state the result here as a formal theorem. We also obtain the following.
\begin{theorem}
\label{thm:fittingindconstrinexp}
  Fitting \IND-ontology construction is possible
  in single exponential time.
\end{theorem}
We do not know whether the bound
in Theorem~\ref{thm:fittingindconstrinexp} is optimal.

We next prove that for frontier-guarded TGDs and for frontier-one TGDs, finite bases
do not always exist. 
\begin{theorem}
\label{thm:fgtgdnobasis}
  There are instances $I$ that have no finite \FGTGD-basis and no finite \FONETGD-basis.
\end{theorem}
To prove Theorem~\ref{thm:fgtgdnobasis}, we use the rather simple instance
$I = \{ R(a,b), R(b,a)\}$.
For every
$n\geq 1$, consider the frontier-one TGD
$$\rho_n = \bigwedge_{i\in[n-1]} R(x_i, x_{i+1}) \wedge R(x_n, x_1) \rightarrow R(x_1,x_1).$$ 
The TGD  expresses that if $x_1$ lies on a cycle of length $n$, then
$x_1$ has a reflexive loop. We have $I\models\rho_n$ for all odd $n$ because no cycle of odd length homomorphically maps to $I$.
Note that the rule bodies of the TGDs $\rho_n$ with $n$ odd get
larger with increasing~$n$. Intuitively, this means that
also the rule bodies of any finite \FGTGD-basis of $I$ must be 
of unbounded size, which means that there is no finite \FGTGD-basis.
In the appendix, we formally prove that $I$ indeed neither has a finite \FGTGD-basis nor
a finite \FONETGD-basis.

We now consider full TGDs. The instance $I$
used in the proof of Theorem~\ref{thm:fgtgdnobasis} is not suitable
here because the set
$$
\begin{array}{rl}
  \Omc_I = \{ & R(x,y) \rightarrow R(y,x) \\[1mm]
  & R(x,y) \wedge R(y,z) \wedge R(z,u)
  \rightarrow R(x,u) \\[1mm]
  & R(x,x) \wedge \mn{true}(y) \wedge \mn{true}(z) \rightarrow R(y,z) \ \}
\end{array}
$$
is a finite \FTGD-basis of $I$. The third
TGD above  represents four TGDs,
as $\mn{true}(v)$ is a placeholder for
either $R(u,v)$ or $R(v,u)$ with $u$ a
fresh variable and $v\in\{y,z\}$. In fact, we even have
the following.
\begin{restatable}{lemma}{lemfinitefullbasisexists}
  $\Omc_I$ is a finite \FTGD-basis of $I$ and also a finite TGD-basis.
\end{restatable}
 Let $J=\{R(u,v)\mid u,v\in\{a,b,c\},\ u\neq v \}$.
 It is well known and easy to see that an undirected graph $G$ has a homomorphism to the undirected 3-clique if and only if
    $G$ is 3-colorable. Note that $J$ is essentially an undirected 3-clique except that undirected edges are replaced
    with bidirectionally directed edges. We 
    may modify any undirected graph $G$ in the same way, making it bidirectional.
    For every graph $G$, consider
    the full TGD
    $$
    \rho_G = \varphi_G \rightarrow R(x,x)
    $$
    where $\varphi_G$ is the conjunction of the edges of $G$ viewed as $R$-atoms and
    $x$ is a vertex in $G$ chosen arbitrarily. By what was said above, it
    is clear that $J \models \rho_G$ if and
    only if $G$ is not 3-colorable. 
    It is known that for any $m \geq 0$, there are non-3-colorable
    graphs of girth exceeding $m$~\cite{Erdoes59}. We remind the reader that the girth of a graph is the length of a shortest cycle in it (and $\infty$ if the graph is acyclic). Intuitively, this means that also the rule bodies of
    any finite \FTGD-basis of $J$ must be
    of unbounded size, and thus there is no such basis.
\begin{restatable}{theorem}{thmftgdnobasis}
\label{thm:ftgdnobasis}
  There are instances $J$ that have no finite \FTGD-basis.
\end{restatable}
We remark in passing that there appears to
be a loose connection between finite 
\FTGD-bases and constraint satisfaction problems. Indeed, it is not very difficult
to see that 
any finite \FTGD-basis \Omc of a finite instance $I$ gives
rise to a datalog-rewriting of the 
complement of the (non-uniform) constraint satisfaction
problem CSP$(I)$ obtained by using $I$ as
a template. For more background
on these notions, we refer to~\cite{siamcomp/FederV98}.

\section{Fitting Existence for TGDs}
\label{sect:fittingTGD}

We study fitting TGD existence and fitting ontology existence for various classes of TGDs. As in Section~\ref{sect:fittingELELI}, we start with
semantic characterizations that are the basis
for developing decision procedures and determining
the computational complexity.
In the case of {\GTGD}s,
the characterization allows us to reprove the 
upper complexity bound from Corollary~\ref{cor:GTGDfitexistExp}. For frontier-guarded
TGDs, frontier-one TGDs, and full TGDs, where the approach via finite bases is precluded, we may nevertheless use our characterizations
to obtain  algorithms for fitting
TGD existence and fitting ontology existence.
The same is true for unrestricted TGDs, for which the
existence of finite bases remains open.
We identify tight complexity bounds for all studied
existence problems, with the exception of full TGDs and unrestricted TGDs where a gap remains, and also tight bounds on the size of
fitting TGDs and ontologies. 

\smallskip

We start with the case of inclusion dependencies. Here we skip a characterization 
because, due to the simplicity of {\IND}s, they are rather
convenient to deal with directly. Since
an \IND that fits a given fitting instance must be of size linear in the size of that
instance, fitting \IND existence is clearly
in \NPclass. By Lemma~\ref{lem:oneneg}, the
same is true for fitting \IND-ontologies.
We prove a matching lower bound by reduction
from 3SAT. A core idea is to use two
relation symbols $R$ and $S$ that provide one
position for each literal in the 3SAT
formula given as an input, and to represent
truth values in an IND by distinguishing whether a position in the head atom holds the same variable as 
the corresponding position in the body atom or an existentially quantified variable.
\begin{restatable}{theorem}{thmfittingexindcomplexity}
  \label{thm:fitting-ex-ind-complexity}
  Fitting \IND existence and fitting \IND-ontology existence are \NPclass-complete.
\end{restatable}

Turning to other classes of TGDs, we first give some preliminaries. Let $I$ be an \Smc-instance. A $k$-tuple
$\bar{a}\subseteq \mn{adom}(I)$ is \emph{total} in $I$ if $\bar{a} \in q(I)$ for all $k$-ary CQs $q(\bar{x})$ over \Smc. Note that
this is the case if and only if $I$ contains every possible fact built from a
relation symbol in \Smc and values from $\bar a$. A set $M \subseteq
\mn{adom}(I)$ is \emph{guarded} if there is a fact $R(\bar{a})\in I$ such that
$M\subseteq \bar{a}$, and $M$ is \emph{maximally guarded} if there is no guarded
set $N\subseteq \mn{adom}(I)$ with $M\subsetneq N$. Fix an arbitrary order on
$\mn{adom}(I)$. For a set $M \subseteq \mn{adom}(I)$, we write $\bar{M}$ to
denote the tuple that contains each element of $M$ exactly once, adhering to the
fixed order. 
By $I\vert_M$ we denote the subset of $I$ that consists of precisely those facts
that use only values from $M$. 
Consider instances $I_1, \ldots, I_k$ and an $n$-tuple
$$\bar{b} \in \mn{adom}(\prod_{i=1}^k I_i)^n.$$
We may write $\bar b$ as $\bar a_1 \times \cdots \times \bar a_k$ where $\bar a_i\in \mn{adom}(I_i)^n$ for all $i\in[k]$. We then use $\bar b[i]$ to denote $\bar a_i$. We next present a
characterization of fitting GTGD existence. Via 
Lemma~\ref{lem:oneneg}, it also applies to
fitting ontology existence.
\begin{restatable}{theorem}{thmmtcGTGDfittingTGD}
  \label{thm:mtc:GTGDfitting} Let $(\Psf, \Nsf)$ be a fitting instance where
  $\Nsf = \{N_1,\dots,N_k\}$. Then no \GTGD fits $(\Psf,\Nsf)$ if and only if
  for every non-empty maximally guarded set $M \subseteq \mn{adom}(\prod \Nsf)$ such that
  $\bar{M}[i]$ is non-total in $N_i$ for all $i\in [k]$,
  the following conditions are satisfied: 
  \begin{enumerate}
    \item the following set is non-empty:
        $$
          \begin{array}{r@{\;}c@{\;}l}
            S_M &=&
            \{(J,\bar{b}) \mid J\in\Psf \text{ and } \bar{b}\in\mn{adom}(J)^{|M|} \text{ such that}\\[1mm]
            && \hspace*{12mm} (\prod \Nsf\vert_M,\bar{M}) \to (J,\bar{b})\}
           \end{array}$$
    \item $\exists i \in [k]$: $(K^*, \bar{c}^*) \to (N_i, \bar{M}[i])$
          where $(K, \bar{c}) =
          \prod S_M$.
  \end{enumerate}
\end{restatable}
Characterizations for other 
classes of TGDs can be obtained by varying the characterization
in Theorem~\ref{thm:mtc:GTGDfitting}. For brevity, we only give the differences.
For the convenience of the reader, all three characterizations are given in an explicit form in the
appendix.
\begin{theorem}
\label{thm:charthreeinone}
  Consider Theorem~\ref{thm:mtc:GTGDfitting} modified so that
  in the definition of the set $S_M$,`$(\prod \Nsf\vert_M,\bar{M})$' is replaced with `$(\prod \Nsf,\bar{M})$'. The resulting theorem holds
  for
  \begin{enumerate}
  
      \item \FGTGD;  

     \item \FONETGD when, additionally, singleton sets
      are considered for $M$ instead of maximally guarded
      sets;
      
      \item \TGD when, additionally, unrestricted sets
      are considered for
      $M$ instead of maximally guarded
      sets.
      
  \end{enumerate}
\end{theorem}
The characterization for full TGDs requires
some more changes, we state a self-contained version.
\begin{restatable}{theorem}{thmmtcFTGDfittingTGD}
  \label{thm:mtc:FTGDfitting}
  Let $(\Psf,\Nsf)$ be a fitting instance where $\Nsf = \{N_1,\dots,N_n\}$. Then
  no \FTGD fits
  $(\Psf,\Nsf)$ if and only if, for all relation symbols
  $R_1,\dots,R_n$ and tuples $\bar a_1,\dots,\bar a_n$
   such that
  $$
    \bar{a}_i \in \mn{adom}(\prod \Nsf)^{\mn{ar}(R_i)} \enspace\text{and}\enspace R_i(\bar a_i[i]) \notin N_i \quad\text{for } i\in[n],
  $$
  there is a $P\in\Psf$ and a homomorphism $h$ from $\prod \Nsf$ to $P$
  such that $R_j(h(\bar a_j))\notin P$
    for some $j\in[n]$.

  Moreover, if $(\Psf,\Nsf)$ admits a fitting \FTGD, then it admits one in which
  the number of head atoms is bounded by the number of examples in \Nsf.
\end{restatable}
The above characterizations give rise to algorithms for fitting TGD
existence and fitting ontology existence.
\begin{restatable}{theorem}{thmtgdfittingupper}
\label{thm:tgdfittingupper}~\\[-4mm]
\begin{enumerate}
  \item For $\Lmc \in \{ \GTGD, \FGTGD, \FONETGD\}$, fitting \Lmc-ontology existence and fitting \Lmc-TGD existence are in \coNExpTime. 

  \item For full TGDs, fitting ontology existence is in $\Sigma^p_2$ (and in \coNPclass if the arities of relation symbols are bounded by a constant), and fitting
  TGD existence is in \coNExpTime.
  
  \item For unrestricted TGDs, fitting ontology
  existence is in \coTwoNExpTime and fitting TGD existence
  is in \coThreeNExpTime.

\end{enumerate}
\end{restatable}
To prove Point~1 of Theorem~\ref{thm:tgdfittingupper}, the characterizations for 
GTGDs can be implemented straightforwardly. Note that the number of sets $M$ is
linear and thus the sets
$S_M$ can be computed deterministically in single
exponential time, checking 
$(\prod \Nsf\vert_M,\bar{M}) \to (J,\bar{b})$ by brute force enumeration. 
For frontier-guarded and frontier-one TGDs, in contrast, there is no obvious way to compute the sets $S_M$ in \coNExpTime. This is 
because $(\prod \Nsf|_M, \bar M)$ is replaced by
$(\prod \Nsf,\bar M)$, which
has single exponentially many values rather than linearly many, and thus brute force enumeration no longer works. In the
appendix, we show how to get around this problem.

For Point~2, the characterization can be implemented straightforwardly. The difference
in complexity between fitting ontology existence and fitting TGD existence is due to the fact that, by Lemma~\ref{lem:oneneg},
fitting ontology existence corresponds to fitting TGD existence with a single negative example. But in this case the product $\prod \Nsf$ is of course of linear size only, rather than of single exponential size.

For Point~3, there are double exponentially many sets $M$
to be considered (as these no longer need to be guarded) and 
the sets  $S_M$ are of double
exponential size and the products
$\prod S_M$ are of triple exponential size.
This explains the \coThreeNExpTime upper bound for fitting TGD existence. If there is only a single negative
example as in ontology fitting via Lemma~\ref{lem:oneneg}, the number of sets $S_M$ and the
size of the products $\prod S_M$ reduce
by one exponential, and the complexity drops to \coTwoNExpTime.

\smallskip

In the proofs of Theorems~\ref{thm:mtc:GTGDfitting},~\ref{thm:charthreeinone},~and~\ref{thm:mtc:FTGDfitting}, concrete fitting TGDs and ontologies
are constructed. A straightforward analysis of their size yields the following.
\begin{restatable}{theorem}{thmtgdfittingsizeupper}~\\[-4mm]
\begin{enumerate}
  \item
 Let $\Lmc \in \{ \GTGD, \FGTGD, \FONETGD \}$ and let $(\Psf,\Nsf)$ be a fitting
 instance. If there is a fitting \Lmc-TGD or a
 fitting  \Lmc-ontology for $(\Psf,\Nsf)$, then there is one of single exponential size that can be constructed in double exponential time. 

  \item The same is true for full TGDs where, in the case of fitting ontologies, there is even a fitting ontology of polynomial size that can be
  constructed in single exponential time.
 
  \item If  there is a fitting TGD (fitting TGD-ontology) for $(\Psf,\Nsf)$, then there is one of triple (double) exponential size that can be constructed in quadruple (triple) exponential time. \end{enumerate}
\end{restatable}

We now establish lower complexity bounds.
\begin{restatable}{theorem}{thmfittingextgdcomplexity}
  \label{thm:fitting-ex-tgd-complexity}~\\[-4mm]
  \begin{enumerate}
    \item   
  Let $\Lmc \in \{ \GTGD, \FGTGD, \FONETGD, \TGD\}$. Then fitting \Lmc-TGD existence and fitting \Lmc-ontology existence are \coNExpTime-hard. 
  
  \item For full TGDs,  fitting TGD existence  is \coNExpTime-hard
  and  fitting ontology existence is DP-hard (and \coNPclass-hard if
  the arities of relation symbols are bounded by a constant).
   \end{enumerate}
\end{restatable}
We thus obtain \coNExpTime-completeness for GTGD, FGTGD, and F1TGD.
All \coNExpTime lower bounds in 
Theorem~\ref{thm:fitting-ex-tgd-complexity}
are proved by 
reduction from the product homomorphism problem~\cite{CateD15} and apply already to schemas that contain only binary relation symbols. The \coNExpTime lower bounds in Points~1 and~2 are based on different constructions. This is because the former
exploit the unrestricted heads of the TGD classes considered there while the latter exploits the unrestricted bodies of full TGDs.

We finally turn to lower bounds on the size of fitting TGDs
and ontologies. These can be derived from results implicit in~\cite{CateD15,cp/Willard10}. The following theorem yields lower bounds for the size of fitting TGDs and ontologies for the cases of \GTGD, \FGTGD, \FONETGD, and unrestricted TGDs.
\begin{restatable}{theorem}{thmTGDfitsizelower}
\label{thm:TGDfitsizelower}
For every $n\geq 1$, there is a fitting instance $(\Psf_n,\Nsf_n)$ such that
  \begin{enumerate}
  
    \item the size of $(\Psf_n,\Nsf_n)$ is bounded by $p(n)$, $p$ a polynomial;
    
    \item $(\Psf_n,\Nsf_n)$ admits a fitting guarded and frontier-one TGD;

    \item the smallest TGD fitting $(\Psf_n,\Nsf_n)$ has size $2^n$.
  \end{enumerate}
  The same is true for ontologies instead of single TGDs. 
\end{restatable}
Note that for GTGD, FGTGD, and F1TGD, these bounds are tight.
We have the same lower bounds also for full TGDs (where they are not tight).
\begin{restatable}{theorem}{thmfulltgdsize}
\label{thm:fulltgdsize}
For every $n\geq 1$, there is a fitting instance $(\Psf_n,\Nsf_n)$ such that
  \begin{enumerate}
  
    \item the size of $(\Psf_n,\Nsf_n)$ is bounded by $p(n)$, $p$ a polynomial;
    
    \item $(\Psf_n,\Nsf_n)$ admits a fitting full TGD;

    \item the smallest full TGD fitting $(\Psf_n,\Nsf_n)$ has size $2^n$.
  \end{enumerate}
  
\end{restatable}
For inclusion dependencies, it is
an easy consequence of Lemma~\ref{lem:oneneg} that if a fitting \IND-ontology exists, then there is one of polynomial size. 

\section{Conclusion}

We have studied finite bases as well as TGD and ontology
fittings for several description logics and classes of
TGDs. 
We mention some interesting open questions. We would like
to know the exact complexity of deciding fitting TGD existence and fitting ontology existence for unrestricted
TGDs. Note that the same gap left open here also shows up
in~\cite{CateD15}. We would also
like to know whether finite bases always exist for unrestricted TGDs. We conjecture that this is not the 
case and that the instance $J$ used in
Section~\ref{sect:finitebasesTGD} to show that finite \FTGD-bases need not exist can be used to show that. 
For the time being, however, we do not have a proof. It 
would also be interesting to characterize precisely
the sets of instances \Hsf that have a finite basis
of frontier-guarded TGDs, and likewise for frontier-one
TGDs and full TGDs. 

\section*{Acknowledgments}
This work is partly supported by BMFTR in DAAD project 57616814 (\href{https://secai.org/}{SECAI}).
The second and third authors were supported by DFG project JU~3197/1-1.

\newpage

\bibliographystyle{named.updated}
\bibliography{kr25}

\cleardoublepage

\appendix

\section{Some Basic Lemmas}\label{appendix}

\begin{lemma}
  \label{lem:simulations} Let $(I,d)$ and $(J,e)$ be pointed instances and $\Lmc \in \{\EL,\ELI\}$. Then
  $(I,d) \preceq_{\Lmc} (J,e)$ implies that for every $\Lmc_\bot$
  concept $C$,
  $
  d\in C^I \text{ implies } e \in C^J.
  $
\end{lemma}

\begin{lemma}
  \label{lem:product_homomorphism} Let $S$ be a non-empty finite set of $k$-ary
  pointed instances with $(I, \bar{a})\in S$ and let $(J,\bar{b}) = \prod S$.
  Then $(J,\bar{b}) \to (I, \bar{a})$.
\end{lemma}

\begin{lemma}
  \label{lem:product_cq} Let $S$ be a non-empty finite set of $k$-ary pointed
  instances and $(J,\bar{b}) = \prod S$. Then for every $k$-ary CQ $q(\bar{x})$:
  $$
  \bar{b} \in q(J) \text{\enspace if and only if\enspace} \bar{a}\in q(I)
  \text{ for all } (I,\bar{a})\in S.
  $$
\end{lemma}

\begin{lemma}
  \label{lem:prodELI} Let $(I_1,a_1),\dots,(I_n,a_n)$ be pointed instances,
  $(P,\bar a)=\prod_{1 \leq i \leq n} (I_i,a_i)$,
  and $C$ an $\ELI_\bot$-concept. Then $\bar a \in C^P$ if and only if $a_i \in C^{I_i}$
  for $1 \leq i \leq n$.
\end{lemma}

\section{Proofs for Section~\ref{sect:prelim}}

\lemsimtothemax*

\noindent\begin{proof}\
  Let $I$ and $J$ be finite interpretations. We
\begin{enumerate}

\item start with
the relation $Z_0$ that consists of all pairs $(d,e) \in \Delta^{I} \times \Delta^{J}$
such that for all concept names $A$, $d \in A^{I}$ implies $e \in A^{J}$
and then

\item construct a sequence of relations $Z_0,Z_1,\dots$ where 
we obtain $Z_{i+1}$ from $Z_i$ by starting with $Z_{i+1}=Z_i$ and then deleting
as follows:  if $(d,e) \in Z_i$ and $(d,d') \in R^{I}$, with $R$ an \Lmc-role, and there is no $(e,e') \in R^{J}$ with $(d',e') \in Z_i$,
then remove $(d,e)$ from $Z_{i+1}$. 

\end{enumerate}
The process stabilizes after $m \leq |\Delta^I| \cdot |\Delta^J|$ rounds because that is the maximum number of
pairs in $Z_0$. Each round clearly also only needs
polynomial time. 
It is well known and easy to prove that the resulting relation $Z_m$ is the maximal \Lmc-simulation (w.r.t.\ set inclusion) from $I$ to $J$. Thus  $(I,d) \preceq_\Lmc (J,e)$ iff $(d,e) \in Z_m$.

The procedure also allows us to identify for every pair $(d,e) \in (\Delta^{I}\times\Delta^{J}) \setminus Z_m$ a concept $C_{d,e}$
with $d \in C_{d,e}^{I}$ and $e \notin C_{d,e}^{J}$:
\begin{itemize}

    \item If $(d,e) \notin
Z_0$, then we may choose as $C_{d,e}$ any concept name $A$ with $d \in A^{I}$ and $e \notin A^{J}$.

    \item If a pair $(d,e)$ is deleted in Step~2 because
    of some $(d,d') \in R^{I}$, then we set $C_{d,e}$ to $\exists R. D$ with $D$ the
    conjunction of $C_{d',e'}$, for all $(e,e') \in R^{J}$ (the empty conjunction being $\top$).
    
\end{itemize}
 Let us analyze the size
of the constructed concepts $C_{d,e}$. An easy analysis
 reveals that the role depth of each concept $C_{d,e}$
is bounded by $|\Delta^\Iel| \cdot |\Delta^J|$, the  maximum number of rounds of the algorithm, and that the outdegree is bounded by
$|\Delta^J|$. Consequently, $C_{d,e}$ is of size at most single exponential. Let us now consider  succinct
representation. Clearly, every concept
$C_{d,e}$ constructed by the algorithm is a concept name or 
of the form $\exists R. D$ with $D$ a
conjunction of concepts $C_{d',e'}$. Moreover, there are 
at most $|\Delta^\Iel| \cdot |\Delta^J|$ concepts. It follows that
the number of subconcepts of each concept $C_{d,e}$ is 
 $O((|\Delta^\Iel| \cdot |\Delta^J|)^2)$, and thus the 
 succinct representation of $C_{d,e}$ is of polynomial size. Moreover, it is easy to see that the succinct representations of the concepts $C_{d,e}$ can be constructed
in polynomial time, in parallel to executing the elimination procedure.
\end{proof}

\lemfittingviabasis*

\noindent\begin{proof}\
  Since ``$\Rightarrow$'' is trivial, we consider ``$\Leftarrow$''. Assume
  that $\Omc_\Psf$ does not fit $(\Psf, \Nsf)$ and take any  \Lmc-ontology \Omc. We have
  $I\models\Omc_\Psf$ for all $I\in\Psf$ and hence there must be a $J\in\Nsf$ such that $J\models \Omc_\Psf$. We have to show that \Omc does not fit $(\Psf,\Nsf)$. If $I \not \models \Omc$ for some $I \in \Psf$, then we
  are done. Otherwise every TGD in \Omc is satisfied in all instances in \Psf
  and thus $\Omc_\Psf \models \Omc$ by definition of finite bases. But then $J \models \Omc$ and thus \Omc does not fit $(\Psf,\Nsf)$.
\end{proof}

\section{Proofs for Section~\ref{sect:finitebasesELELI}}

We want to show that the ontology $\Omc_I$ constructed in the main part of the paper is a finite \Lmc-basis of $I$. A central step is to
prove the following lemma, where $\Omc \models C \equiv D$ is an abbreviation
for $\Omc \models C \sqsubseteq D$ and  $\Omc \models D \sqsubseteq C$.
\begin{restatable}{lemma}{lemmain}
  \label{lem:main}
  $\Omc_\Iel \models C \equiv E^*_{C^\Iel}$ for all \Lmc-concepts $C$.
\end{restatable}

\noindent\begin{proof}\
  The proof is by induction on the structure of $C$. We show both directions
  simultaneously. In the induction start, $C$ is either a concept name, $\top$
  or $\bot$, if $\bot$ is an \Lmc-concept. First assume that $C = A$ is a concept name. The semantics yields $\Iel \models A \equiv A^\Iel$ and thus
  $\Omc_\Iel \models A \equiv E^*_{A^\Iel}$ by Points~1 and~3 of the
  construction of $\Omc_\Iel$. For $\top$ and $\bot$,  $\Omc_I \models E^*_{\top^\Iel} \sqsubseteq \top$ and
  $\Omc_I \models \bot \sqsubseteq E^*_{\bot^\Iel}$ hold trivially by the semantics. Regarding
  the  other directions, Point~6 yields the desired result for $\top$
  and Point~7 for $\bot$.

  For the induction step, we begin with $C = D_1 \sqcap D_2$.  The semantics
  yields $I \models D_1^I \sqcap D_2^I \sqsubseteq C^I$ and 
   $I \models C^I \sqcap C^I \sqsubseteq D_i^I$ for $i \in \{1,2\}$,
   and thus 
  from Point~5
  of the construction of $\Omc_I$ we obtain
    \begin{align*}
     & \Omc_\Iel \models E^*_{D_1^\Iel} \sqcap E^*_{D_2^\Iel} \sqsubseteq E^*_{C^\Iel} \text{ and} \\
     & \Omc_\Iel \models E^*_{C^\Iel} \sqsubseteq E^*_{D_i^\Iel} \text{ for } i \in \{1,2\}.
  \end{align*}
Therefore
$$\Omc_\Iel \models  E^*_{C^\Iel} \equiv E^*_{D_1^\Iel} \sqcap E^*_{D_2^\Iel} 
.$$
    Applying the induction hypothesis yields $\Omc_\Iel \models D_i \equiv
  E^*_{D_i^\Iel}$ for $i \in \{1,2\}$ and thus we may conclude from 
  the above that
  $$\Omc_\Iel \models D_1 \sqcap D_2 \equiv  E^*_{C^\Iel}.$$

  Now consider $C = \exists R. D$ with $R$ an \Lmc-role. By the semantics, 
$\Iel \models \exists R.
  D^\Iel \equiv (\exists R. D)^\Iel$. By Points~2 and~4 of the construction of
  $\Omc_\Iel$, we obtain
  \begin{align*}
    \Omc_\Iel \models \exists R. E^*_{D^\Iel}\equiv E^*_{(\exists R. D)^\Iel}.
  \end{align*}
  The induction hypothesis yields $\Omc_\Iel \models D \equiv
  E^*_{D^\Iel}$, which, combined with the above gives $\Omc_\Iel \models \exists R. D
  \equiv E^*_{(\exists R. D)^\Iel}$, as required.
\end{proof}

Based on Lemma~\ref{lem:main}, the following is easy to show.
\lemelfinitebasis*
\noindent\begin{proof}\
  We have to show that
  $\Omc_I \models C \sqsubseteq D$ iff \mbox{$I \models C \sqsubseteq D$}, for all \Lmc-concept inclusions $C \sqsubseteq D$.

  \smallskip
  \noindent
  ``$\Rightarrow$''. It suffices to observe that $I \models \Omc_I$,
  which is immediate from the construction of $\Omc_I$.

  \smallskip
  \noindent
  ``$\Leftarrow$''. Assume that $\Iel \models C \sqsubseteq D$. Then
  $\Iel \models E^*_{C^\Iel} \sqsubseteq E^*_{D^\Iel}$ by definition
  of $E^*_{C^\Iel}$ and $E^*_{D^\Iel}$. By Point~5 of
  the construction of~$\Omc_\Iel$, this implies $E^*_{C^\Iel} \sqcap E^*_{C^\Iel}
  \sqsubseteq E^*_{D^\Iel} \in \Omc_\Iel$,  which of course gives
  $\Omc_\Iel \models E^*_{C^\Iel} \sqsubseteq E^*_{D^\Iel}$. Lemma~\ref{lem:main}
  yields $\Omc_\Iel \models C \sqsubseteq D$, as required.
\end{proof}

\lemsmallconcepts*

\noindent\begin{proof}\
Given an instance $I$ and a set $X \subseteq \Delta^I$, the algorithm claimed
to exist by Lemma~\ref{lem:smallconcepts} returns $\bot$ if $X=\emptyset$
and $\bot$ is an \Lmc-concept. If this special case does not apply, then
it first constructs
$(P,\bar d)= \prod_{d \in X} (I,d)$ (in single exponential time)  and then checks that $(P,\bar d) \not\preceq_\Lmc (I,e)$
for every $e \in \Delta^I \setminus X$, using the 
algorithm from Lemma~\ref{lem:simtothemax}. If the check fails for some~$e$, it outputs `undefinable'. Otherwise, for each $e$ it identifies an \Lmc-concept $C_{\bar d,e}$ with
$X \subseteq C_{\bar d,e}^I$ and $e \notin C_{\bar d,e}^I$ and returns the conjunction of all these concepts. This is clearly an \Lmc-concept that defines $X$. The desired
concepts $C_{\bar d,e}$ are in fact the concepts obtained from Lemma~\ref{lem:simtothemax}.  They satisfy $\bar d \in C_{\bar d,e}^P$ and $e \notin C_{\bar d,e}^I$, and by
 Lemma~\ref{lem:prodELI} the former implies $X \subseteq C_{\bar d,e}^I$, as required.
\end{proof}

\section{Proofs for Section~\ref{sect:fittingELELI}}

We first define
characteristic concepts of bounded depth. Let $\Lmc \in \{\EL,\ELI\}$ and let $(I,d)$ be a finite pointed instance.
For every
$i\in\mathbb{N}$, the concept $C^I_{\Lmc,i}(d)$ is defined by:
\begin{itemize}
  \item $C^I_{\Lmc,0}(d) = \top \sqcap \bigsqcap\{A\mid d\in
        A^\Iel\}$
  \item $C^I_{\Lmc,i+1}(d) = C^I_{\Lmc,0}(d) \sqcap
        \bigsqcap\limits_{\substack{R \Lmc\text{-role} \\
            (d,e)\in R^\Iel}} \exists R. C^I_{\Lmc,i}(e)$.
\end{itemize}
Note that the size of $C^I_{\Lmc,i}(d)$ is $O(|I| \cdot |\Delta^I|^i$)
and the number of subconcepts in  $C^I_{\Lmc,i}(d)$ is $O(|I|\cdot|\Delta^I|^2 \cdot i$).

\begin{restatable}{lemma}{lemFiniteUnravelling}
  \label{lem:FiniteUnravelling} Let $(\Iel, d)$ and $(\Jel, e)$ be finite
  pointed \interpretations. 
  If $e\in C^I_{\Lmc,n}(d)^\Jel$, with $n =
  \lvert \Delta^\Iel \rvert \cdot \lvert \Delta^\Jel \rvert$, then $(\Iel, d)\preceq_\Lmc (\Jel, e)$.
\end{restatable}

\noindent\begin{proof}\
  Towards proving the contrapositive, assume that
  $(\Iel, d) \not\preceq_\Lmc (\Jel, e)$.
  Then Lemma~\ref{lem:simtothemax} yields an \Lmc-concept
  $C$ of role depth at most $n$ such that $d \in C^I$
  and $e \notin C^J$. However, the former clearly implies
  $\emptyset \models C^I_{\Lmc,n}(d) \sqsubseteq C$ by
  definition of characteristic concepts, and thus 
   $e \notin C^J$ implies $e\notin C^I_{\Lmc,n}(d)^\Jel$,
   which is what we have to show.
\end{proof}

\thmmtcELfittingTGD*

\noindent\begin{proof}\
  ``$\Leftarrow$''. Assume that for all  $\bar d =(d_1,\dots,d_k)\in \Delta^{\prod \Nsf}$ with $d_i$ non-\Lmc-total in $N_i$ for all $i \in [k]$, Conditions~1
  and~2 are satisfied. Further assume to the contrary of what we have to show
  that there exists an \Lmc-concept inclusion $C \sqsubseteq D$ that fits $(\Psf, \Nsf)$.  Then for every $i \in [k]$
  there is a $d_i \in \Delta^{N_i}$ such that $d_i \in C^{N_i}\setminus D^{N_i}$. Consider the tuple $\bar d=(d_1,\dots,d_k)$. First note that, by Lemma~\ref{lem:prodELI}, $\bar d \in C^{\prod \Nsf}$.
  Also note that $d_i$ is not \Lmc-total in $N_i$ for all $i \in [k]$ and therefore $\bar d$ must satisfy
  Conditions~1 and~2. By Condition~1, the product $(\Jel, \bar f) =
  \prod S_{\bar{d}}$ is defined. Condition~2 yields $(\Jel, \bar f) \preceq_\Lmc (N_i,d_i)$ for some $i \in [k]$ and
  since $d_i \not\in D^{N_i}$, it follows that $\bar f\not\in D^\Jel$. By
  Lemma~\ref{lem:prodELI} this implies $e \not\in D^P$ for some $(P,e) \in
  S_{\bar{d}}$. But by definition of $S_{\bar{d}}$ we have $(\prod \Nsf,\bar d)\preceq_\Lmc (P,e)$ and thus from $\bar d \in C^{\prod \Nsf}$ we obtain
  $e\in C^P$. It follows that $P\not\models C\sqsubseteq D$, and since \Lmc is invariant under disjoint union, this contradicts the
  assumption that $C \sqsubseteq D$ fits $(\Psf,\Nsf)$.

  \smallskip

  \noindent ``$\Rightarrow$''. We show the contrapositive. Assume that 
  there is a non-\Lmc-total $\bar d \in \Delta^{\prod \Nsf}$ such that Conditions~1
  and~2 do not both hold. Let $\prod S_{\bar d} = (\Jel,\bar f)$ and $n =
  \lvert \Delta^{\prod \Nsf}\rvert \cdot \lvert \Delta^\Pel \rvert$.

  First assume that $d$ satisfies Condition~1 (and thus $\prod S_{\bar{d}}$ is defined), but violates Condition~2. Consider the concept inclusion
  $$C^{\prod \Nsf}_{\Lmc,n}(\bar d)  \sqsubseteq C^J_{\Lmc,m}(\bar f)$$
  where  $m = \lvert \Delta^\Jel \rvert \cdot \max_{\Iel\in\Nsf} |\Delta^\Iel|$.
  We show that every example in \Psf
  satisfies this inclusion while all examples in \Nsf violate it. For the former, it suffices
  to show that the disjoint union $P$ satisfies the inclusion. Let $e \in (C^{\prod \Nsf}_{\Lmc,n}(\bar{d}))^P$. Then
  Lemma~\ref{lem:FiniteUnravelling} yields
  $(\prod\Nsf, \bar d)\preceq_\Lmc (P, e)$  and thus
  $(P,e) \in S_{\bar{d}}$ is one of the \interpretations in the product $(\Jel,\bar f)$. By
  Lemma~\ref{lem:prodELI}, this implies $e \in (C^J_{\Lmc,m}(\bar f))^P$, as required.
  To show that every $N \in \Nsf$ violates ${C^{\prod \Nsf}_{\Lmc,n}(\bar d)} \sqsubseteq C^J_{\Lmc,m}(\bar f)$, first note that $\bar d \in (C^{\prod \Nsf}_{\Lmc,n}(\bar d))^{\prod \Nsf}$ and thus Lemma~\ref{lem:prodELI} yields $d_i \in (C^{\prod \Nsf}_{\Lmc,n}(\bar d))^{N_i}$ for all $i \in [k]$. Moreover, $(\Jel,\bar f) \not\preceq_\Lmc (N_i,d_i)$ for all $i \in [k]$ because
  Condition~2 is violated.
  From Lemma~\ref{lem:FiniteUnravelling} we obtain
  $d_i\not\in (C^J_{\Lmc,\ell}(\bar f))^{N_i}$ for $\ell=|\Delta^{N_i}| \cdot|\Delta^J|$. This clearly implies $d_i\not\in (C^J_{\Lmc,m}(\bar f))^{N_i}$ since $m \geq \ell$, and thus we are done.

  Now assume that Condition~1 is violated. Let $(K, t)$ be the pointed interpretation with $\Delta^K = \{t\}$
  and such that $t$ satisfies all concept names in the schema \Smc  of $(\Psf,\Nsf)$ and there
  is a reflexive loop on $t$ for every role name in \Smc.
  Then clearly
  $t$ is
  \Lmc-total. Consider the concept inclusion $$C^{\prod \Nsf}_{\Lmc,n}(\bar d) \sqsubseteq
  {C^K_{\Lmc,m}(t)}$$ where $m = \max_{\Iel\in\Nsf} |\Delta^\Iel|$. We again show that every example in \Psf satisfies this
  inclusion while no $N \in \Nsf$ does. Concerning the latter, it is again clear that $d_i \in (C^{\prod \Nsf}_{\Lmc,n}(\bar d))^{N_i}$ for all $i \in [k]$
  and thus  it suffices to show that $d_i\notin (C^K_{\Lmc,m}(t))^{N_i}$.
  Assume to the contrary. Then by Lemma~\ref{lem:FiniteUnravelling} $(\Kmc, t)\preceq_\Lmc
  (N_i, d_i)$. This implies that every concept satisfied by $(\Kmc, t)$ is also
  satisfied by $(N_i, d_i)$. As $t$ is \Lmc-total in $K$, $d_i$ must be \Lmc-total in $N_i$ too, a
  contradiction.
  To show that every example in \Psf satisfies
  $C^{\prod \Nsf}_{\Lmc,n}(\bar d) \sqsubseteq
  {C^K_{\Lmc,m}(t)}$,
  it suffices to show that the disjoint union $P$ satisfies the
  inclusion, and in particular that $(C^{\prod \Nsf}_{\Lmc,n}(\bar d))^P = \emptyset$. This, however, follows from
  Lemma~\ref{lem:FiniteUnravelling} and the fact that Condition~1 is violated.

  The proof of the claim for $\Lmc_\bot$ can be obtained from the proof for \Lmc above
  by minor changes. For the direction ``$\Leftarrow$'' and the case of
  violation of Condition~2 in the ``$\Rightarrow$'' direction, it suffices to
  lift the restriction to non-\Lmc-total values in the assumptions, which has no
  effect on the arguments. When it comes to the violation of Condition~1 in the
  ``$\Rightarrow$'' direction, the proof becomes simpler. The inclusion $C^{\prod \Nsf}_{\Lmc,n}(\bar d) \sqsubseteq
  {C^K_{\Lmc,m}(t)}$ can then be replaced by
  $C^{\prod \Nsf}_{\Lmc,n}(\bar d) \sqsubseteq
   \bot$.
\end{proof}

\thmELIFittingInExp*

\noindent\begin{proof}\
Let \Lmc be any of the DLs mentioned in the theorem.
To decide whether a given fitting instance $(\Psf,\Nsf)$ admits
a fitting \Lmc-TGD, we have to check whether there is a $\bar d =(d_1,\dots,d_k)\in \Delta^{\prod \Nsf}$, with $d_i$ non-\Lmc-total in $N_i$ for all $i \in [k]$ if \Lmc does not admit $\bot$, such that Conditions~1
and~2 of Theorem~\ref{thm:mtc:ELfittingTGD} are not both satisfied.
Making use of the facts that the existence of
a simulation can be decided in polynomial time and the involved products
are of single exponential size and can be computed
in output polynomial time, it is easy to implement the required checks
to obtain an \ExpTime upper bound.\footnote{Note that $d_i$ is \Lmc-total
in  $N_i$ iff $(I_\bot,d) \preceq_\EL (N_i,d_i)$ where $I_\bot$ is the 
instance that contains $A(d)$ for every concept name $A$ in the schema if
$(\Psf,\Nsf)$ and $R(d,d)$ for every role name $R$ in this schema. As a
consequence, totality can be decided in polynomial time.}  The fitting TGDs obtained if the 
check fails,
as  in the proof of Theorem~\ref{thm:mtc:ELfittingTGD}, are at
most double exponential in size and at most single exponential if
represented succinctly.

Regarding fitting ontology existence and construction, Lemma~\ref{lem:oneneg} yields a simple
reduction to the TGD fitting case that gives the desired results. 
We remark that, unlike the fitting ontologies obtained from
finite bases,
the resulting ontologies contain at most one concept inclusion per instance in~\Nsf.
\end{proof}

\thmelfittingexistence*

\noindent\begin{proof}\ 
  Let $\Lmc \in \{ \EL, \ELI, \EL_\bot, \ELI_\bot \}$. The
  \textit{product \EL-simulation problem} means to decide, given  pointed
  \interpretations $(\Iel_1,a_1),\ldots, (\Iel_n, a_n), (\Jel, b)$, whether
  \mbox{$\prod^n_{i=1}(\Iel_i, a_i) \preceq_\EL (\Jel, b)$}. This problem is
  known to be \ExpTime-hard~\cite{ijcai/FunkJLPW19}. We provide a
  polynomial time reduction from the product \EL-simulation problem to fitting
  \Lmc-ontology existence and fitting \Lmc-TGD existence. In fact, we shall use
  fitting instances with a single negative example for which, by
  Lemma~\ref{lem:oneneg}, the existence of fitting ontologies and of fitting
  TGDs coincides.
  
  Let the input to the product \EL-simulation problem consist of the pointed
  instances $(\Iel_1,a_1),\ldots, (\Iel_n, a_n), (\Jel, b)$ over some schema
  \Smc. We may assume without loss of generality that all these instances are
  pairwise disjoint.

  We construct a fitting instance $(\Psf,\Nsf)$ over an extended schema 
  $$\Smc' = \Smc \cup \{ A_c \mid c\in\Delta^\Jel \cup \{u,v\}\} \cup \{R \}$$
  where the $A_c$ are fresh concept names and $R$ is a fresh role name. We also
  introduce  fresh values $u,v$. Set $\Nsf = \{\Nel\}$ where
  $$
  \begin{array}{rcl}
  N &=& \Jel \cup \{A_c(c)\mid c\in
  \Delta^\Jel \cup \{u,v\}\} \cup \{R(u,b)\} \\[1mm]
   && \cup \, \{ S(v,c) \mid
  c \in \Delta^{J} \cup \{u,v\} \text{ and } S \in \Smc' \} \\[1mm]
  && \cup \, \{ A(v) \mid A \in \Smc'\}.
  \end{array}
  $$
  To define \Psf, let $N_1, \ldots, N_n$ be pairwise disjoint copies of~$N$. For
  every $d \in \Delta^{N}$ we use $\langle d,i \rangle$ to denote the copy of
  $d$ in~$N_i$. Moreover,  set $\langle d,i \rangle^\downarrow =d$. Now define
  $\Psf = \{\Pel\}$ with $\Pel$ the disjoint union of the instances
  $I_1,\dots,I_n$ and $N_1,\dots,N_n$, extended with the facts $R(\langle u,i
  \rangle, a_i)$, $i\in[n]$. 

  We  show that $\prod^n_{i=1}(\Iel_i, a_i) \preceq_\EL (\Jel, b)$ if and only
  if $(\Psf, \Nsf)$ has no fitting \Lmc-TGD, making use of the characterization
  provided by Theorem~\ref{thm:mtc:ELfittingTGD}. Note that because we only
  have a single negative example $N$, the product $\prod \Nsf$ in
  Theorem~\ref{thm:mtc:ELfittingTGD} is simply $N$, and the second condition
  simplifies to $\prod S_{\bar{d}} \preceq_\Lmc (N,d)$.
  Take any $d\in \Delta^\Nel$ and consider the set $S_d = \{(P,e) \mid (\Nel,
  d)\preceq_\EL (\Pel, e)\}$. We must have
  $$
  \begin{array}{rcl}
    S_d&=&\{(P,\langle d,1\rangle),\dots,(P,\langle d,n\rangle ) \} \, \cup \\[1mm]
    && \{(P,\langle v,1\rangle),\dots,(P,\langle v,n\rangle ) \}.
  \end{array}  
  $$ 
  In fact, the ``$\supseteq$'' direction is immediate and the ``$\subseteq$''
  follows from the use of the fresh concept names $A_{d}$. In particular,  $S_d$
  is non-empty and thus Condition~1 of Theorem~\ref{thm:mtc:ELfittingTGD} is
  satisfied for~$d$.

  As a consequence of this initial consideration and of
  Theorem~\ref{thm:mtc:ELfittingTGD}, it suffices to show the
  following.
  \\[2mm]
  {\bf Claim 1.} $\prod^n_{i=1}(\Iel_i, a_i)\preceq_\EL (\Jel, b)$ if and only
  if $\prod S_d \preceq_\Lmc (N,d)$ for every $d\in \Delta^{\Nel}$. \\[2mm]
  This should be clear in the case $\Lmc \in \{ \EL_\bot,\ELI_\bot\}$ where
  Conditions~1 and~2 of Theorem~\ref{thm:mtc:ELfittingTGD} have to be satisfied
  for all $d \in \Delta^N$, not just for the non-\Lmc-total ones.  Note that,
  due to the use of the fresh concept names $A_c$, none of the  values $d \in
  \Delta^N \setminus \{v\}$ is \Lmc-total while $v$ is \Lmc-total. But showing
  the claim also suffices in the case $\Lmc \in \{ \EL,\ELI\}$: for all
  \Lmc-total $d$ in $N$, we trivially have $\prod S_d \preceq_\Lmc (N,d)$ if
  $S_d$ is non-empty, which we have argued to be the case.

  \smallskip  

  To simplify matters, we consider the set
  $$
  T_d=\{(P,\langle d,1\rangle),\dots,(P,\langle d,n\rangle ) \}
  $$
  and observe the following.
  \\[2mm]
  {\bf Claim 2.} $\prod S_d \preceq_\Lmc (N,d)$ if and only if $\prod T_d
  \preceq_\Lmc (N,d)$ for every $d\in \Delta^{\Nel}$. \\[2mm]
  To prove Claim~2, it suffices to observe that from any \Lmc-simulation $Z$
  that witnesses  $\prod S_d \preceq_\Lmc (N,d)$, we can obtain an
  \Lmc-simulation $Z'$ that witnesses $\prod T_d \preceq_\Lmc (N,d)$ by using
  $$
    \{ ((b_1,\dots,b_n),c) \mid ((b_1,\dots,b_n,\langle v,1\rangle,\dots,\langle v,n\rangle),c) \in Z  \}.
  $$
  This is indeed an \Lmc-simulation because $Z$ is and each $\langle v,i\rangle$
   satisfies all concept names and has a reflexive loop for all role names.
   Conversely, from any \Lmc-simulation $Z$ that witnesses  $\prod T_d
   \preceq_\Lmc (N,d)$, we can obtain an \Lmc-simulation $Z'$ that witnesses
   $\prod S_d \preceq_\Lmc (N,d)$ by using
  $$
  \begin{array}{r@{}l}
    \{ ((b_1,\dots,b_n,c_1,\dots,c_n),c) \mid ((b_1,\dots,b_n),c) \in Z  
    \text{ and } \\[1mm]
    \hspace*{4cm}c_1,\dots,c_n \in \Delta^P\}.
  \end{array}
  $$

  \smallskip
    
  The virtue of Claim~2 is that it suffices to prove Claim~1 with $\prod T_d$ in
  place of $\prod S_d$. This is what we do in what follows. As a preparation,
  note that for distinct $d,d' \in \Delta^N$, the products $\prod T_d$ and
  $\prod T_{d'}$ are based on the same instance, which is the $n$-fold product
  of $P$ with itself, denoted $P^n$, and differ only in the distinguished tuple,
  which is $(\langle d,1\rangle,\dots,\langle d,n\rangle)$ and $(\langle
  d',1\rangle,\dots,\langle d',n\rangle)$, respectively. We continue to show
  Claim~1 with $\prod T_d$ in place of $\prod S_d$.
  \smallskip

  \noindent ``$\Rightarrow$''. Let $\prod^n_{i=1}(\Iel_i, a_i)\preceq_\EL (\Jel, b)$ and
  let $Z$ be a witnessing \EL-simulation. 
  We call a value $(d_1,\dots,d_n) \in \Delta^{P^n}$ \emph{well-sliced} if
  $d_i \in \Delta^{I_i} \cup \Delta^{N_i}$ for $1 \leq i \leq n$. Extend $Z$ to
  a relation $Z'$, as follows:
  \begin{enumerate}
  
  \item add, for every well-sliced $(d_1,\dots,d_n) \in \Delta^{P^n}$ that
  contains at least one copy of a value in $N$, the pair
  $$
    ((d_1,\dots,d_n),d_j^\downarrow)
  $$
  where $d_j$ is the left-most value in $(d_1,\dots,d_n)$ that is a copy of a
  value in $N$ (this choice is arbitrary);

  \item add, for every $(d_1,\dots,d_n) \in \Delta^{P^n}$, the pair
  $$
    ((d_1,\dots,d_n),v).
  $$
\end{enumerate}
  We show that $Z'$ is an \ELI-simulation from $\prod T_d$ to $(N,d)$, for every
  $d \in \Delta^N$. This suffices also when $\Lmc \in \{ \EL, \EL_\bot \}$ since
  every \ELI-simulation is also an \EL-simulation. By Point~1 of the extension,
  $Z'$ contains the pair $((\langle d,1\rangle,\dots,\langle d,n\rangle),d)$ for
  every $d \in \Delta^N$. It thus remains to show that $Z'$ is an
  \ELI-simulation from $P^n$ to $N$.
  
  Let $(\bar{e},e)\in Z \subseteq Z'$. We have to show that the two conditions
  of \ELI-simulations are satisfied. Condition~1 is satisfied since $Z$ is an
  \EL-simulation. For Condition~2, let $(\bar{e},e) \in Z$ and $(\bar e,\bar f)
  \in S^{P^n}$. If $S$ is a role name, then it suffices to observe that $Z$
  satisfies Condition~2 of \EL-simulations. If $S$ is an inverse role, it
  suffices to observe that $(e,v) \in S^N$ and $(\bar f,v) \in Z'$ by Point~2 of
  the extension.
    
  Now assume that $(\bar{e},e)\in Z' \setminus Z$ was added in Point~1 of the
  extension, with  $\bar e = (e_1,\dots,e_n)$. Then there is  some $e_j$ that is
  a copy of a value of $N$ and for the leftmost such  $e_j$ we have
  $e_j^\downarrow=e$. The two conditions of \ELI-simulations are again satisfied:
  \begin{enumerate}

    \item Assume that $\bar e \in A^{P^n}$ for some concept name $A$. Then $e
    \in A^N$, by definition of products and because $\bar e \in A^{P^n}$ and
    $e_j$ is a copy of $e$.

   \item Assume that $(\bar e,\bar f) \in S^{P^n}$, for some \ELI-role $S$. Let
   $\bar f = (f_1,\dots,f_n)$. We distinguish two cases.

    \emph{Case 1}. $S \notin \{ R,R^-\}$. Then $(\bar e,\bar f) \in S^{P^n}$
    implies that $f_\ell$ is a copy of a value in $N$ if and only if $e_\ell$
    is a copy of a value in $N$, for $1 \leq \ell \leq n$. In particular,
    this means that $f_j$ is the left-most value in $\bar f$ that is a copy of
    a value of~$N$. Point~1 of the definition of $Z'$ yields $(\bar f,f_j)
    \in Z'$. By definition of products, $(\bar e,\bar f) \in S^{P^n}$ implies
    that $(e_j^\downarrow,f_j^\downarrow) \in S^N$ and we are done.

    \emph{Case 2}. $S=R$.    
    First assume that $\bar f$ does not contain a copy of a value in $N$.
    Since $\bar e$ is well-sliced and $S=R$, we must then have $e_i = \langle u
   ,i \rangle$ and $f_i = a_i$ for $1 \leq i \leq n$. Note that $(u,b) \in
    R^N$. Moreover, $Z \subseteq Z'$ contains the pair $(\bar f,b)$ and thus we
    are done.

    Now assume that $\bar f$ contains a copy of a value in $N$. Let $f_k$ be
    the left-most such copy. It follows together with $(\bar e,\bar f) \in
    R^{P^n}$ that $e_k$ is also a copy of a value in $N$ and hence by
    definition of product we have $(e_k^\downarrow,f_k^\downarrow) \in R^N$.
    Moreover, Point~1 of the definition of $Z'$ yields $(\bar f,f_k^\downarrow)
    \in Z'$ as required.

    \emph{Case 3}. $S=R^-$. Recall $e_j^\downarrow=e$. By construction we have
    $(e_j^\downarrow, v) \in {R^-}^N$. From Point~2 of the definition of $Z'$
    we obtain $(\bar f,v) \in Z'$ and thus we are done.
  \end{enumerate}
  Finally assume that $(\bar{e},e)\in Z' \setminus Z$ was added in Point~2 of
  the extension. Then $e=v$. Recall that $v$ satisfies all concept names from
  $\Smc'$ and has a reflexive loop for every $\Smc'$-role name. Together with
  Point~2 of the definition of $Z'$, this clearly implies that Conditions~1
  and~2 of \ELI-simulations are satisfied.
    
  \smallskip

  \noindent
  ``$\Leftarrow$''. Assume that $\prod T_d \preceq_\Lmc (N,d)$ for every $d\in
  \Delta^{\Nel}$. Then in particular $\prod T_u \preceq_\Lmc (N,u)$. Let $Z$ be
  a witnessing simulation. Then specifically $(\bar{u}, u)\in Z$ where $\bar u =
  ( \langle u,1\rangle,\dots,\langle u,n \rangle)$. Define $Z' = Z \cap
  (\Delta^{\prod^n_{i=1}\Iel_i} \times \Delta^\Jel)$. We show that $Z'$ is an
  \EL-simulation that witnesses $\prod^n_{i=1}(\Iel_i, a_i)\preceq_\EL (\Jel,
  b)$.

  Note that $R(\bar{u}, \bar{a})\in \Pel^n$ where $\bar a = (a_1,\dots,a_n)$.
  Since $b$ is the only $R$-successor of $u$ in $\Nel$, it follows that
  $(\bar{a},b)\in Z$. Hence, $(\bar{a},b)\in Z'$. It remains to show that $Z'$
  is an \EL-simulation. This, however, follows from the fact that $Z$ is an
  \EL-simulation and that all values $\Delta^N \setminus \Delta^J$ that are
  reachable in $N$ from some value in $\Delta^J$ are reachable only via an
  inverse role, but not via a `forward' role.
\end{proof}

For the subsequent proof, we first recall a result from~\cite{Funk19}.

\begin{theorem}\label{thm:size-lower-el-concepts} Let $n\geq 1$. There are
  pointed instances $(I_1,d_1),\ldots,(I_n,d_n),(J,e)$ such that
  \begin{enumerate}
    \item the size of the instances is bounded by $p(n)$, $p$ a polynomial;
    \item $\prod_i (I_i,d_i)\not\preceq_\EL (J,e)$;
    \item the smallest \EL-concept $C$ such that $d_i\in C^{I_i}$ for all $i\in\{1,\ldots,n\}$ but $e\notin C^{J}$ has size $2^{2^n}$ and is of role depth $2^n$.\footnote{Reference~\cite{Funk19} does not explicitly mention a lower bound on the role depth, but it is not hard to derive it using the analysis given there.}
  \end{enumerate}
\end{theorem}

We use Theorem~\ref{thm:size-lower-el-concepts} to
prove the following result.

\thmsizelowerelfittings*

\noindent\begin{proof}\ 
    Let $\Lmc\in \{\EL,\ELI,\EL_\bot,\ELI_\bot\}$ and $n\geq 1$. Let $(I_1,d_1),\ldots,(I_n,d_n),(J,e)$ be the instances from Theorem~\ref{thm:size-lower-el-concepts}, and let their schema be \Smc. Without loss of generality, we can assume that the domains of these instances are disjoint. By Condition~2 of Theorem~\ref{thm:size-lower-el-concepts} and Lemma~\ref{lem:simulations}, there is an \EL-concept $C_0$ such that $d_i \in C_0^{I_i}$ for all $i\in\{1,\ldots,n\}$ but $e\notin C_0^{J}$.
    
  We construct sets of instances $\Psf_n=\{I_1',\ldots,I_n'\}$ and $\Nsf_n=\{J'\}$ over an extended schema 
  \[\Smc'=\Smc\cup\{A,S\}\]
  where $A$ is a fresh concept name and $S$ a fresh role name, 
  as follows:
  \begin{itemize}
  
    \item $J'$ is obtained from $J$ by adding two fresh values $e',f$ as well as 
    
    \begin{itemize}
    
        \item assertions $A(e')$ and $S(e',e)$ and

        \item assertions $R(f,d),B(f)$ for all role names $R\in\Smc'$, concept names $B\in \Smc'$, and values $d\in\mn{adom}(J)\cup\{f\}$.
        
    \end{itemize} 

    \item $I_i'$ is obtained from the union of $I_i$ and $J'$ by adding a fresh value $d_i'$ and the statements $S(d_i',d_i)$, $S(d_i',e)$, and $A(d_i')$.
  \end{itemize}
  Clearly, the sizes of $I_1',\ldots,I_n',J'$ are bounded by some fixed polynomial because the sizes of $I_1,\ldots,I_n,J$ are. It remains to verify Conditions~2 and~3.

  For Condition~2, observe that the TGD
  \[A\sqsubseteq \exists S.C_0\]
  is satisfied in every $I_i'$ but not in $J'$. Thus, 
  this TGD fits $(\Psf_n,\Nsf_n)$, and so does the
ontology that contains exactly this TGD.

Before verifying Condition~3, we observe two properties of the construction.

\smallskip\noindent\textbf{Claim.} For every $d\in \mn{adom}(J)\cup \{f\}$ and $i\in\{1,\ldots,n\}$, we have~(i) $(J',d)\preceq_\ELI (I'_i,d)$ and~(ii) $(I'_i,d)\preceq_\ELI (J',d)$.

\smallskip\noindent\textbf{Proof of the claim.} Point~(i) is clear since $J'\subseteq I_i$, for every~$i$, and hence the identity is a witnessing \ELI-simulation. For Point~(ii), observe that the relation $S_i$ defined by
\begin{align*}
S_i ={} & \{(d,d)\mid d\in \mn{adom}(J')\}\cup{}\\ & \{(d,f)\mid d\in \mn{adom}(I_i)\setminus\mn{adom}(J')\}
\end{align*}
is the witnessing \ELI-simulation, for every $i$. This finishes the proof of the claim.
    
\medskip

  To show Condition~3, let $X\sqsubseteq Y$ be a smallest TGD that fits $(\Psf_n,\Nsf_n)$. Then  $I_i'\models X\sqsubseteq Y$ for all $I_i'$, but $J'\not\models 
  X \sqsubseteq Y$.
  By the latter, we know that there is some $d\in \Delta^{J'}$ such that $d\in X^{J'}$ but $d\notin Y^{J'}$. 
  Suppose first that $d\in\mn{adom}(J)\cup\{f\}$. Then the Claim together with Lemma~\ref{lem:simulations} implies that $d$ satisfies the same $\ELI_\bot$-concepts in $J'$ and $I_i'$, contradicting $I_i'\models X\sqsubseteq Y$.
  Thus, $d=e'$ is the fresh value introduced in the construction of $J'$.
  Observe that by construction $(J',e')\preceq_\ELI (I'_i,d_i')$ for every $i\in\{1,\ldots,n\}$.
  Since $e'\in X^{J'}$, Lemma~\ref{lem:simulations} implies $d_i'\in X^{I'_i}$, for all $i$. Since all $I_i'$ satisfy $X\sqsubseteq Y$, also $d_i'\in Y^{I'_i}$. 
  
  By the structure of $I_i'$ and $J'$, $Y$
  is of shape $\exists S.Z$. Note that $e'\notin Y^{J'}$ implies $e\notin Z^{J'}$. By Point~(ii) of the claim and Lemma~\ref{lem:simulations}, it follows that $e\notin Z^{I_i'}$, for every $i$.
  Hence, $d_i\in Z^{I_i}$ for all $i$. 
  If $Z$ is an \EL-concept, then by the choice of $I_1,\ldots,I_n,J$ and Condition~3 of Theorem~\ref{thm:size-lower-el-concepts}, this $Z$ has to have size at least $2^{2^n}$ and depth $2^n$. Thus, the TGD $X\sqsubseteq Y$ has size at least $2^{2^n}$ and $2^n$ when represented succinctly.

  It remains to consider the case when $Z$ is an \ELI-concept. Let $U=\exists R^-.Z'$ be a subconcept of $Z$ and let $\bar Z$ be the concept obtained from $Z$ by replacing $\exists R^-.Z'$ with $\top$. Notice that we still have $e\notin \bar Z^{J'}$ since every value in $\mn{adom}(J)$ satisfies $U$ in $J'$, due to the presence of $f$. Moreover, $d_i\in \bar Z^{I_i'}$ for all $i$ and hence 
  $X\sqsubseteq \exists S.\bar Z$ is a smaller fitting TGD than $X \sqsubseteq Y$, in contradiction to $X \sqsubseteq Y$ being of smallest possible size.

  The argument for smallest fitting ontologies \Omc is almost identical. We start with choosing from \Omc some
  TGD  $X\sqsubseteq Y\in \Omc$ such that $I_i'\models X\sqsubseteq Y$ for all $I_i'$, but $J'\not\models X
  \sqsubseteq Y$ and then continue as above, in the
  end obtaining a smaller fitting ontology $\{ X\sqsubseteq \exists S.\bar Z \}$.
\end{proof}

\section{Proofs for Section~\ref{sect:finitebasesTGD}}

\lemFiniteBasesGTGDCorrect*
\noindent\begin{proof}\ 
We have to show that for every guarded TGD $\rho$:
  $$
  \Omc_\Hsf \models \rho \text{ if and only if } I\models \rho
  \text{ for all } I\in\Hsf.
  $$
``$\Rightarrow$''. It suffices to show that
  if $\rho \in \Omc_\Hsf$, then $I \models \rho$ for all $I \in \Hsf$. Let $\rho
  = \varphi(\bar{x},\bar{y}) \rightarrow \exists \bar{z} \, \psi (\bar{x},
  \bar{z})$. For brevity, let $q(\bar{x})=\exists \bar{y} \, \varphi$ and
  $p(\bar{x})=\exists \bar{z} \, \psi$. Consider any $I \in \Hsf$. If $q(I) =
  \emptyset$, then $I\models\rho$ holds trivially. Otherwise take any $\bar{a}
  \in q(I)$. We  have to show that $\bar{a}\in p(I)$. By construction of the
  rules in $\Omc_\Hsf$, the right-hand side $p$ of $\rho$ is obtained from the
  CQ $q_{(P^*,\bar{b}^*)}$ by renaming the free variables, where
  $$
  (P,\bar{b})=\prod S \quad \text{ and } \quad S = \{(J,\bar{c})\mid J\in \Hsf,
  \bar{c}\in q(J)\}.
  $$
  Clearly, $(I,\bar{a})$ is in $S$ and thus by
  Lemma~\ref{lem:product_homomorphism} there exists a homomorphism $h$ from
  $(P,\bar{b})$ to $(I,\bar{a})$. We can easily extend $h$ to a homomorphism
  $h'$ from $(P^*,\bar{b}^*)$ to $(I,\bar{a})$: if value $a^*$ is a clone of
  value $a$, then set $h'(a^*)=h(a)$. It is now simple to convert $h'$ into a
  homomorphism $g$ from $p$ to $I$ with $g(\bar{x})=\bar{a}$ by renaming the
  objects in the domain of $h'$. Thus $\bar{a}\in p(I)$ as required.

  \smallskip \noindent ``$\Leftarrow$''. Assume that $I \models \rho$ for all $I \in
  \Hsf$. Moreover, let $I$ be some instance with $I \models \Omc_\Hsf$. We have
  to show that $I \models \rho$. Let $\rho = \varphi(\bar{x},\bar{y})
  \rightarrow \exists \bar{z} \, \psi (\bar{x}, \bar{z})$ and, for brevity, let
  $q(\bar{x})=\exists \bar{y} \, \varphi$ and  $p(\bar{x})=\exists \bar{z} \,
  \psi$.  If $q(I) = \emptyset$, then $I\models\rho$ holds trivially. Thus
  assume that $q(I)\neq\emptyset$ and take any $\bar{a}\in q(I)$. We want to
  show that $\bar{a} \in p(I)$. Consider two cases.

  First, let $q(J)=\emptyset$ for all $J\in\Hsf$. Then $\Omc_\Hsf$ includes the
  rule $\varphi \rightarrow \psi'$ with $\psi'$ the conjunction of all atoms
  over \Smc that use only the variables from $\bar{x}$. Since $I \models
  \Omc_\Hsf$ and $\bar{a} \in q(I)$, it follows that $I$ contains all facts over
  \Smc that use only values from $\bar{a}$. But then we find a homomorphism $h$
  from $p$ to $I$ with
  $h(\bar{x})=\bar{a}$, no matter what $p$ is.
  Thus $\bar{a} \in p(I)$, as required.

  Second, assume that $q(J)\neq\emptyset$ for some $J\in\Hsf$. Then $\Omc_\Hsf$
  contains the rule $\varphi \rightarrow \exists \bar{z} \,
  \psi(\bar{x},\bar{z})$ where $\exists \bar{z} \, \psi(\bar{x},\bar{z})$ is
  obtained from the CQ $q_{(P^*,\bar{b}^*)}$ by renaming the free variables,
  where
  $$
  (P,\bar{b})=\prod S \quad \text{ and } \quad S = \{(J,\bar{c})\mid J\in\Hsf,
  \bar{c}\in q(J)\}.
  $$
  For brevity, let $p_\Pi(\bar{x})=\exists \bar{z} \, \psi(\bar{x},\bar{z})$.
  Since $I \models \Omc_\Hsf$, $\bar{a} \in p_\Pi(I)$ and thus there is a
  homomorphism $h$ from $p_\Pi$ to $I$ with $h(\bar{x})=\bar{a}$. 
  Since $J \models\rho$ for all $J \in \Hsf$, we
  have $\bar{c} \in p(J)$ for all $(J,\bar{c}) \in S$. By
  Lemma~\ref{lem:product_cq}, this implies $\bar{b} \in p(P)$, that is, there is
  a homomorphism $g$ from $p$ to $P$ with $g(\bar{x})=\bar{b}$. 
  Note that the variables in $\bar x=x_1 \cdots x_k$ are all distinct as $\bar x$ is simply
  the list of frontier variables of $\rho$. By definition of diversification,
  we obtain from $g$ a homomorphism $g'$ from $p$ to $P^*$ with $g'(\bar x)=\bar b^*$ by defining $g'(x_i)$ to be the fresh value $a^*_i$ introduced by diversification as
  a clone of the $i$-th value in $\bar b$.  
  The composition
  $h \circ g'$ is a homomorphism from $p$ to $I$ and satisfies $h \circ
  g'(\bar{x})=\bar{a}$. Consequently, $\bar{a} \in p(I)$, as required.
\end{proof}

\thmsmallgtgdbasis*

\noindent\begin{proof}\
    We start with analyzing the size. Due to guardedness, we can assume w.l.o.g.\ that only variables $x_1,\ldots,x_\ell$ are used in TGD bodies in $\Omc_\Hsf$. Then the number of possible atoms is
    \[\sum_{P\in\Smc} \ell^{\mn{ar}(P)}\leq s\cdot \ell^\ell.\]
    The number of possible TGD bodies built from at most $n=||H||+1$ of the possible atoms is
    \[\binom{s\cdot \ell^\ell}{n} \in O\left( (s\cdot \ell^\ell)^{n}\right),\]
    as required.
    
    To show that $\Omc_\Hsf'$ is indeed a basis, it suffices to verify that $\Omc_\Hsf'\models\Omc_\Hsf$. Let $q(\bar x)\to p(\bar x)\in \Omc_\Hsf$. 
    We claim that there is a TGD $q'(\bar x)\to p(\bar x)\in \Omc_\Hsf'$ such that all atoms in $q'$ are atoms in $q$.  Clearly, this implies $q'(\bar x)\to p(\bar x)\models q(\bar x)\to p(\bar x)$, which closes the proof. 

    Let $q(\bar x)=\exists \bar{y} \,\varphi(\bar x,\bar y)$ and let $R(\bar z)$ be the guard atom in $\varphi(\bar x,\bar y)$  where $\bar z=\bar x\cup \bar y$. For simplicity, suppose that the guard atom takes the shape $R(z_1,\ldots,z_n)$ and that $\bar x=z_1,\ldots,z_k$ and $\bar y=z_{k+1},\ldots,z_n$ for some $k$. The other cases can be treated similarly.

    We make two observations on $q(I)$ for each $I\in \Hsf$: 
    \begin{enumerate}
        \item $q(I)\subseteq \{ (a_1,\ldots,a_k)\mid R(a_1,\ldots,a_n)\in I\}$, and 
       \item for every $R(\bar b)\in I$, $\bar b = (b_1,\ldots,b_n)$ such that $(b_1,\ldots,b_k)\notin q(I)$, we find an atom $\alpha_{I,\bar b}=P(z_{i_1},\ldots,z_{i_m})$ in $q$ such that $P(b_{i_1},\ldots,b_{i_m})\notin I$. 
    \end{enumerate}
    Let $q'=\exists \bar{y} \,\varphi'(\bar x,\bar y)$ where $\varphi'$ is the conjunction of $R(z_1,\ldots,z_n)$ and all atoms $\alpha_{I,\bar b}$ identified in Item~2 above. By definition, all atoms in $q'$ are atoms in $q$, as intended. By construction of $q'$, we additionally have \[\{(I,\bar a)\mid I\in\Hsf,\bar a\in q(I)\}=\{(I,\bar a)\mid I\in\Hsf,\bar a\in q'(I)\}.\]
    Moreover, $q'$ has at most $||\Hsf||+1$ atoms. Hence, $q'(x)\to p(x)\in \Omc_\Hsf'$ as claimed.
\end{proof}

The proof of Theorem~\ref{thm:size-lower-gtgd-ontologies-improved} relies on a standard characterization of entailments of TGD-ontologies via the chase, which we briefly recall next. Let $J$ be an instance, $\rho=p(\bar x)\to q(\bar x)$ be a TGD, and $\bar a$ a tuple over $\mn{adom}(J)$. We say that $\rho$ is \emph{applicable to} $J,\bar a$ if there is a homomorphism from $p(\bar x)$ to $(J,\bar a)$. The \emph{result of applying $\rho$ to $J,\bar a$} is obtained as the union of $J$ and $I_q[\bar x/\bar a]$, which is the canonical instance $I_q$ of $q$ with $\bar x$ renamed to $\bar a$. A sequence of instances $J_0,J_1,\ldots$ is called a \emph{chase sequence for $J$ and $\Omc$} if $J=J_0$ and each $J_{i+1}$ is the result of applying some $\rho\in \Omc$ to $J_i,\bar a$ for some $\bar a$. The sequence $J_0,J_1,\ldots$ is \emph{fair} if for each $i\geq 0$ and each tuple $\bar a$ over $\mn{adom}(J_i)$ such that some $\rho\in\Omc$ is applicable to $J_i,\bar a$, this $\rho$ is applied to $J_j,\bar a$ for some $j$. The \emph{result} of the chase sequence $J_0,J_1,\ldots$ is the instance $\bigcup_{i\geq 0}J_i$. The following characterization is well known, see for example~\cite[Proposition~3]{pods/DeutschNR08}.

\begin{lemma}\label{lem:chase}
Let $\Omc$ be some TGD-ontology and $\rho=p(\bar x)\to q(\bar x)$ some TGD. Then
\[\Omc\models \rho\quad \text{iff}\quad \widehat J,\bar a\models q(\bar a)\]
for some (equivalently: all) result $\widehat J$ of a fair chase sequence for $I_p$ and $\Omc$.
\end{lemma}

\thmsizelowergtgdontologiesimproved*

\noindent\begin{proof}\
Let $P$ and $R$ be unary and binary symbols, respectively.
For $m\geq 1$, let $L_m$ denote the ``lasso'' instance, with values 
$a_0^m, \ldots, a_{2m-1}^m$
and facts
\begin{align*}
L_m ={} & \{R(a_i^m, a_{i+1}^m)\mid i<2m-1\}\cup {} \\ & \{R(a_{2m-1}^m,a_m^m), P(a_m^m)\}
\end{align*}
We now define, for $n\geq 1$, $I_n$ as 
\[I_n=\bigcup_{i=1}^n \left(L_{p_i}\cup \{A(a_0^{p_i})\}\right),\]
where $p_i$ denotes the $i$-th prime number (that is, $p_1=2, p_2=3, \ldots$).

We claim that the constructed instances are as required. For Point~1 observe that, 
by the prime number theorem, $p_n = O(n\log n)$, and thus the size of $I_n$ is bounded by some polynomial in $n$. 

For Point~2, we rely on the following fact from~\cite{ipl/CateFJL24}:
\begin{itemize}
    \item [$(\dagger)$] Any unary CQ $q(x)$ which contains a $P$-atom connected to $x$ and such that the TGD $A(x)\to q(x)$ is satisfied in $I_n$ of size at least $2^n$.
\end{itemize}
Here, \emph{connected} is defined as usual in terms of the undirected graph $(\mn{var}(q), \{ \{x,y\} \mid x,y\text{ co-occur in some atom in $q$}\})$ induced by $q$. The fact~$(\dagger)$ can be proved along the lines of the proof of Theorem~3.2 in~\cite{ipl/CateFJL24}. Since it is rather close to that proof we omit the details. The intuition is that $P$-atoms occur only in prime distance to $A$ atoms, so the assumption that $A(x)\to q(x)$ is satisfied in $I_n$ means that the distance of any $P$-atom from $x$ is the product of all the used primes which is at least $2 ^n$.

We proceed with the proof of Point~2. Let $\Omc$ be a smallest GTGD-basis for $I_n$, and consider the TGD $\rho_n = A(x)\to q_1(x_1)$ with \[q_1(x_1)=\exists x_2\ldots x_k.R(x_1,x_2)\wedge\ldots R(x_{k-1},x_k)\wedge P(x_k)\] where $k=\Pi_{i=1}^n p_i$. Since $I_n\models \rho_n$, we must have $\Omc\models \rho_n$. By Lemma~\ref{lem:chase}, we have $\widehat J\models q_1(a)$ for the result $\widehat J$ of some fair chase sequence for $\{A(a)\}$ and \Omc. Since \Omc is a GTGD-ontology, the left-hand sides of TGDs in \Omc have a very restricted shape. Essentially, they can be only one of the following CQs:
\begin{align}
& A(x)\qquad R(x,y)\qquad A(x),R(x,y)\label{eq:possible-lhs1}\\
& P(x)\qquad R(x,y),P(y) \qquad P(x), R(x,y)\label{eq:possible-lhs2}
\end{align}
(There are more guarded CQs but those will have no match into $\widehat J$, e.g. $A(x),P(x)$.)

For the analysis, let $J_0,J_1,\ldots$ be the fair chase sequence that leads to $\widehat J$ and let $i$ be the first index such that $J_i$ contains some $P$-fact connected to the value $a$ in the initial instance~$J_0$. Since we can consider any fair chase sequence, it is without loss of generality to assume that before that step $i$ only rules with rule heads of shape~\eqref{eq:possible-lhs1} have been applied. We distinguish now cases on the head of the rule applied in the last step:
\begin{itemize}
    \item Consider first rule head $A(x)$, that is, a rule of shape 
    \[\rho = A(x)\to q(x)\]
    for some CQ $q(x)$. By assumption, $q$ contains some atom $P(y)$ with $y$ connected to $x$. Since $\rho\in\Omc$ and $\Omc$ is a GTGD-basis for $I_n$, we have $I_n\models \rho$. Fact~$(\dagger)$ implies that $q(x)$ and thus $\Omc$ is of size at least $2^n$.
    
    \item Consider next rule head $R(x,y)$, that is, a rule of shape 
    \[\rho = R(x,y)\to q(x,y)\]
    for some CQ $q(x,y)$. As in the previous case, we can conclude that $q(x,y)$ contains an atom $P(z)$ for some $z$ connected to $x$. Let $q'(x,y)$ be connected component of $x$ in $q(x,y)$.
    Since $\rho\in\Omc$, we know that there is a homomorphism from $q'(x,y)$ to $(I_n,(b_1,b_2))$ for every atom $R(b_1,b_2)\in I_n$. A straightforward analysis shows that $q'$ cannot contain a $P$ atom, a contradiction. (Informally, for each distance $\ell$ we can find an atom $R(b_1,b_2)$ so that there is no $P$-atom in distance $\ell$ from $b_1$ in $I_n$.)
    
    \item The case of rule head $A(x),R(x,y)$ can be treated as the first case. 
    
    \end{itemize}
    This finishes the proof of Theorem~\ref{thm:size-lower-gtgd-ontologies-improved}.
\end{proof}

\thmsizelowerfulltgdbases*

\noindent\begin{proof}\
  The instance $I_n$ contains, for $1 \leq i \leq n$, the facts
  $$
    S(\bar a_i) \quad R(\bar b_i) \quad R(\bar c_i)
  $$
  with $\bar a_i$ a tuple of $2n$ pairwise distinct values, $\bar b_i$~identical to $\bar a_i$ except that  a fresh
  value is in position $2i-1$, and $\bar c_i$ 
  identical to $\bar a_i$ except that  
  a fresh value  is in position~$2i$.
  Then among the {\IND}s true in $I_i$ we
  find all $2^n$ {\IND}s of the form
  $$
    S(x_1,y_1,\dots,x_n,y_n)
      \rightarrow \exists z_1 \cdots \exists z_n \,
    R(u_1,v_1,\dots,u_n,v_n)
  $$
  where for $1 \leq i \leq n$, either
  $(u_i,v_i)=(x_i,z_i)$ or
  $(u_i,v_i)=(z_i,y_i)$. 
   Moreover, it is not difficult
   to see that (i)~each of these {\IND}s $\rho$ is logically strongest among the {\IND}s in
   $I_n$, that is, there is no \IND $\rho'$ 
   true in $I_n$ such that $\rho' \models \rho$;
   and (ii)~for
   a set of {\IND}s $\Omc$ and an \IND $\rho$, $\Omc \models \rho$ 
   if and only if there is a $\rho' \in \Omc$
   such that $\rho' \models \rho$. As a 
   consequence, any finite \IND-basis of 
   $I_n$ must contain all of the $2^n$ {\IND}s above.
\end{proof}
%

\begin{restatable}{lemma}{lemnofinitebasis}
\label{lem:nofinitebasis}
  The instance $I$ has no finite \FGTGD-basis and no
  finite \FONETGD-basis.
\end{restatable}

\noindent\begin{proof}\ 
    Assume to the contrary that $\Omc$ is a finite \FGTGD-basis or \FONETGD-basis of $I$, and let $n$ be the maximal number of variables contained in the body of
    any TGD in $\Omc$. Set $m = 2n+1$ and let $J$ denote the instance that is a bi-directional $R$-cycle of length $m$, that is, $R(d_i, d_{i+1}) \in J$
    and $R(d_{i+1}, d_i) \in J$  for all $i \in [m]$ (indices taken modulo $m$). 

  Consider $\rho_m$. Since $m$ is odd and $I$ satisfies $\rho_k$  for odd $k$,
  $I\models\rho_m$. Therefore, $\Omc \models \rho_m$. Regarding $J$, by construction every $d_i$ lies on a
  cycle of length $m$ but $R(d_i,d_i)\notin J$, so $J\not\models\rho_m$. We
  proceed by showing that $J\models \Omc$, which implies $\Omc \not\models\rho_m$,
  yielding the desired contradiction.

  Take any TGD $\rho\in\Omc$ and let $\rho = \varphi(\bar{x},\bar{y}) \rightarrow \exists
  \bar{z} \, \psi (\bar{x}, \bar{z})$. For brevity, define $q(\bar{x})=\exists \bar{y} \,
  \varphi$ and $p(\bar{x})=\exists \bar{z} \, \psi$. 
  Let $h_{q\to J}$ be a homomorphism from
  $q$ to $J$. Define $J'$ as the maximal subinstance of $J$ whose active domain is the image of $h_{q\to J}$.
  Clearly, $J'$ has at most $n$
  values. This allows us to define a
  homomorphism $h_{J' \to I}$ from $J'$ to $I$: the elements of  any maximal contiguous sequence $d_i, \ldots, d_{i+k} \in \mn{adom}(J')$,
   are mapped alternately to $a$ and $b$. 

  The composition $h_{q \to I} = h_{J' \to I} \circ h_{q \to J}$ defines a
  homomorphism from $q$ to $I$. Since $I$ satisfies
  $\rho$, there is thus also a homomorphism $h_{p \to I}$ from $p$ to $I$ with $h_{p \to
      I}(\bar{x}) = h_{q \to I}(\bar{x})$.
            
  Next observe that the restriction $g$ of $h_{J' \to I}$
  to the set of elements $S=\{ h_{q \to J}(x) \mid x \text{ frontier-variable in } \rho \}$ is injective. This is 
  trivially the case if $\rho$ contains at most one 
  frontier variable. It is also the case if $\rho$ 
  contains two frontier variables since $\rho$ is frontier-guarded and $I$ contains no reflexive $R$-loop. Therefore,
  the inverse $g^-$ of $g$ is a well-defined mapping
  from $\mn{adom}(I)$ to $\mn{adom}(J)$. 

  Moreover, it is easy to verify that $g^-$ is a homomorphism from $I$ to $J$ because the cycle
  that we had added in the construction of $J$ is 
  bi-directional and thus for every edge $R(d,e) \in J'$
  there is also the edge $R(e,d)$. It follows that $h_{p \to J} = h_{p \to I} \circ g^-$
  is a homomorphism from $p$ to $J$ with $h_{p \to J}(\bar{x}) = h_{q \to
      J}(\bar{x})$, and therefore $J\models\rho$.
  We conclude $J\models \Omc$, as required.
\end{proof}

$$
\begin{array}{rl}
  \Omc_I = \{ & R(x,y) \rightarrow R(y,x) \\[1mm]
  & R(x,y) \wedge R(y,z) \wedge R(z,u)
  \rightarrow R(x,u) \\[1mm]
  & E(x,x) \wedge \mn{true}(x) \wedge \mn{true}(y) \rightarrow R(y,z) \ \}
\end{array}
$$

\lemfinitefullbasisexists*
\noindent\begin{proof}\
  Let $J$ be a model of $\Omc_I$ and
    take any full TGD  $\rho = q \rightarrow R(x_1,x'_1) \wedge \cdots \wedge R(x_k,x'_k)$  with $I \models \rho$.  Let $h$ be a homomorphism from
  $q$ to $J$. We have to show that 
  $R(h(x_i),h(x'_i)) \in J$ for $1 \leq i \leq k$.

  First assume that $q$ does not admit a homomorphism to~$I$. By choice of $I$,  the canonical instance $I_q$ of $q$ then contains a cycle of odd length, and so does $J$. But since the first and second TGD in $\Omc_I$ are satisfied in $J$,
this implies that $J$ must contain a reflexive loop. By the third TGD, $J$ is total, that is, it contains $R(a,b)$ for all $a,b \in \mn{ind}(J)$. Therefore, we
  clearly have  $R(h(x_i),h(x'_i)) \in J$ for $1 \leq i \leq k$, as required.

  Next assume that $q$  admits a 
  homomorphism $g$ to~$I$. Then the canonical instance $I_q$ of $q$ is bipartite with partition $P_1,P_2$ where $P_1$ contains all variables  $x \in \mn{var}(q)$ with $g(x)=a$, and likewise
  for $P_2$ and $g(x)=b$. 

 Take any head atom $R(x_i,x'_i)$.
  Then $x'_i$ are neither both in $P_1$
  nor  in $P_2$ because otherwise
  $g$ would witness that $I$ does not satisfy $\rho$. What is more, 
  $x_i$ and $x'_i$ must belong to the same maximally connected component of $I_q$.
  For if they belonged to different connected components, we could find a homomorphism
  from $q$ to $I$ that maps both $x_i$ and $x'_i$ 
  to the same value $a$ or $b$, but $I$ has
  no reflexive loops which contradicts satisfaction of $\rho$ in $I$. 

  Let $Q_1$ be the set of all values $h(x) \in \mn{adom}(J)$
  such that $x \in P_1$ and $x_1$ is in the same maximally connected component as $x_i$ and  $x'_i$, and likewise for $Q_2$. Since $J$ satisfies the
  first and second TGD in $\Omc_I$, there must be 
  an edge $R(c_1,c_2) \in J$ for all $c_1 \in Q_1$ and $c_2 \in Q_2$; note that this
  relies on our careful use of maximally 
  connected components. But since $x_i$
  and $x'_i$ are neither both in $P_1$
  nor  in $P_2$, this implies $R(h(x_i),h(x'_i)) \in J$, as required.
\end{proof}

\thmftgdnobasis*
\noindent\begin{proof}\ 
    Assume to the contrary that $\Omc$ is a finite \FTGD-basis of $J$, and let $n$ be the maximal number of variables contained in the body of
    any TGD in $\Omc$. 

Take any graph
    $G$ that is not 3-colorable and has girth exceeding $n$ and view it as an
    instance, that is, make it  bidirectional.
  Consider the TGD $\rho_G$. Since $G$ is not 3-colorable and $J$ satisfies $\rho_G$  for non-3-colorable $G$, we have
  $J\models\rho_G$. Therefore, $\Omc \models \rho_G$. We
  proceed to show that $G\models \Omc$, which implies $\Omc \not\models\rho_G$
since clearly $G \not \models \rho_G$,
  yielding the desired contradiction.

  Take any TGD $\rho\in\Omc$, with $\rho = q \rightarrow R(x_1,x'_1) \wedge \cdots \wedge R(x_k,x'_k)$. 
   Let $h$ be a homomorphism from
  $q$ to $G$. Define $G'$ as the  subinstance/graph of $G$ induced by
  the values/vertices in the domain of $h$.
  Clearly, $G'$ has at most $n$
  values. Since the girth of $G$ exceeds $n$, $G'$ is acyclic and
  thus 3-colorable. Moreover, for any two
  vertices $u,v$ in $G'$, we find a 3-coloring of
  $G'$ that colors $u$ and $v$ with different colors. Choosing $u=h(x_i)$ and $v=h(x'_i)$ for some $i \in [n]$, this
  implies that there is a
  homomorphism $g$ from $G'$ to $J$ such
  that $R(g(h(x_i)),g(h(x'_i))) \notin J$. 
    The composition $g\circ h$ is a
  homomorphism from $q$ to~$J$. Since $J$ satisfies
  $\rho$, we must thus have $R(g \circ h(x_i),g \circ h(x'_i)) \in J$, a contradiction.
\end{proof}

\section{Proofs for Section~\ref{sect:fittingTGD}}

\thmfittingexindcomplexity*

\noindent\begin{proof}\
  \emph{3SAT} is the problem to decide the satisfiability of propositional logic formulas  that are in conjunctive
  normal form with exactly three distinct literals per clause. It is famously known to be \NPclass-hard.
  Without loss of generality, we additionally require that no clause contains both a variable and its negation. We provide a polynomial time reduction from \emph{3SAT} to
  fitting \IND-ontology existence and fitting \IND existence. The fitting
  instances used in the reduction contain a single negative example and thus,
  by Lemma~\ref{lem:oneneg}, the existence of a fitting ontology and a fitting
  \IND coincide.

  Let the input to \emph{3SAT} be the propositional formula
   $\varphi = C_1 \wedge \cdots \wedge C_m$ using the variables $p_1, \dots, p_n$.
  We construct a fitting instance $(\Psf = \{I_1, \ldots, I_m\}, \Nsf = \{J\})$
  over the schema $\{R,S\}$, where $\mn{ar}(R) = \mn{ar}(S) = 2n$. With every
  position $i\in [2n]$, we associate the literal $p_i$, if $i\leq n$, and $\neg
  p_{i-n}$, if $i > n$. For brevity, we speak about \emph{position $p_i$}
  to mean position $i$ and \emph{position $\neg p_i$} to mean
  position $n+i$.
  
  Fix a tuple $\bar{a} = (a_1, \ldots, a_{2n})$  of values. 
  We will consider tuples of values obtained from $\bar a$ by replacing
  some values with fresh values. 
Each clause $C_i$ gives rise to a set  $M_i$ of exactly 8 such tuples, obtained from $\bar a$ by
  replacing, for $\ell_1,\ell_2,\ell_3$ the literals in $C_i$,
   either the value in position $\ell_i$ or the
  value in position $\neg \ell_i$  with a fresh value, for all $i \in \{1,2,3\}$. The tuple in $M_i$ that has fresh values in all three positions
  $\ell_1, \ell_2, \ell_3$ is referred to as the \emph{falsifying tuple}.
  We now define the instances $I_i$ and $J$ as follows:
  \begin{itemize}
      \item $I_i$ contains $R(\bar a)$ and all facts $S(\bar b)$ with $\bar b \in M_i$ non-falsifying;
            \item $J$ contains $R(\bar a)$ and, for $1 \leq i \leq m$, the fact $S(\bar b)$ where $\bar b \in M_i$ is falsifying.

  \end{itemize}
  We need to show that $\varphi$ is satisfiable if and only if $(\Psf, \Nsf)$
  has a fitting \IND. 
  Whenever we consider tuples of variables
  $\bar x$ of length $2n$ and a literal $\ell$, then we use $x_\ell$ to denote the variable in $\bar x$ in position $\ell$, and likewise for $\bar y$ etc.

  \medskip
  \noindent ``$\Rightarrow$'' Assume $\varphi$ is satisfiable and let $V$
  be a satisfying variable assignment. Consider the \IND
  $$\rho = R(\bar x) \to \exists \bar{z} \, S(\bar y),$$ where 
  $\bar x$, $\bar y$, $\bar z$ are tuples of variables of
  length $2n$. The variables in $\bar x$ and $\bar z$ are all
  distinct. If $V(p)=1$, then the variable in position
  $p$ of $\bar y$ is $x_p$ and the variable in position $\neg p$ is $z_{\neg p}$.
  If $V(p)=0$, then the variable in position
  $p$  of $\bar y$ is $z_p$ and the variable in position $\neg p$ is $x_{\neg p}$.
  We show that $\rho$ fits $(\Psf,
  \Nsf)$, starting with the positive examples. 
  
  Let $I_i\in\Psf$ and let $C_i =
  \ell_1 \vee \ell_2 \vee \ell_3$ be the corresponding clause. There is exactly one homomorphism $h$ from
  the body of $\rho$ to $I_i$, mapping the atom $R(\bar x)$ to the fact $R(\bar a)$. Consider
  the fact $S(\bar b)$ where $\bar b$ is obtained from $\bar a$ by replacing, for all $i \in \{1,2,3\}$, the value in position $\neg \ell_i$ by a fresh value if 
  $V \models \ell_i$ and the value in position  $\ell_i$ by a fresh value if $V\not\models \ell_i$. Since $V$ satisfies
  $C_i$, the tuple $\bar b$ is not the falsifying
  tuple. By construction, it follows that $I_i$ contains $S(\bar b)$. Using the construction of $\rho$, it is now easy to verify that the
  mapping $g$ defined by $g(\bar z)=\bar b$ is a
  homomorphism from the head of $\rho$ to $I_i$ that maps the atom $S(\bar y)$ to $S(\bar b)$. This homomorphism witnesses satisfaction of the IND.
  
Now consider the negative example. Again,  there is exactly one homomorphism $h$ from the body of $\rho$ to $J$, mapping the atom $R(\bar x)$ to the fact $R(\bar a)$. Consider any fact $S(\bar b)$ in $J$. We show that $S(\bar b)$ does witness satisfaction of the
IND. By construction of $J$, $S(\bar b)$ is a falsifying tuple, say for clause
$C_i=\ell_1 \vee \ell_2 \vee \ell_3$. We have $V(\ell_i)=1$ for
some $i \in \{1,2,3\}$. But then the tuple $\bar y$ has variable
$x_{\ell_i}$ in position $\ell_i$ whereas the tuple $\bar b$ has a
fresh value (rather than the value from position $\ell_i$ of
$\bar a$) in position $\ell_i$. Thus, $S(\bar y)$ cannot be mapped
to $S(\bar b)$ by a homomorphism that agrees with $h$ on the variables
that occur both in $\bar y$ and $\bar x$.

    \medskip
  \noindent ``$\Leftarrow$'' Assume that $(\Psf, \Nsf)$ has a fitting \IND $\rho$. We first prove that $\rho$ satisfies
  the following properties:
  \begin{enumerate}
    \item there are no repeated variables in the body atom, and neither in the head atom;
    \item every frontier variable of $\rho$ has the same position in the head atom as
    it has in the body atom;
    \item the relation symbol in the body of $\rho$ is $R$, the relation symbol in the head  is $S$;
    \item for every variable $p$, not more than one of the variables
    in positions $p$ and $\neg p$ in the body atom occurs in the frontier
    of $\rho$.
  \end{enumerate}
  \emph{Property~1}.
  It is easy to verify that for every instance $I$ in $(\Psf, \Nsf)$ and every atom $\alpha$ over schema $\{R, S\}$ that contains variables $\bar x$, we have $I \models \exists \bar{x} \, \alpha$ if and only if $\alpha$ does not
  contain any variable more than once. Thus, if an atom with repeated variables occurs in the body
  of~$\rho$, then $J\models \rho$; if no such atom occurs in the body,
  but some such atom  occurs in the head, then $I\not\models \rho$ for all $I\in\Psf$. Either case contradicts
  that $\rho$ is a fitting, hence $\rho$ satisfies Property~1. 

  \emph{Property~2}. Since Property~1 is satisfied, the body atom of $\rho$ contains every variable at most once. Every such atom admits a homomorphism to any positive example $I_i$. But in $I_i$, 
  every value
  occurs only in exactly one position across all facts. Since $I_i\models \rho$, this implies that Property~2 must be satisfied.
  
  \emph{Property~3}. We first observe that
  $\rho$ must use both $R$ and $S$ as, otherwise,  Properties~1 and~2 imply that $\rho$ is a tautology, in contradiction to $J \not\models \rho$. The remaining case to be ruled out is thus that 
   $\rho$ uses $S$ in the body and $R$ in the head. The frontier
   must contain some variable as otherwise $J \models \rho$.
   Let $i$ be the position in which $x$ occurs in the head atom $R(\bar y)$.
   By construction, there is a positive example $I_j$
   that contains a fact $S(\bar b)$ with a fresh value in
   position $i$. But the only $R$-fact in $I_j$ is $R(\bar a)$,
   which contains a non-fresh (and thus distinct) value in position $i$.
   It follows that $I_j \not\models \rho$, a contradiction.

  \emph{Property~4}. Let $C_i$ be a clause in which variable $p$ occurs
  and consider the positive example $I_i$. There is only one 
  homomorphism from the body $R(\bar x)$ of $\rho$ to $I_i$,
  mapping $R(\bar x)$ to $R(\bar a)$. By construction of $I_i$,
  every $S$-fact in $I_i$ has a fresh value on position $p$ or $\neg p$.
  Consequently, if the head atom $S(\bar y)$ in $\rho$ contains a frontier
  variable in both position $p$ and $\neg p$, then $I_i \not \models \rho$, a contradiction.

\smallskip
By Properties~1--3, we may assume that 
 $$\rho = R(\bar x) \to \exists \bar{z} \, S(\bar y)$$
where $\bar x$ and $\bar y$ contain no repeated variables
and if some variable $x$ from $\bar x$ occurs in $\bar y$,
then this is in the same place as the occurrence of $x$ in $\bar x$.
 
 We construct an assignment $V$ that~$\varphi$:
 set $V(p)=1$ if the variable in position $p$ of $\bar y$ is from
 $\bar x$
 and $V(p)=0$ if it is from $\bar z$. We have to argue the
 $V$ satisfies all clauses in $\varphi$. 
 
 Take any such clause
 $C$. Then $J$ contains the 
 fact $S(\bar b)$ where $\bar b$ is the falsifying tuple for
 $C$. Since $\rho$ fits the negative example, the homomorphism
 $h$ that maps $R(\bar x)$ to $R(\bar a$) cannot be extended to 
 a homomorphism that maps $S(\bar y)$ to $S(\bar b)$. This can only
 be the case if there is a position $i$ in $\bar y$ that contain a 
 variable from $\bar x$, but such that position $i$ of $\bar b$
 contain a fresh value (rather than a value from $\bar a$). First
 assume that position $i$ is associated with a variable $p$. Since
 $\bar y$ contains a variable from $\bar x$ in position $p$, we 
 have $V(p)=1$. Since position $p$ of $\bar b$ contains a fresh
 value and $b$ is the falsifying tuple, $p$ is one of the literals 
in $C$. The case that position $i$ is associated
 with a negated variable is similar, additionally using Property~4 above.
\end{proof}

\thmmtcGTGDfittingTGD*

\noindent\begin{proof}\
  ``$\Rightarrow$''. We show the contrapositive. Assume that there is some
  non-empty maximally guarded $M \subseteq \mn{adom}(\prod \Nsf)$ such that
  $\bar{M}[i]$ is non-total in $N_i$ for all $i\in [k]$ and at least one of the
  Conditions~1 and~2 is violated. We have to show that $(\Psf, \Nsf)$ has a
  fitting  guarded TGD.

  To this end we will construct a \GTGD $\rho$ such that $N\not\models\rho$ for all $N\in\Nsf$, and $J\models\rho$ for all $J\in\Psf$. Consider $( \prod \Nsf
  \vert_M, \bar{M})$. Note that since $M$ is non-empty and
  maximally guarded, $\prod \Nsf \vert_M$ cannot be empty, and thus the
  canonical CQ $q_{(\prod \Nsf \vert_M, \bar{M})}$ is defined. Then the body
  of $\rho$ is $\varphi(\bar{x})$, where the CQ $q_M(\bar{x}) =
  \varphi(\bar{x})$ is obtained from $q_{(\prod \Nsf \vert_M, \bar{M})}$ by
  renaming the free variables. Clearly $\bar{M}\in q_M(\prod \Nsf)$, and thus Lemma~\ref{lem:product_cq} yields $\bar{M}[i]\in q_M(N_i)$ for every $i\in[n]$.
  Since $M$ is guarded, there is a fact in $\prod \Nsf \vert_M$
  that contains every value from $\mn{adom}(\prod \Nsf \vert_M)$. Then by
  definition there exists an atom in $q_M$ that mentions every variable from
  $\mn{var}(q_M)$, and thus $q_M$ is guarded, ensuring that any TGD with
  $\varphi(\bar{x})$ as body is guarded. Concerning the head of $\rho$, we make
  a case distinction.

  First assume the violation of Condition~1, so $S_M = \emptyset$.
  Since for every $i\in [k]$, the tuple $\bar{M}[i]$ is non-total in
  $N_i$, there exists a CQ $q_i(\bar x) = \exists\bar{z}_i\, \psi_i(\bar x)$, such that $\bar{M}[i]\notin q_i(N_i)$. 
  Let $\rho = \varphi(\bar{x}, \bar{y}) \rightarrow
  \exists \bar{z} \, \psi(\bar{x}, \bar{z})$, where $\psi = \bigwedge_{i\in [k]} \psi_i$, and $\bar z = \bigcup_{i\in [k]} \bar z_i$. For brevity define $q_H(\bar x) = \exists \bar{z} \, \psi(\bar x, \bar z)$. Clearly $\bar{M}[i]\notin q_H(N_i)$, and thus $N_i\not\models \rho$ for $i\in[n]$.
  Now let $J\in\Psf$.  To show that
  $J\models\rho$, it suffices to argue that $q_M(J) = \emptyset$. Suppose to the
  contrary that there is a tuple $\bar{a}$ with $\bar{a}\in q_M(J)$. Then there
  is a homomorphism $h$ from $q_M(\bar{x})$ to $(J,\bar{a})$. By appropriately
  renaming the elements in the domain of $h$ to values from $\mn{adom}(\prod
  \Nsf \vert_M)$, we obtain a homomorphism witnessing $(\prod \Nsf\vert_M,
  \bar{M}) \to (J, \bar{a})$. Hence $(J, \bar{a})\in S_M$, which contradicts the
  assumption that $S_M=\emptyset$.

  Now assume that Condition~1 is satisfied (and thus $S_M$ is non-empty), but
  Condition~2 is violated. Let $(K, \bar{c}) = \prod S_M$. Let $\rho =
  \varphi(\bar{x}, \bar{y}) \rightarrow \exists \bar{z} \, \psi(\bar{x},
  \bar{z})$, where $\exists \bar{z} \, \psi(\bar{x},\bar{z})$ is obtained from
  the CQ $q_{(K^*, \bar{c}^*)}$ by renaming the free variables. For brevity, let
  $q_\Pi(\bar{x})=\exists \bar{z} \, \psi(\bar{x},\bar{z})$. Then the violation
  of Condition~2 by $M$ implies $\bar{M}[i]\not\in q_\Pi(N_i)$, and thus
  $N_i\not\models \rho$ for $i\in[k]$. To see that all positive examples satisfy
  $\rho$, let $J\in\Psf$ and $\bar{a}\in q_M(J)$. Then $(J, \bar{a})\in S_M$,
  which by Lemma~\ref{lem:product_homomorphism} implies $(K, \bar{c})\rightarrow
  (J, \bar{a})$. Clearly $(K^*, \bar{c}^*)\rightarrow (K, \bar{c})$ and thus
  $\bar{a}\in q_\Pi(J)$, so $J\models \rho$.

  \noindent ``$\Leftarrow$''. Assume that for all non-empty maximally guarded $M
  \subseteq \mn{adom}(\prod \Nsf)$, such that $\bar{M}[i]$ is non-total in $N_i$
  for $i\in [k]$, Conditions~1 and~2 are satisfied. Further assume, contrary to
  what we aim to show, that there exists a \GTGD that fits $(\Psf, \Nsf)$.

  Let $\rho = \varphi(\bar{x}, \bar{y}) \rightarrow \exists \bar{z} \,
  \psi(\bar{x}, \bar{z})$ be such a TGD. For brevity, define $q(\bar{x}) =
  \exists \bar{y} \, \varphi$ and $p(\bar{x}) = \exists \bar{z} \, \psi$. As
  $\rho$ fits $(\Psf, \Nsf)$, it follows that for every $i\in[k]$ there is some
  $\bar{a}_i\in \mn{adom}(N_i)^{|\bar{x}|}$ with $\bar{a}_i \in q(N_i)$ and
  $\bar{a}_i \notin p(N_i)$, witnessed by a homomorphism $h_{q \to N_i}$ from
  $q(\bar x)$ to $(N_i, \bar a_i)$. Our goal is to show that for some $i$, we
  have $\bar a_i\in p(N_i)$, thereby obtaining the desired contradiction. To
  utilize both conditions, we first need to find an appropriate non-empty
  maximally guarded set $M\subseteq \mn{adom}(\prod \Nsf)$.

  Consider the instance $(\prod \Nsf, \bar b)$, where $\bar b = \bar{a}_1 \times
  \cdots \times \bar a_k$. Define $h_{q \to \Pi}: \mn{adom}(I_q) \to
  \mn{adom}(\prod \Nsf)$ by $h_{q \to \Pi}(y) = (h_{q \to N_1}(y),\ldots, h_{q
  \to N_k}(y))$. Since $\rho$ is guarded, so is $q$, and thus,
  there exists a maximally guarded set $M\subseteq \mn{adom}(\prod \Nsf)$ such
  that $h_{q \to \Pi}(\mn{adom}(I_q)) \subseteq M$. The non-emptiness of $M$
  follows from the fact that $I_q$, as the canonical instance of $q$, trivially
  has a non-empty active domain. By the definition of products, $h_{q \to \Pi}$
  is a homomorphism witnessing $\bar b\in q(\prod \Nsf)$, and since $\bar b\subseteq M$, this implies that $h_{q \to \Pi}$ also witnesses $\bar b\in
  q(\prod \Nsf\vert_M)$.

  To ensure that $M$ satisfies Conditions~1 and~2, it remains to show that
  $\bar{M}[i]$ is non-total in $N_i$ for all $i \in [k]$. Let $i \in [k]$ and
  observe that $\bar{b}[i] = \bar{a}_i$, and thus $\bar{b}[i]$ is non-total in
  $N_i$ because $\bar{a}_i \notin p(N_i)$. As an immediate consequence of the
  definition of totality, every $M'\subseteq \mn{adom}(N_i)$ with $\bar b[i]
  \subseteq M'$ is also non-total in $N_i$. Note that $(d_1,\ldots, d_k) \in M$
  implies $d_i \in \bar{M}[i]$ by definition of products. Therefore
  $\bar{b}\subseteq M$ implies $\bar{b}[i] \subseteq \bar{M}[i]$ and thus
  $\bar{M}[i]$ is non-total in $N_i$. Hence, $M$ satisfies
  Conditions~1 and~2.

  This allows us to invoke Condition~1, which states that the set $S_M$ is
  non-empty, so let $(P_1, \bar d_1), \ldots, (P_m, \bar d_m)$ be an enumeration
  of the pointed instances in $S_M$, with $m >0$. Consider $(P_j, \bar d_j)$ for
  some $j\in[m]$. The definition of $S_M$ yields $(\prod \Nsf\vert_M, \bar M)
  \to (P_j, \bar d_j)$. Composing $h_{q \to \Pi}$ with the witnessing
  homomorphism yields a homomorphism $h_{q \to P_j}$ from $q$ to $P_j$ such that
  $h_{q \to P_j}(\bar x) \subseteq \bar d_j$. Recall that $\rho$ is a fitting
  TGD and therefore $P_j\in\Psf$ implies $h_{q \to P_j} (\bar{x}) \in p(P_j)$.
  Hence, there exists a homomorphism $g_{p \to P_j}$ from $p(\bar x)$ to
  $(P_j,h_{q \to P_j} (\bar x))$.

  Let $(K, \bar c) = \prod S_M$ and define $g_{p \to K}: \mn{adom}(I_p) \to
  \mn{adom}(K)$ by setting $g_{p \to K}(y) = (g_{p \to P_1}(y), \ldots, g_{p \to
  P_m}(y))$. By construction $g_{p \to K}(\bar x) \subseteq \bar c$ and by
  definition  of products $g_{p \to K}$ is a homomorphism. Define a homomorphism
  $g_{p \to K^*}$ from $p(\bar x)$ to $(K^*, \bar c^*)$ by $g_{p \to K^*}(y) =
  c^*_i$ if $g_{p \to K}(y) = c_i$ and $g_{p \to K^*}(y) = g_{p \to K}(y)$
  otherwise. Clearly $g_{p \to K^*}(\bar x)\subseteq \bar c^*$ and since $g_{p
  \to K}$ is a homomorphism, the definition of diversification guarantees that
  $g_{p \to K^*}$ is also a homomorphism.

  By Condition~2 we have $(K^*, c^*) \to (N_i, \bar M[i])$ for some $i\in [k]$.
  Composing $g_{p \to K^*}$ with the witnessing homomorphism yields  a
  homomorphism $g_{p \to N_i}$ from $p$ to $N_i$ such that $g_{p \to N_i}(\bar x) \subseteq \bar M[i]$. We next establish that in fact $g_{p \to N_i}(\bar x)
  = \bar a_i$, thus showing $\bar a_i\in p(N_i)$ and thereby obtaining a
  contradiction, as desired. Let $x\in \bar x$ and let $\bar M = (M_1,\ldots,
  M_n)$. Observe that if $h_{q \to \Pi}(x) = M_\ell$ for $\ell\in[n]$, then
  $h_{q \to N_i}(x)$ is the $\ell$-th element of $\bar{M}[i]$ due to
  construction of $h_{q \to \Pi}$. Furthermore the construction of the
  homomorphisms $h_{q \to P_j}$, $g_{p \to P_j}$, $g_{p \to K}$, $g_{p \to K^*}$
  and $g_{p \to N_i}$ all preserve the order of the original mapping of $\bar x$
  to $\bar M$ induced by $h_{q \to \Pi}$. Thus $g_{p \to N_i}(x) = h_{q \to
  N_i}(x)$, which entails $g_{p \to N_i}(\bar x) = \bar a_i$, as required. 
\end{proof}

For easier reference, we spell out the characterizations for \FONETGD,
\FGTGD and \TGD explicitly below. In order to prove them, we need the following
lemma.

\begin{lemma}\label{lem:nonemptyfrontier} Let $\rho$ be a TGD over \Smc such
  that $\rho$ has no frontier variables. Then, there exists a TGD $\rho'$ over
  \Smc with exactly one frontier variable such that, for all finite
  \Smc-instances $I$, the following equivalence holds: $$I\models \rho \text{ if
  and only if } I \models \rho'.$$
\end{lemma}
\noindent\begin{proof}\ 
  Let $\rho = \varphi(\bar y) \rightarrow \exists \bar{z} \, \psi
  (\bar z)$ be a TGD over \Smc with an empty frontier and let $I$ be a finite
  \Smc-instance. Consider an arbitrary atom $R(\bar x)$ of $\varphi(\bar y)$ and
  some $x\in \bar x$. Let $R(\bar v)$ be the result of replacing every $x_i$,
  except $x$, by a variable $v_i$ not occurring in $\rho$. Consider the TGD
  $\rho' = \varphi(\bar y) \rightarrow \exists \bar{w} \exists \bar{z} \,
  (\psi(\bar z) \wedge R(\bar v))$, where $\bar w = \bar v \setminus x$. Since
  the frontier of $\rho'$ is $x$, $\rho'$ is frontier-one and it is easy to see
  that for every instance $I$ the equivalence $I\models \rho \Leftrightarrow I
  \models \rho'$ holds.
\end{proof}

Before we attend to the characterization of fitting FGTGDs, we show another lemma. The proof 
of the \FGTGD characterization relies on guarded sets rather than maximally guarded sets. However, 
a characterization in terms of maximally guarded sets is desirable. As we shall see later, it enables a complexity analysis that yields a tight upper bound for fitting \FGTGD and ontology existence. The following lemma will be used to bridge the gap between guarded sets and maximally guarded sets.

\begin{lemma}\label{lem:supsets}
  Let $(\Psf, \Nsf)$ be a fitting instance where $\Nsf = \{N_1,\ldots, N_k\}$.
  Let $\emptyset \subsetneq M \subseteq M' \subseteq \mn{adom}(\prod \Nsf)$ be such that 
  $M'$ satisfies the following conditions:
  \begin{enumerate}
    \item the following set is non-empty:
        $$
          \begin{array}{r@{\;}c@{\;}l}
            S_{M'} &=&
            \{(J,\bar{b}) \mid J\in\Psf \text{ and } \bar{b}\in\mn{adom}(J)^{|M'|} \text{ such that}\\[1mm]
            && \hspace*{12mm} (\prod \Nsf,\bar{M}') \to (J,\bar{b})\}
           \end{array}$$
    \item $(K^*, \bar{c}^*) \to (N_i, \bar{M}'[i])$ for some $i\in[k]$ where $(K,
           \bar{c}) = \prod S_{M'}$. 
  \end{enumerate}
  Then $M$ also satisfies Conditions~1 and~2.
\end{lemma}
\noindent\begin{proof}\ 
  Let $\emptyset \subsetneq M \subseteq M' \subseteq
  \mn{adom}(\prod \Nsf)$ such that $M'$ satisfies Conditions~1 and~2. Let $\pi$
  be the projection mapping a tuple of length $|M'|$ to a subtuple of length
  $|M|$, induced by $\bar M'$ and $\bar M$, that is $\pi(\bar M') = \bar M$.
  Note that $\pi$ is well-defined because $M\subseteq M'$ and $\bar M'$ contains
  no value more than once. Observe that
  \begin{enumerate}
    \item every homomorphism witnessing $(\prod \Nsf, \bar M') \to (I, \bar a)$
    also witnesses $(\prod \Nsf, \bar M) \to (I, \pi(\bar a))$, and
    \item every homomorphism witnessing $(\prod \Nsf, \bar M) \to (I, \bar a)$
    also witnesses $(\prod \Nsf, \bar M') \to (I, \bar a')$ for some $\bar a'$ with $\pi(\bar a') = \bar a$.
  \end{enumerate}

  Since $M'$ satisfies Condition~1, $S_{M'}$ is non-empty. Then it is an
  immediate consequence of Observation~1 that $S_M$ is also non-empty.

  Define $(K, \bar c) = \prod S_M$ and $(K', \bar c') = \prod S_{M'}$. In order
  to prove that $M$ satisfies Condition~2, we first show that $(K^*, \bar c^*)
  \to (K'{}^*, \pi(\bar c'{}^*))$. Let $|S_M| = n$ and $|S_{M'}|= m$. It is not
  hard to see from Observations~1 and~2 that $n \leq m$ and that any $I\in\Psf$
  occurring in some pointed instance of $S_M$, must also occur in some pointed
  instance of $S'_M$ and vice versa. Define $f: [m] \to [n]$ such that for
  $j\in[m]$, if $(I, \bar a)$ is the $j$-th component of $(K', \bar c')$, then
  $f(j) = i$, where $i$ is the position of $(I, \pi(\bar a))$ as a component of
  $(K, \bar c)$.
  
  This allows us to define a homomorphism from $(K^*, \bar c^*)$ to $(K'{}^*, \pi(\bar c'{}^*))$. Let $h: \mn{adom}(K^*) \to \mn{adom}(K'{}^*)$,
  such that $h(\bar c^*) = \pi(\bar c'{}^*)$ and for every $a =
  (a_1,\ldots,a_n)\not\in \bar c^*$, $h(a) = (a_{f(1)},\ldots, a_{f(m)})$.
  Intuitively, every $a$ that is not a clone is sent to an $h(a)$ that has only
  components that already occur in $a$. To see that $h$ indeed is a
  homomorphism, let $R(\bar a)\in K^*$. If none of the values of the fact $R(\bar a)$ is from $\bar c^*$, by the definition of $h$ and $f$, for every $j\in[m]$, $h(\bar a)[j] =
  \bar a[i]$ for some $i\in[n]$. Observe that by the definition of products, $R(\bar a)\in K^*$ implies that $R(\bar a[i])$ is a fact of the $i$-th component of $S_M$ for
  every $i\in[n]$. Thus by the definition of products $R(h(\bar a))\in K' \subseteq K'{}^*$. If some values from $\bar c^*$ do occur in $\bar a$, diversification guarantees that there is a fact $R(\bar a')\in K$, where
  in place of any cloned $c^*$, the corresponding $c$ occurs. Since $R(\bar a')$
  contains no cloned values, the previous argument applies and thus $R(h(\bar a'))\in K'$. Observe that $f$ ensures that $h$ preserves the order induced by
  $\pi$, that is, if $c$ is the $i$-th element of $\bar c$, then $h(c)$ is the
  $i$-th element of $\pi(\bar c')$. Since by definition $h(\bar c^*) = \pi(\bar c'{}^*)$, the fact $R(h(\bar a))$ is obtained from $R(h(\bar a'))$ by
  diversification, and thus $R(h(\bar a))\in K'{}^*$.

  We will now show that $(K^*, \bar c^*) \to (K'{}^*, \pi(\bar c'{}^*))$ implies
  that $M$ satisfies Condition~2. Since $M'$ satisfies Condition~2, there is an
  $i\in[k]$ such that $(K'{}^*, (\bar c'{}^*)) \to (N_i, \bar M'[i])$, which
  clearly implies $(K'{}^*, \pi(\bar c'{}^*)) \to (N_i, \pi(\bar M'[i]))$. The
  composition yields a homomorphism witnessing $(K^*, \bar c^*)\to (N_i,
  \pi(\bar M'[i]))$. By definition $\pi(\bar M')[i] = M[i]$, and it
  can be easily verified that $\pi(\bar M')[i] = \pi(\bar M'[i])$. Thus $(K^*,
  \bar c^*)\to (N_i, M[i])$, which concludes the proof.
\end{proof}
\begin{theorem}
  \label{thm:mtc:FGTGDfitting} Let $(\Psf, \Nsf)$ be a fitting instance where
  $\Nsf = \{N_1,\dots,N_k\}$. Then no \FGTGD fits $(\Psf,\Nsf)$ if and only if
  for every non-empty maximally guarded set $M \subseteq \mn{adom}(\prod \Nsf)$ such that
  $\bar{M}[i]$ is non-total in $N_i$ for all $i\in [k]$, the following
  conditions are satisfied:
  \begin{enumerate}
    \item the following set is non-empty:
        $$
          \begin{array}{r@{\;}c@{\;}l}
            S_M &=&
            \{(J,\bar{b}) \mid J\in\Psf \text{ and } \bar{b}\in\mn{adom}(J)^{|M|} \text{ such that}\\[1mm]
            && \hspace*{12mm} (\prod \Nsf,\bar{M}) \to (J,\bar{b})\}
           \end{array}$$
    \item $(K^*, \bar{c}^*) \to (N_i, \bar{M}[i])$ for some $i\in[k]$ where $(K,
           \bar{c}) = \prod S_M$. 
  \end{enumerate}
\end{theorem}
\noindent\begin{proof}\ 
    Instead of showing the above statement, we prove an equivalent one, where instead of maximally guarded sets, guarded sets are considered. 
    First we show that both statements are indeed equivalent. Let $(\Psf, \Nsf)$ be a fitting instance. It is sufficient to show that every non-empty maximally guarded set $M\subseteq \mn{adom}(\prod \Nsf)$, such that $\bar{M}[i]$ is non-total in $N_i$ for all $i\in[k]$, satisfies Conditions~1 and~2 if and only if the same holds for every non-empty guarded set $M'\subseteq \mn{adom}(\prod \Nsf)$.
    Since every maximally guarded set is guarded, the direction from guarded to maximally guarded sets holds trivially. 
    Regarding the other direction, assume that every maximally guarded set $M\subseteq \mn{adom}(\prod \Nsf)$, such that $\bar{M}[i]$ is non-total in $N_i$ for all $i\in[k]$, satisfies Conditions~1 and~2. 
    Consider a non-empty guarded set $M'\subseteq \mn{adom}(\prod \Nsf)$, such that $\bar{M}'[i]$ is non-total in $N_i$ for all $i\in[k]$. Then there is a non-empty maximally guarded set $M\subseteq \mn{adom}(\prod \Nsf)$ with $M'\subseteq M$.
    It is not hard to see from the definitions of totality and products, that $\bar{M}'[i]$ being non-total in $N_i$ for all $i\in[k]$ implies that $\bar{M}[i]$ must also be non-total in $N_i$ for all $i\in[k]$, and thus $M$ satisfies Conditions~1 and~2. By Lemma~\ref{lem:supsets} this implies that $M'$ satisfies Conditions~1 and~2, establishing the desired equivalence.
    \medskip

    We will now continue to show the statement for guarded sets, rather than maximally guarded sets. The rest of the proof follows the same structure as the proof of Theorem~\ref{thm:mtc:GTGDfitting}, with significant changes in the reasoning explicitly marked by the symbol~$(\dagger)$.  
  \medskip
  
  \noindent ``$\Rightarrow$''. We show the contrapositive. Assume that there is some
  non-empty guarded $M \subseteq \mn{adom}(\prod \Nsf)$ such that $\bar{M}[i]$
  is non-total in $N_i$ for all $i\in [k]$, and at least one of the Conditions~1
  and~2 is violated. We have to show that $(\Psf, \Nsf)$ has a fitting
  frontier-guarded TGD.

  To this end we will construct an \FGTGD $\rho$ such that $N \not\models\rho$ for all $N\in\Nsf$, and $J\models\rho$ for all $J\in\Psf$. Consider $( \prod
  \Nsf, \bar{M})$. Note that since $M$ is non-empty, $\prod \Nsf$ cannot be
  empty, and thus the canonical CQ $q_{(\prod \Nsf, \bar{M})}$ is defined. Then
  the body of $\rho$ is $\varphi(\bar{x}, \bar{y})$, where the CQ $q_M(\bar{x})
  = \exists \bar{y} \, \varphi(\bar{x}, \bar{y})$ is obtained from $q_{(\prod
  \Nsf, \bar{M})}$ by renaming the free variables. Clearly $\bar{M}\in q_M(\prod
  \Nsf)$, and thus Lemma~\ref{lem:product_cq} yields $\bar M[i]\in q_M(N_i)$ for every $i\in[n]$.
  \begin{itemize}
    \item[$(\dagger)$] Since $M$ is guarded, there is a fact in
    $\prod \Nsf$ that contains every value from $\bar M$. Then by definition
    there exists an atom in $q_M$ that mentions every variable from $\bar x$,
    ensuring that any TGD with body $\varphi(\bar{x}, \bar{y})$ and frontier
    $\bar x$ is frontier-guarded. Concerning the head of $\rho$, we make a case distinction.
  \end{itemize}  

  First assume the violation of Condition~1, so $S_M = \emptyset$. Since for every $i\in [k]$, the tuple $\bar{M}[i]$ is non-total in
  $N_i$, there exists a CQ $q_i(\bar x) = \exists\bar{z}_i\, \psi_i(\bar x)$, such that $\bar{M}[i]\notin q_i(N_i)$. 
  Let $\rho = \varphi(\bar{x}, \bar{y}) \rightarrow
  \exists \bar{z} \, \psi(\bar{x}, \bar{z})$, where $\psi = \bigwedge_{i\in [k]} \psi_i$, and $\bar z = \bigcup_{i\in [k]} \bar z_i$. For brevity define $q_H(\bar x) = \exists \bar{z} \, \psi(\bar x, \bar z)$. Clearly $\bar{M}[i]\notin q_H(N_i)$, and thus $N_i\not\models \rho$ for $i\in[n]$. Now let $J\in\Psf$.  To show that
  $J\models\rho$, it suffices to argue that $q_M(J) = \emptyset$. Suppose to the
  contrary that there is a tuple $\bar{a}$ with $\bar{a}\in q_M(J)$. Then there
  is a homomorphism $h$ from $q_M(\bar{x})$ to $(J,\bar{a})$. By appropriately
  renaming the elements in the domain of $h$ to values from $\mn{adom}( \prod
  \Nsf)$,  we obtain a homomorphism witnessing $(\prod \Nsf, \bar{M}) \to (J,
  \bar{a})$. Hence $(J, \bar{a})\in S_M$, which contradicts the assumption that
  $S_M=\emptyset$.

  Now assume that Condition~1 is satisfied (and thus $S_M$ is non-empty), but
  Condition~2 is violated. Let $(K, \bar{c}) = \prod S_M$. Let $\rho =
  \varphi(\bar{x}, \bar{y}) \rightarrow \exists \bar{z} \, \psi(\bar{x},
  \bar{z})$, where $\exists \bar{z} \, \psi(\bar{x},\bar{z})$ is obtained from
  the CQ $q_{(K^*, \bar{c}^*)}$ by renaming the free variables. For brevity, let
  $q_\Pi(\bar{x})=\exists \bar{z} \, \psi(\bar{x},\bar{z})$. Then the violation
  of Condition~2 by $M$ implies $\bar{M}[i]\not\in q_\Pi(N_i)$, and thus
  $N_i\not\models \rho$ for $i\in[k]$. To see that all positive examples satisfy
  $\rho$, let $J\in\Psf$ and $\bar{a}\in q_M(J)$. Then $(J, \bar{a})\in S_M$,
  which by Lemma~\ref{lem:product_homomorphism} implies $(K, \bar{c})\rightarrow
  (J, \bar{a})$. Clearly $(K^*, \bar{c}^*)\rightarrow (K, \bar{c})$ and thus
  $\bar{a}\in q_\Pi(J)$, so $J\models \rho$.

  \noindent ``$\Leftarrow$''. Assume that for all non-empty guarded $M \subseteq
  \mn{adom}(\prod \Nsf)$, such that $\bar{M}[i]$ is non-total in $N_i$ for $i\in
  [k]$, Conditions~1 and~2 are satisfied. Further assume, contrary to what we
  aim to show, that there exists an \FGTGD that fits $(\Psf,
  \Nsf)$.

  \begin{itemize}
    \item[$(\dagger)$] By Lemma~\ref{lem:nonemptyfrontier}, this implies the
    existence of a fitting \FGTGD with a non-empty frontier.
  \end{itemize}

  Let $\rho = \varphi(\bar{x}, \bar{y}) \rightarrow \exists \bar{z} \, \psi(\bar{x}, \bar{z})$ be such a TGD. For brevity, define $q(\bar{x}) =
  \exists \bar{y} \, \varphi$ and $p(\bar{x}) = \exists \bar{z} \, \psi$. As
  $\rho$ fits $(\Psf, \Nsf)$, it follows that for every $i\in[k]$ there is some
  $\bar{a}_i\in \mn{adom}(N_i)^{|\bar{x}|}$ with $\bar{a}_i \in q(N_i)$ and
  $\bar{a}_i \notin p(N_i)$, witnessed by a homomorphism $h_{q \to N_i}$ from
  $q(\bar x)$ to $(N_i, \bar a_i)$. Our goal is to show that for some $i$, we
  have $\bar a_i\in p(N_i)$, thereby obtaining the desired contradiction. To
  utilize both conditions, we first need to find an appropriate non-empty
  guarded set $M\subseteq \mn{adom}(\prod \Nsf)$.

  Consider the instance $(\prod \Nsf, \bar b)$, where $\bar b = \bar{a}_1 \times
  \cdots \times \bar a_k$. Define $h_{q \to \Pi}: \mn{adom}(I_q) \to
  \mn{adom}(\prod \Nsf)$ by $h_{q \to \Pi}(y) = (h_{q \to N_1}(y),\ldots, h_{q
  \to N_k}(y))$.

  \begin{itemize}
    \item[$(\dagger)$] By the definition of products, $h_{q \to \Pi}$ is a
    homomorphism witnessing that $\bar b\in q(\prod \Nsf)$. Let $M = \{b\in
    \mn{adom}(\prod \Nsf) \mid b\in \bar b\}$. Since $\rho$ has a non-empty
    frontier and is frontier-guarded, and $h_{q \to \Pi}$ maps $\bar x$ to $\bar b$, $M$ is non-empty and guarded.
  \end{itemize}
  
  To ensure that $M$ satisfies Conditions~1 and~2, it remains to show that
  $\bar{M}[i]$ is non-total in $N_i$ for all $i \in [k]$. Let $i \in [k]$ and
  observe that $\bar{b}[i] = \bar{a}_i$, and thus $\bar{b}[i]$ is non-total in
  $N_i$ because $\bar{a}_i \notin p(N_i)$. As an immediate consequence of the
  definition of totality, every $M'\subseteq \mn{adom}(N_i)$ with $\bar b[i]
  \subseteq M'$ is also non-total in $N_i$. Note that $(d_1,\ldots, d_k) \in M$
  implies $d_i \in \bar{M}[i]$ by definition of products. Therefore
  $\bar{b}\subseteq M$ implies $\bar{b}[i] \subseteq \bar{M}[i]$ and thus
  $\bar{M}[i]$ is non-total in $N_i$. Hence, $M$ satisfies
  Conditions~1 and~2.

  This allows us to invoke Condition~1, which states that the set $S_M$ is
  non-empty, so let $(P_1, \bar d_1), \ldots, (P_m, \bar d_m)$ be an enumeration
  of the pointed instances in $S_M$, with $m >0$. Consider $(P_j, \bar d_j)$ for
  some $j\in[m]$. The definition of $S_M$ yields $( \prod \Nsf, \bar M) \to
  (P_j, \bar d_j)$. Composing $h_{q \to \Pi}$ with the witnessing homomorphism
  yields a homomorphism $h_{q \to P_j}$ from $q$ to $P_j$ such that $h_{q \to
  P_j}(\bar x) \subseteq \bar d_j$. Recall that $\rho$ is a fitting TGD and
  therefore $P_j\in\Psf$ implies $h_{q \to P_j} (\bar{x}) \in p(P_j)$. Hence,
  there exists a homomorphism $g_{p \to P_j}$ from $p(\bar x)$ to $(P_j,h_{q \to
  P_j} (\bar x))$.

  Let $(K, \bar c) = \prod S_M$ and define $g_{p \to K}: \mn{adom}(I_p) \to
  \mn{adom}(K)$ by setting $g_{p \to K}(y) = (g_{p \to P_1}(y), \ldots, g_{p \to
  P_m}(y))$. By construction $g_{p \to K}(\bar x) \subseteq \bar c$ and by
  definition  of products $g_{p \to K}$ is a homomorphism. Define a homomorphism
  $g_{p \to K^*}$ from $p(\bar x)$ to $(K^*, \bar c^*)$ by $g_{p \to K^*}(y)
  = c^*_i$ if $g_{p \to K}(y) = c_i$ and $g_{p \to K^*}(y) = g_{p \to K}(y)$
  otherwise. Clearly $g_{p \to K^*}(\bar x)\subseteq \bar c^*$ and since $g_{p
  \to K}$ is a homomorphism, the definition of diversification guarantees that
  $g_{p \to K^*}$ is also a homomorphism.

  By Condition~2 we have $(K^*, c^*) \to (N_i, \bar M[i])$ for some $i\in [k]$.
  Composing $g_{p \to K^*}$ with the witnessing homomorphism yields  a
  homomorphism $g_{p \to N_i}$ from $p$ to $N_i$ such that $g_{p \to N_i}(\bar x) \subseteq \bar M[i]$. We next establish that in fact $g_{p \to N_i}(\bar x)
  = \bar a_i$, thus showing $\bar a_i\in p(N_i)$ and thereby obtaining a
  contradiction, as desired. Let $x\in \bar x$ and let $\bar M = (M_1,\ldots,
  M_n)$. Observe that if $h_{q \to \Pi}(x) = M_\ell$ for $\ell\in[n]$, then
  $h_{q \to N_i}(x)$ is the $\ell$-th element of $\bar{M}[i]$ due to
  construction of $h_{q \to \Pi}$. Furthermore the construction of the
  homomorphisms $h_{q \to P_j}$, $g_{p \to P_j}$, $g_{p \to K}$, $g_{p \to K^*}$
  and $g_{p \to N_i}$ all preserve the order of the original mapping of $\bar x$
  to $\bar M$ induced by $h_{q \to \Pi}$. Thus $g_{p \to N_i}(x) = h_{q \to
  N_i}(x)$, which entails $g_{p \to N_i}(\bar x) = \bar a_i$, as required. 
\end{proof}
For \FONETGD, we provide a characterization that slightly differs from the one in Theorem~\ref{thm:charthreeinone}. 
Instead of singleton sets $M \subseteq \mn{adom}(\prod \Nsf)$, it considers values $ \bar a \in \mn{adom}(\prod \Nsf)$. Moreover, it uses the interpretation $\prod S_{\bar a}$ in the second condition, rather
than its diversification, as the latter is redundant for
\FONETGD.
\begin{theorem}
    \label{thm:mtc:FONETGDfitting}
    Let $(\Psf, \Nsf)$ be a fitting instance where $\Nsf = \{N_1,\dots,N_k\}$.
	Then no \FONETGD fits $(\Psf,\Nsf)$ if and only if for every $\bar a =
	(a_1,\ldots, a_k) \in \mn{adom}(\prod \Nsf)$ such that $a_i$ is non-total in
	$N_i$ for all $i\in [k]$, the following conditions are satisfied:
	\begin{enumerate}
		\item the following set is non-empty:
		      $$
		      \begin{array}{r@{\;}c@{\;}l}
			      S_{\bar a} & = &
			      \{(J, b) \mid J\in\Psf \text{ and } b\in\mn{adom}(J) \text{ such that} \\[1mm]
			          &   & \hspace*{12mm} (\prod \Nsf,\bar{a}) \to (J,b)\}
		      \end{array}$$
		\item $\prod S_{\bar a} \to (N_i, a_i)$ for some $i\in[k]$.
	\end{enumerate}
\end{theorem}
\noindent\begin{proof}\ 
    The proof follows the same structure as the proof of Theorem~\ref{thm:mtc:GTGDfitting}, 
    with significant changes in the reasoning explicitly marked by the symbol~$(\dagger)$.

  \medskip
  
  \noindent ``$\Rightarrow$''. We show the contrapositive. Assume that there is some $\bar a =(a_1,\ldots, a_k) \in \mn{adom}(\prod \Nsf)$ such that $a_i$ is
  non-total in $N_i$ for all $i\in [k]$, and at least one of the Conditions~1
  and~2 is violated. We have to show that $(\Psf, \Nsf)$ has a fitting
  frontier-one TGD.

  To this end we will construct an \FONETGD $\rho$ such that $N
  \not\models\rho$ for all $N\in \Nsf$, and $J\models\rho$ for all $J\in\Psf$. Consider $(\prod \Nsf,
  \bar{a})$.

  \begin{itemize}
    \item[$(\dagger)$] Note that since $\bar{a} \in \mn{adom}(\prod \Nsf)$, the canonical CQ $q_{(\prod
    \Nsf, \bar{a})}$ is defined and has exactly one free variable.
  \end{itemize}
  
  Then the body of $\rho$ is $\varphi(x, \bar{y})$, where the CQ $q_{\bar a}(x) =
  \exists \bar{y} \, \varphi(x, \bar{y})$ is obtained from $q_{(\prod \Nsf,
  \bar{a})}$ by renaming the free variable. Clearly $\bar{a}\in q_{\bar a}(\prod
  \Nsf)$, and thus Lemma~\ref{lem:product_cq} yields $a_i\in q_{\bar a}(N_i)$ for every $i\in[n]$. Concerning the head of $\rho$, we make a case distinction.
  
  First assume the violation of Condition~1, so $S_{\bar a} = \emptyset$.
  Since for every $i\in [k]$, the value $a_i$ is non-total in
  $N_i$, there exists a CQ $q_i(x) = \exists\bar{z}_i\, \psi_i(x)$, such that $a_i\notin q_i(N_i)$. 
  Let $\rho = \varphi(x, \bar{y}) \rightarrow
  \exists \bar{z} \, \psi(x, \bar{z})$, where $\psi = \bigwedge_{i\in [k]} \psi_i$, and $\bar z = \bigcup_{i\in [k]} \bar z_i$. For brevity define $q_H(x) = \exists \bar{z} \, \psi(x, \bar z)$. Clearly $a_i\notin q_H(N_i)$, and thus $N_i\not\models \rho$ for $i\in[n]$.
  Now let $J\in\Psf$. To show that $J\models\rho$, it
  suffices to argue that $q_{\bar a}(J) = \emptyset$. Suppose to the contrary that
  there is a value $b$ with $b\in q_{\bar a}(J)$. Then there is a homomorphism $h$ from
  $q_{\bar a}(x)$ to $(J,b)$. By appropriately renaming the elements in the domain of
  $h$ to values from $\mn{adom}(\prod \Nsf)$, we obtain a homomorphism
  witnessing $(\prod \Nsf, \bar{a}) \to (J, b)$. Hence $(J, b)\in S_{\bar a}$, which
  contradicts the assumption that $S_{\bar a}=\emptyset$.

  Now assume that Condition~1 is satisfied (and thus $S_{\bar a}$ is non-empty), but
  Condition~2 is violated. Let $(K, c) = \prod S_{\bar a}$. Let $\rho = \varphi(x,
  \bar{y}) \rightarrow \exists \bar{z} \, \psi(x, \bar{z})$, where $\exists
  \bar{z} \, \psi(x,\bar{z})$ is obtained from the CQ $q_{(K,c)}$ by
  renaming the free variable. For brevity, let $q_\Pi(x)=\exists \bar{z} \,
  \psi(x,\bar{z})$. Then the violation
  of Condition~2 by $M$ implies $a_i\not\in q_\Pi(N_i)$, and thus
  $N_i\not\models \rho$ for $i\in[k]$. To
  see that all positive examples satisfy $\rho$, let $J\in\Psf$ and $b\in
  q_{\bar a}(J)$. Then $(J, b)\in S_{\bar a}$, which by Lemma~\ref{lem:product_homomorphism}
  implies $(K, c)\rightarrow (J, b)$. It follows that $b\in q_\Pi(J)$, and thus $J\models \rho$.

  \noindent ``$\Leftarrow$''. Assume that for all $\bar{a} = (a_1, \ldots, a_k) \in \mn{adom}(\prod
  \Nsf)$, such that $a_i$ is non-total in $N_i$ for $i\in [k]$,
  Conditions~1 and~2 are satisfied. Further assume, contrary to what we aim to
  show, that there exists an \FONETGD that fits $(\Psf, \Nsf)$.

  \begin{itemize}
    \item[$(\dagger)$] By Lemma~\ref{lem:nonemptyfrontier}, this implies the
    existence of a fitting \FONETGD with a frontier of size exactly $1$.
  \end{itemize}

  Let $\rho = \varphi(x, \bar{y}) \rightarrow \exists \bar{z} \, \psi(x,
  \bar{z})$ be such a TGD. For brevity, define $q(x) = \exists \bar{y} \,
  \varphi$ and $p(x) = \exists \bar{z} \, \psi$. As $\rho$ fits $(\Psf, \Nsf)$,
  it follows that for every $i\in[k]$ there is some $a_i\in \mn{adom}(N_i)$ with
  $a_i \in q(N_i)$ and $a_i \notin p(N_i)$, witnessed by a homomorphism $h_{q
  \to N_i}$ from $q(x)$ to $(N_i, a_i)$. Our goal is to show that for some $i$,
  we have $a_i\in p(N_i)$, thereby obtaining the desired contradiction. To
  utilize both conditions, we first need to find an appropriate
  $\bar{a}\in \mn{adom}(\prod \Nsf)$.

  Consider the instance $(\prod \Nsf, \bar{a})$, where $\bar{a} = (a_1,\ldots, a_k)$. Define
  $h_{q \to \Pi}: \mn{adom}(I_q) \to \mn{adom}(\prod \Nsf)$ by $h_{q \to \Pi}(y)
  = (h_{q \to N_1}(y),\ldots, h_{q \to N_k}(y))$. By the definition of products, $h_{q \to \Pi}$ is a homomorphism witnessing that $\bar{a}\in q(\prod \Nsf)$.

  To ensure that $\bar{a}$ satisfies Conditions~1 and~2, it remains to show that
  $a_i$ is non-total in $N_i$ for all $i \in [k]$. However, for every $i\in[k]$, this is an immediate consequence of $a_i \notin p(N_i)$, and thus $\bar{a}$ indeed satisfies Conditions~1 and~2.

  This allows us to invoke Condition~1, which states that the set $S_{\bar a}$ is
  non-empty, so let $(P_1, d_1), \ldots, (P_m, d_m)$ be an enumeration of the
  pointed instances in $S_{\bar a}$, with $m >0$. Consider $(P_j, d_j)$ for some
  $j\in[m]$. The definition of $S_{\bar a}$ yields $(\prod \Nsf, \bar a) \to (P_j,
  d_j)$. Composing $h_{q \to \Pi}$ with the witnessing homomorphism yields a
  homomorphism $h_{q \to P_j}$ from $q$ to $P_j$ such that $h_{q \to P_j}(x) =
  d_j$. Recall that $\rho$ is a fitting TGD and therefore $P_j\in\Psf$ implies
  $d_j \in p(P_j)$. Hence, there exists a homomorphism $g_{p \to P_j}$ from
  $p(x)$ to $(P_j,d_j)$.

  Let $(K, c) = \prod S_{\bar a}$ and define $g_{p \to K}: \mn{adom}(I_p) \to
  \mn{adom}(K)$ by setting $g_{p \to K}(y) = (g_{p \to P_1}(y), \ldots, g_{p \to
  P_m}(y))$. By construction $g_{p \to K}(x) = c$ and by definition of products
  $g_{p \to K}$ is a homomorphism. 
  
  By Condition~2 we have $(K, c) \to (N_i, a_i)$ for some $i\in [k]$.
  Composing $g_{p \to K}$ with the witnessing homomorphism yields a
  homomorphism $g_{p \to N_i}$ from $p$ to $N_i$ such that $g_{p \to N_i}(x) =
  a_i$, thus showing $a_i\in p(N_i)$, a contradiction. 
\end{proof}
\begin{theorem}
  \label{thm:mtc:TGDfitting} Let $(\Psf, \Nsf)$ be a fitting instance where
  $\Nsf = \{N_1,\dots,N_k\}$. Then no \TGD fits $(\Psf,\Nsf)$ if and only if for
  every non-empty set $M \subseteq \mn{adom}(\prod \Nsf)$ such that 
  $\bar{M}[i]$ is non-total in $N_i$ for all $i\in [k]$, the following
  conditions are satisfied:
  \begin{enumerate}
    \item the following set is non-empty:
        $$
          \begin{array}{r@{\;}c@{\;}l}
            S_M &=&
            \{(J,\bar{b}) \mid J\in\Psf \text{ and } \bar{b}\in\mn{adom}(J)^{|M|} \text{ such that}\\[1mm]
            && \hspace*{12mm} (\prod \Nsf,\bar{M}) \to (J,\bar{b})\}
           \end{array}$$
    \item $(K^*, \bar{c}^*) \to (N_i, \bar{M}[i])$ for some $i\in[k]$ where $(K,
           \bar{c}) = \prod S_M$. 
  \end{enumerate}
\end{theorem}
\noindent\begin{proof}\
    The proof follows the same structure as the proof of Theorem~\ref{thm:mtc:GTGDfitting}, 
    with significant changes in the reasoning explicitly marked by the symbol~$(\dagger)$.
  
  \medskip
  
  \noindent ``$\Rightarrow$''. We show the contrapositive. Assume that there is some
  non-empty  $M \subseteq \mn{adom}(\prod \Nsf)$ such that $\bar{M}[i]$ is
  non-total in $N_i$ for all $i\in [k]$, and at least one of the Conditions~1
  and~2 is violated. We have to show that $(\Psf, \Nsf)$ has a fitting TGD.

  To this end we will construct a TGD $\rho$ such that $N \not\models\rho$ for all $N\in\Nsf$, and $J\models\rho$ for all $J\in\Psf$. Consider $(\prod
  \Nsf, \bar{M})$. Note that since $M$ is non-empty, $\prod \Nsf$ cannot be
  empty, and thus the canonical CQ $q_{(\prod \Nsf, \bar{M})}$ is defined. Then
  the body of $\rho$ is $\varphi(\bar{x}, \bar{y})$, where the CQ $q_M(\bar{x})
  = \exists \bar{y} \, \varphi(\bar{x}, \bar{y})$ is obtained from $q_{(\prod
  \Nsf, \bar{M})}$ by renaming the free variables. Clearly $\bar{M}\in q_M(\prod \Nsf)$, and thus Lemma~\ref{lem:product_cq} yields $\bar{M}[i]\in q_M(N_i)$ for every $i\in[n]$. Concerning the head of $\rho$, we make a case distinction.

  First assume the violation of Condition~1, so $S_M = \emptyset$.
  Since for every $i\in [k]$, the tuple $\bar{M}[i]$ is non-total in
  $N_i$, there exists a CQ $q_i(\bar x) = \exists\bar{z}_i\, \psi_i(\bar x)$, such that $\bar{M}[i]\notin q_i(N_i)$. 
  Let $\rho = \varphi(\bar{x}, \bar{y}) \rightarrow
  \exists \bar{z} \, \psi(\bar{x}, \bar{z})$, where $\psi = \bigwedge_{i\in [k]} \psi_i$, and $\bar z = \bigcup_{i\in [k]} \bar z_i$. For brevity define $q_H(\bar x) = \exists \bar{z} \, \psi(\bar x, \bar z)$. Clearly $\bar{M}[i]\notin q_H(N_i)$, and thus $N_i\not\models \rho$ for $i\in[n]$.
  Now let $J\in\Psf$.  To show that
  $J\models\rho$, it suffices to argue that $q_M(J) = \emptyset$. Suppose to the
  contrary that there is a tuple $\bar{a}$ with $\bar{a}\in q_M(J)$. Then there
  is a homomorphism $h$ from $q_M(\bar{x})$ to $(J,\bar{a})$. By appropriately
  renaming the elements in the domain of $h$ to values from $\mn{adom}( \prod
  \Nsf)$,  we obtain a homomorphism witnessing $(\prod \Nsf, \bar{M}) \to (J,
  \bar{a})$. Hence $(J, \bar{a})\in S_M$, which contradicts the assumption that
  $S_M=\emptyset$.

  Now assume that Condition~1 is satisfied (and thus $S_M$ is non-empty), but
  Condition~2 is violated. Let $(K, \bar{c}) = \prod S_M$. Let $\rho =
  \varphi(\bar{x}, \bar{y}) \rightarrow \exists \bar{z} \, \psi(\bar{x},
  \bar{z})$, where $\exists \bar{z} \, \psi(\bar{x},\bar{z})$ is obtained from
  the CQ $q_{(K^*, \bar{c}^*)}$ by renaming the free variables. For brevity, let
  $q_\Pi(\bar{x})=\exists \bar{z} \, \psi(\bar{x},\bar{z})$. Then the violation
  of Condition~2 by $M$ implies $\bar{M}[i]\not\in q_\Pi(N_i)$, and thus
  $N_i\not\models \rho$ for $i\in[k]$. To see that all positive examples satisfy
  $\rho$, let $J\in\Psf$ and $\bar{a}\in q_M(J)$. Then $(J, \bar{a})\in S_M$,
  which by Lemma~\ref{lem:product_homomorphism} implies $(K, \bar{c})\rightarrow
  (J, \bar{a})$. Clearly $(K^*, \bar{c}^*)\rightarrow (K, \bar{c})$ and thus
  $\bar{a}\in q_\Pi(J)$, so $J\models \rho$.

  \noindent ``$\Leftarrow$''. Assume that for all $M \subseteq \mn{adom}(\prod \Nsf)$, such that
  $\bar{M}[i]$ is non-total in $N_i$ for $i\in [k]$, Conditions~1 and~2 are
  satisfied. Further assume, contrary to what we aim to show, that there exists
  a  TGD that fits $(\Psf, \Nsf)$.

  \begin{itemize}
    \item[$(\dagger)$] By Lemma~\ref{lem:nonemptyfrontier}, this implies the
    existence of a fitting TGD with a non-empty frontier.
  \end{itemize}

  Let $\rho = \varphi(\bar{x}, \bar{y}) \rightarrow \exists \bar{z} \, \psi(\bar{x}, \bar{z})$ be such a TGD. For brevity, define $q(\bar{x}) =
  \exists \bar{y} \, \varphi$ and $p(\bar{x}) = \exists \bar{z} \, \psi$. As
  $\rho$ fits $(\Psf, \Nsf)$, it follows that for every $i\in[k]$ there is some
  $\bar{a}_i\in \mn{adom}(N_i)^{|\bar{x}|}$ with $\bar{a}_i \in q(N_i)$ and
  $\bar{a}_i \notin p(N_i)$, witnessed by a homomorphism $h_{q \to N_i}$ from
  $q(\bar x)$ to $(N_i, \bar a_i)$. Our goal is to show that for some $i$, we
  have $\bar a_i\in p(N_i)$, thereby obtaining the desired contradiction. To
  utilize both conditions, we first need to find an appropriate non-empty set
  $M\subseteq \mn{adom}(\prod \Nsf)$.

  Consider the instance $(\prod \Nsf, \bar b)$, where $\bar b = \bar{a}_1 \times
  \cdots \times \bar a_k$. Define $h_{q \to \Pi}: \mn{adom}(I_q) \to
  \mn{adom}(\prod \Nsf)$ by $h_{q \to \Pi}(y) = (h_{q \to N_1}(y),\ldots, h_{q
  \to N_k}(y))$.
  
  \begin{itemize}
    \item[$(\dagger)$] By the definition of products, $h_{q \to \Pi}$ is a
    homomorphism witnessing that $\bar b\in q(\prod \Nsf)$. Let $M = \{b\in
    \mn{adom}(\prod \Nsf) \mid b\in \bar b\}$. Since $\rho$ has a non-empty
    frontier, $M$ is non-empty.
  \end{itemize}

  To ensure that $M$ satisfies Conditions~1 and~2, it remains to show that
  $\bar{M}[i]$ is non-total in $N_i$ for all $i \in [k]$. Let $i \in [k]$ and
  observe that $\bar{b}[i] = \bar{a}_i$, and thus $\bar{b}[i]$ is non-total in
  $N_i$ because $\bar{a}_i \notin p(N_i)$. As an immediate consequence of the
  definition of totality, every $M'\subseteq \mn{adom}(N_i)$ with $\bar b[i]
  \subseteq M'$ is also non-total in $N_i$. Note that $(d_1,\ldots, d_k) \in M$
  implies $d_i \in \bar{M}[i]$ by definition of products. Therefore
  $\bar{b}\subseteq M$ implies $\bar{b}[i] \subseteq \bar{M}[i]$ and thus
  $\bar{M}[i]$ is non-total in $N_i$. Hence, $M$ satisfies
  Conditions~1 and~2.

  This allows us to invoke Condition~1, which states that the set $S_M$ is
  non-empty, so let $(P_1, \bar d_1), \ldots, (P_m, \bar d_m)$ be an enumeration
  of the pointed instances in $S_M$, with $m >0$. Consider $(P_j, \bar d_j)$ for
  some $j\in[m]$. The definition of $S_M$ yields $( \prod \Nsf, \bar M) \to
  (P_j, \bar d_j)$. Composing $h_{q \to \Pi}$ with the witnessing homomorphism
  yields a homomorphism $h_{q \to P_j}$ from $q$ to $P_j$ such that $h_{q \to
  P_j}(\bar x) \subseteq \bar d_j$. Recall that $\rho$ is a fitting TGD and
  therefore $P_j\in\Psf$ implies $h_{q \to P_j} (\bar{x}) \in p(P_j)$. Hence,
  there exists a homomorphism $g_{p \to P_j}$ from $p(\bar x)$ to $(P_j,h_{q \to
  P_j} (\bar x))$.

  Let $(K, \bar c) = \prod S_M$ and define $g_{p \to K}: \mn{adom}(I_p) \to
  \mn{adom}(K)$ by setting $g_{p \to K}(y) = (g_{p \to P_1}(y), \ldots, g_{p \to
  P_m}(y))$. By construction $g_{p \to K}(\bar x) \subseteq \bar c$ and by
  definition  of products $g_{p \to K}$ is a homomorphism. Define a homomorphism
  $g_{p \to K^*}$ from $p(\bar x)$ to $(K^*, \bar c^*)$ by $g_{p \to K^*}(y)
  = c^*_i$ if $g_{p \to K}(y) = c_i$ and $g_{p \to K^*}(y) = g_{p \to K}(y)$
  otherwise. Clearly $g_{p \to K^*}(\bar x)\subseteq \bar c^*$ and since $g_{p
  \to K}$ is a homomorphism, the definition of diversification guarantees that
  $g_{p \to K^*}$ is also a homomorphism.

  By Condition~2 we have $(K^*, c^*) \to (N_i, \bar M[i])$ for some $i\in [k]$.
  Composing $g_{p \to K^*}$ with the witnessing homomorphism yields  a
  homomorphism $g_{p \to N_i}$ from $p$ to $N_i$ such that $g_{p \to N_i}(\bar x) \subseteq \bar M[i]$. We next establish that in fact $g_{p \to N_i}(\bar x)
  = \bar a_i$, thus showing $\bar a_i\in p(N_i)$ and thereby obtaining a
  contradiction, as desired. Let $x\in \bar x$ and let $\bar M = (M_1,\ldots,
  M_n)$. Observe that if $h_{q \to \Pi}(x) = M_\ell$ for $\ell\in[n]$, then
  $h_{q \to N_i}(x)$ is the $\ell$-th element of $\bar{M}[i]$ due to
  construction of $h_{q \to \Pi}$. Furthermore the construction of the
  homomorphisms $h_{q \to P_j}$, $g_{p \to P_j}$, $g_{p \to K}$, $g_{p \to K^*}$
  and $g_{p \to N_i}$ all preserve the order of the original mapping of $\bar x$
  to $\bar M$ induced by $h_{q \to \Pi}$. Thus $g_{p \to N_i}(x) = h_{q \to
  N_i}(x)$, which entails $g_{p \to N_i}(\bar x) = \bar a_i$, as required. 
\end{proof}

\thmmtcFTGDfittingTGD*
\noindent\begin{proof}\ ``$\Rightarrow$''. We show the contrapositive. Suppose
  there exist $R_1, \ldots, R_n$ and
  $\bar{a}_1, \ldots, \bar{a}_n$ with
  $\bar{a}_i\in \mn{adom}(\prod \Nsf)^{\mn{ar}(R_i)}$ and
  $R_i(\bar{a}_i[i])\notin N_i$ for $i\in[n]$, and such that for all $P\in \Psf$
  and $h$ witnessing $\prod \Nsf \to P$, we have $R_{1}(h(\bar{a}_1)), \ldots,
  R_{n}(h(\bar{a}_n))\in P$. We need to show that $(\Psf, \Nsf)$ has a fitting
  \FTGD.

  We will construct a \FTGD $\rho$ such that $N \not\models \rho$ for all $N\in\Nsf$ and $P\models
  \rho$ for all $P\in\Psf$. Define $\bar{a} = \bigcup_{i\in[n]}
  \bar{a}_i$ and
  $$
    \rho = \varphi(\bar{x},\bar y) \to R_1(\bar x_1) \wedge \cdots \wedge R_n(\bar x_n) \enspace \text{with}\enspace \bar{x} = \bigcup_{i\in[n]}
    \bar{x}_i,
  $$
  where $q_{\bar a}(\bar x)=\exists \bar{y} \,\varphi(\bar x,\bar y)$ is obtained
  from the canonical CQ of $(\prod\Nsf,\bar a)$ by renaming the free variables.
  Then $\bar{a}\in q_{\bar{a}}(\prod \Nsf)$, which yields $\bar{a}[i]\in
  q_{\bar{a}}(N_i)$ for all $i\in [n]$ by Lemma~\ref{lem:product_cq}. Since by
  assumption $R_i(\bar{a}_i[i])\notin N_i$, it follows that $N_i \not\models
  \varphi(\bar{x},\bar y) \to R_i(\bar x_i)$, which implies $N_i\not\models
  \rho$ for $i\in[n]$. It remains to show that $P\models \rho$ for every
  $P\in\Psf$. Let $P\in\Psf$. 
  
  \emph{Case 1.} Assume $\prod \Nsf \to P$,
  witnessed by $h$ with $h(\bar{a}) = \bar{b}$. This entails
  $R_{1}(h(\bar{a}_1)), \ldots, R_{n}(h(\bar{a}_n))\in P$ by assumption, hence
  $P\models\rho$. 
  
  \emph{Case 2.} No homomorphism from $\prod \Nsf$ to $P$
  exists. Then $P\not\models q_{\bar a}$, and therefore $P\models\rho$ holds
  vacuously. Thus, $\rho$ is a fitting \FTGD for $(\Psf, \Nsf)$.

  \smallskip\noindent ``$\Leftarrow$''. Assume that for all $R_1, \ldots, R_n$
  and $\bar{a}_1, \ldots, \bar{a}_n$ with 
  $\bar{a}_i\in \mn{adom}(\prod \Nsf)^{\mn{ar}(R_i)}$ and
  $R_i(\bar{a}_i[i])\notin N_i$ for $i\in[n]$, there
  exists a $P\in\Psf$ and a homomorphism $h$ from $\prod \Nsf$ to $P$, such that
  $R_j(h(\bar a_j)) \notin P$ for some $j \in [n]$. We show that $(\Psf, \Nsf)$
  has no fitting \FTGD. Towards a contradiction, suppose there is a fitting
  \FTGD
  $$
    \rho = \varphi(\bar x, \bar y) \to \psi(\bar x).
  $$ 
  For brevity, set $q(\bar{x}) = \exists \bar{y} \, \varphi$. Since
  $\rho$ is a fitting \FTGD, there is a homomorphism $h_i$ witnessing $(I_q,
  \bar x) \to (N_i, h_i(\bar x))$ and an atom $R_i(\bar x_i)$ in $\psi$ such
  that $R_i(h_i(\bar x_i))\notin N_i$ for every $i\in[n]$.
  
  From the atoms $R_i(\bar x_i)$ and homomorphisms $h_i$ we construct facts $\bar a_1,\dots,\bar a_n$ with $\bar a_j \in \mn{adom}(\prod \Nsf)^{\mn{ar}(R_j)}$
  and $R_j(\bar a_j[j]) \notin N_j$ for $j\in [n]$. Let
  $$
    \bar a_i = h_1(\bar x_i) \times \cdots \times h_n(\bar x_i)\quad \text{ for } i\in[n].
  $$
  
    Fix an $i\in[n]$. To see that indeed $\bar a_i\in \mn{adom}(\prod \Nsf)^{\mn{ar}(R_i)}$, let $\bar a_i = (\bar b_1,\ldots, \bar b_{|\bar a_i|})$ and let $k\in [|\bar a_i|]$. We will show that $\bar b_k\in \mn{adom}(\prod \Nsf)$. Observe that for $x$ in position $k$ of $\bar x_i$, $q(\bar x)$ must contain an atom $S(\bar z)$ with $\bar z\subseteq \bar x$, and $x \in \bar z$, because $\bar x_i \subseteq \bar x$. Then $h_j(\bar x)\in
    q(N_j)$ implies $S(h_j(\bar z)) \in N_j$ for $j\in[n]$. Thus, by definition of products,
    $\prod \Nsf$ contains the fact $S(h_1(\bar z) \times \cdots \times  h_n(\bar z))$, which has $(h_1(x), \ldots, h_n(x)) = \bar b_k$ as one of its values. Therefore $\bar b_k\in \mn{adom}(\prod N)$, as required.
       
  Since $\bar a_i[i] = h_i(\bar x_i)$, we immediately obtain
  $R_i(\bar{a}_i[i])\notin N_i$ for $i\in[n]$. Therefore, we may apply the assumption to $R_1,\ldots, R_n$ and $\bar a_1, \ldots, \bar a_n$,  which yields a $P\in\Psf$ and a homomorphism $h$ from $\prod
  \Nsf$ to $P$, such that $R_j(h(\bar a_j)) \notin P$ for some $j \in [n]$. Define $$\bar a = h_1(\bar x) \times \cdots \times h_n(\bar x).$$
  For every $i\in [n]$, $(I_q,
  \bar x) \to (N_i, h_i(\bar x))$ implies $h_i(\bar x)\in q(N_i)$. We immediately obtain $\bar a\in q(\prod \Nsf)$ by Lemma~\ref{lem:product_cq}.
  Then there exists a homomorphism $g$ witnessing $(I_q, \bar x) \to (\prod \Nsf, \bar a)$. Thus, $g \circ h$ is a witness to $(I_q, \bar x) \to (P, h(\bar a))$, and
  therefore $h(\bar a)\in q(P)$. By definition $h(\bar a_j) \subseteq h(\bar a)$, and since $R_j$ is an atom in $\psi$, $R_j(h(\bar a_j)) \notin P$ implies
  $P \not\models \rho$, the desired contradiction.

  The claimed bound on the number of atoms in the head of the fitting \FTGD by the number of examples in \Nsf follows directly from the constructed \FTGD in the ``$\Rightarrow$'' direction of the proof.
\end{proof}

\thmtgdfittingupper*

\noindent\begin{proof}\
  We start with $\Lmc \in \{ \GTGD, \FGTGD, \FONETGD\}$
  and
  concentrate on fitting TGD existence since Lemma~\ref{lem:oneneg} yields a simple
  reduction from fitting ontology existence to fitting TGD existence that gives the desired results. 

\smallskip

%
We start with fitting GTGD existence, using the characterization from Theorem~\ref{thm:mtc:GTGDfitting} 
to argue that fitting non-existence is in \NExpTime. 

To decide whether a given fitting instance $(\Psf,\Nsf)$ with $\Nsf = \{N_1,\dots,N_k\}$ admits
no fitting \GTGD, we have to check whether for every maximally guarded set $M \subseteq \mn{adom}(\prod
  \Nsf)$ such that $\bar{M}[i]$ is non-total in $N_i$ for all $i\in
  [k]$,  Conditions~1
and~2 of Theorem~\ref{thm:mtc:GTGDfitting} are  satisfied.

We first analyze the cardinality of the sets $S_M$ defined in Condition~1 and of the products $\prod S_M$ used in Condition~2.
By definition, the cardinality of the sets $S_M$ is bounded by $\sum_{P \in \Psf} |P|$. This is because $M$ is a maximally guarded set in $\prod
  \Nsf|_M$ and thus $(J,\bar b) \in S_M$ implies that $J$ contains a fact that
contains all values from $\bar b$ and no other values. Consequently, the
product $\prod S_M$ is of at most single exponential size. 

We also observe that each set $S_M$ can be computed in single exponential
time. To see this first note that each instance $\prod \Nsf\vert_M$ contains at
most linearly many values simply because $M$ is a maximally guarded set. 
We can thus decide in single exponential
time whether $(\prod \Nsf\vert_M,\bar{M}) \to (J,\bar{b})$ by a brute
force algorithm. 

Making use of the above observations and of the fact that the existence of
a homomorphism can be decided in \NPclass, it is easy to  derive a 
 \NExpTime algorithm. We iterate over all relevant sets $M \subseteq \mn{adom}(\prod \Nsf)$, of which there
 are only single exponentially many because these sets must be maximally guarded, and then check Conditions~1 and~2. Condition~1 can be checked
deterministically in single exponential time. For Condition~2, we can guess and
check the required homomorphism. We remark that,
despite the use of diversification, the structure
$(K^*,\bar c^*)$ is still of single exponential size.

\smallskip
The frontier-guarded case is very similar. A noteworthy difference is that
in Condition~1, we now need to decide the existence of a homomorphism from
$(\prod \Nsf,\bar M)$ to $(J,\bar b)$, rather than from $(\prod \Nsf|_M, \bar M)$.
However, the structure $\prod \Nsf$ has single exponentially many values rather
than linearly many, and thus a brute force approach no longer works. We solve
this problem as follows:
\begin{enumerate}

\item we observe that the characterization still holds if Condition~2
  is rephrased as follows:

  \begin{itemize}

     \item[2$'$.] $(K^*, \bar{c}^*) \to (N_i, \bar{M}[i])$ for some $i\in[k]$ and with  $(K,
           \bar{c}) = \prod S'_M$ for some non-empty $S'_M \subseteq S_M$.

  \end{itemize}
  In fact, Conditions~2 and~2$'$ are easily seen to be equivalent.

\item to verify Condition~1, guess $J\in\Psf$ and $\bar{b}\in\mn{adom}(J)^{|M|}$ and a homomorphism from $(\prod \Nsf,\bar{M})$ to $(J,\bar{b})$

\item to verify Condition~2, guess a set $S'_M$ of pairs $(J,\bar{b})$ with $J\in\Psf$ and $\bar{b}\in\mn{adom}(J)^{|M|}$, verify that $S'_M \subseteq S_M$
by guessing a homomorphism from  $(\prod \Nsf,\bar{M})$ to $(J,\bar{b})$ for each
$(J,\bar{b}) \in S'_M$, and finally guess an $i\in[k]$ and a homomorphism from $(K^*, \bar{c}^*) \to (N_i, \bar{M}[i])$ where $(K,
           \bar{c}) = \prod S'_M$. 

\end{enumerate}
The case of frontier-one TGDs is identical to that of frontier-guarded TGDs. The only difference is that the cardinality of the sets $S_M$ is bounded by $\sum_{P \in \Psf} |\mn{adom}(P)|$, since only singleton sets $M$ need to be considered. 

\smallskip

For unrestricted TGDs,  the algorithm exactly parallels the one for GTGDs, that is,
the modifications for frontier-guarded TGDs and frontier-one
TGDs are not needed. Let us verify that this algorithm 
yields a \coThreeNExpTime upper bound.
The sets $M \subseteq \mn{adom}(\prod \Nsf)$ that we iterate over in the outermost loop are no longer maximally guarded. There are therefore double exponentially many such sets. The cardinality
of $S_M$ is at most double exponential rather
than polynomial as in the case of GTGDs, the reason also
being that the sets $M$ are no longer maximally guarded.
The product $\prod S_M$ in Condition~2 is thus
of triple exponential size, which explains
the \coThreeNExpTime upper bound.

\medskip

For fitting ontology existence in place of fitting TGD 
existence, the same algorithm yields a \coTwoNExpTime
upper bound when combined with Lemma~\ref{lem:oneneg}, 
which allows us to reduce fitting ontology existence 
to fitting TGD existence with a single negative example. 
In that case,
there are only single exponentially many
sets $M \subseteq \mn{adom}(\prod \Nsf)$ to iterate
over in the outermost loop. Moreover, the cardinality of $S_M$ and the size
of  $\prod S_M$ drops by one exponential. As a consequence,
we can compute $(K^*,{\bar c}{}^*)$ in double exponential time
and guess the homomorphism in Condition~2.

\medskip

We now turn to fitting \FTGD existence and show that it is in \coNExpTime. We
use the characterization from Theorem~\ref{thm:mtc:FTGDfitting} for fitting
\FTGD non-existence.

The theorem states that no \FTGD fits $(\Psf, \Nsf)$ with $\Nsf = \{N_1,\ldots,
N_n\}$ if and only if for all choices of relation symbols $R_1, \dots, R_n$ and
tuples $\bar a_1, \dots, \bar a_n$ such that $\bar a_i \in \mn{adom}(\prod
\Nsf)^{\mn{ar}(R_i)}$ and $R_i(\bar a_i[i]) \notin N_i$ for all $i \in [n]$,
there exists a $P \in \Psf$ and a homomorphism $h$ from $\prod \Nsf$ to $P$ such
that $R_j(h(\bar a_j)) \notin P$ for some $j \in [n]$.

First, we analyze the size of the objects involved. The product instance $\prod
\Nsf$ can be of exponential size. The arity of the occurring relation symbols can
grow with the input, so for a relation symbol $R$, the number of possible tuples
$\bar a\in \mn{adom}(\prod \Nsf)^{\mn{ar}(R)}$ is exponential in the
worst case. Since $n = |\Nsf|$, the number of combinations of relation
symbols $R_1,\ldots, R_n$ and tuples $\bar a_1, \ldots, \bar a_n$ that need to be considered,
is bounded exponentially. This allows us to iterate over all combinations in single
exponential time. 

For each combination, we need to verify that there is a $P\in\Psf$ and a
homomorphism $h$ from $\prod \Nsf$ to $\Psf$, such that $R_i(h(\bar a_i))\notin P$
for some $R_i$ and $\bar a_i$ with $i\in[n]$. We can guess an $R_i$, a $P\in
\Psf$, and a homomorphism $h$ witnessing $\prod\Nsf \to P$. This puts us in \NExpTime 
because $\prod\Nsf$ is of exponential size and guessing a homomorphism is in \NPclass.
Verification of $R_i(h(\bar a_i))\notin P$ clearly is
possible in polynomial time. Exponentially many such guesses can be done in
\NExpTime, which is therefore the overall complexity. Since this analysis covers
the complement of \FTGD existence, the latter is decidable in \coNExpTime.

We show that fitting \FTGD-ontology existence is decidable in $\Sigma^p_2$. By
Lemma~\ref{lem:oneneg} it suffices to analyze fitting \FTGD existence for $(\Psf, \Nsf)$
with $|\Nsf| = 1$. We again use the negation of the characterization.
It states that an \FTGD fits $(\Psf, \Nsf)$ with $\Nsf = \{N\}$ if and only if
there is a relation symbol $R$ and a tuple $\bar a\in \mn{adom}(N)^{\mn{ar}(R)}$
with $R(\bar a) \notin N$, and such that for all $P\in\Psf$ and $\bar b\in
\mn{adom}(P)^{\mn{ar}(R)}$, if $(N, \bar a) \to (P, \bar b)$, then $R(\bar b)\in
P$. We can non-deterministically guess in polynomial time an $R$ and $\bar a\in\mn{adom}(N)^{\mn{ar}(R)}$ with $R(\bar a)\notin N$, and then use a
\coNPclass oracle to verify that all $P\in\Psf$ and $\bar b\in
\mn{adom}(P)^{\mn{ar}(R)}$ with $(N, \bar a) \to (P, \bar b)$ imply $R(\bar b)\in
P$. This puts fitting \FTGD-ontology existence in $\Sigma^p_2$.

If the arity of the relation symbols occurring in $(\Psf, \Nsf)$ is bounded by a
constant, fitting \FTGD-ontology existence can be decided in \coNPclass. To see
that, observe that the number of tuples $\bar a$ with $R(\bar a)\notin N$ is
polynomial in this case. This implies that fitting non-existence is in \NPclass,
as a polynomial-sized certificate can specify a witnessing homomorphism for each
of the polynomially many $(R, \bar a)$. Thus, fitting \FTGD-ontology existence
with bounded arity is in \coNPclass.
\end{proof}

\thmfittingextgdcomplexity*

\noindent\begin{proof}\
  We prove all \coNExpTime-hardness-results by reducing the \emph{product homomorphism problem (PHP)}. The product homomorphism problem takes as input finite instances $I_1,\ldots,
  I_n, J$ over the same schema and asks, whether $\prod_{i\in[n]} I_i \to J$.
  This problem is known to be \NExpTime-hard, even over a fixed schema consisting only of a binary relation symbol $E$~\cite{CateD15}.

\smallskip We start with Point~1.
  For $\Lmc \in \{\GTGD, \FGTGD, \FONETGD, \TGD\}$, we provide a polynomial time reduction from the product homomorphism problem
  to the complement of fitting \Lmc-ontology existence and to the complement of
  fitting \Lmc-TGD existence. We will use
  fitting instances with a single negative example, where by
  Lemma~\ref{lem:oneneg}, the existence of fitting ontologies and fitting TGDs
  coincides.
  
  Let $I_1,\ldots, I_n, J$ be an input to the product homomorphism problem. We
  assume w.l.o.g.\ that $J$ has a domain that is disjoint from those of
  $I_1,\ldots, I_n$.
  We construct a fitting instance $(\Psf, \Nsf)$ as follows. Take a fresh value
  $a$ and for every $b\in \mn{adom}(J)\cup\{a\}$ a fresh unary relation symbol
  $R_b$. Define instances $$J' = J \cup \{R_b(b)\mid b\in
  \mn{adom}(J)\cup\{a\}\}$$ and $$I'_i = I_i \cup J' \text{ for every }
  i\in[n].$$ Then set $\Psf = \{I'_i \mid i\in[n]\}$ and $\Nsf = \{J'\}$. We
  show that $\prod_{i\in[n]}I_i \to J$ if and only if $(\Psf, \Nsf)$ has no
  fitting \Lmc-TGD, based on the characterization given for fitting \Lmc-TGD existence. 
  
  First, let $\Lmc = \GTGD$. Then by Theorem~\ref{thm:mtc:GTGDfitting} $(\Psf, \Nsf)$ has no fitting \GTGD if and
  only if for every non-empty maximally guarded set $M\subseteq
  \mn{adom}(J')$ such that $\bar{M}$ is non-total in $J'$, the two conditions in
  Theorem~\ref{thm:mtc:GTGDfitting} are satisfied.
  
  Observe that $\bar{M}$ is non-total in $J'$ for every non-empty maximally
  guarded $M\subseteq \mn{adom}(J')$, as a result of the newly introduced
  relation symbols $R_b$ and the fresh value $a$.\footnote{The presence of $a$
  guarantees non-totality even if $\mn{adom}(J)$ is a singleton.} We now show
  that the first condition of Theorem~\ref{thm:mtc:GTGDfitting} is satisfied for
  every such $M$. Let $M\subseteq \mn{adom}(J')$ be non-empty and maximally
  guarded. We have to argue that 
  $$
    \begin{array}{r@{\;}c@{\;}l}
      S_M &=&
      \{(I',\bar{b}) \mid I'\in\Psf \text{ and } \bar{b}\in\mn{adom}(I')^{|M|} \text{ such that}\\[1mm]
      && \hspace*{12mm} (J'\vert_M,\bar{M}) \to (I',\bar{b})\}
    \end{array}
  $$
  is non-empty. In fact, we have that
  $$
    S_M = \{(I'_i, \bar{M}) \mid i\in[n]\}.
  $$
  The ``$\supseteq$'' direction holds since $J' \subseteq I'_i$ and the
  ``$\subseteq$'' direction is due to the use of the fresh relation symbols
  $R_b$.

  Therefore it suffices to show that $\prod_{i\in[n]}I_i \to J$ if and only if
  the second condition of Theorem~\ref{thm:mtc:GTGDfitting} is satisfied by
  $(\Psf, \Nsf)$, for every non-empty maximally guarded
  $M\subseteq \mn{adom}(J')$. That is:
  \\[2mm]
  {\bf Claim.} $\prod_{i\in[n]}I_i \to J$ if and only if $(K^*, \bar{c}^*) \to
  (J', \bar{M})$, where $(K, \bar{c}) = \prod S_M$, for every non-empty
  maximally guarded $M\subseteq \mn{adom}(J')$. \\[2mm]
    \noindent ``$\Rightarrow$''. Assume that there exists a homomorphism $h$ from
    $\prod_{i\in[n]}I_i$ to $J$ and let $M\subseteq \mn{adom}(J')$ be non-empty
    and maximally guarded and $\prod S_M = (K,\bar{c})$. 
    We have $(K^*, \bar{c}^*) \to (K, \bar{c})$. Therefore it suffices to show
    that $h$ can be extended to a homomorphism $h'$ that witnesses $(K, \bar{c})
    \to (J', \bar{M})$. To this end, let $h'$ be the extension of $h$ with
    $$
      h'((b_1,\dots,b_n))=b_i
    $$
    for every $(b_1,\dots,b_n) \in \mn{adom}(K) \setminus
    \mn{adom}(\prod_{i\in[n]}I_i)$, where $i \in [n]$ is smallest with $b_i \in
    \mn{adom}(J')$. Note that such an $i$ must exist, since otherwise
    $(b_1,\dots,b_n) \in \mn{adom}(\prod_{i\in[n]}I_i)$.
  
    It remains to show that $h'$ is a homomorphism. Thus let $R(\bar{d}_1,\ldots,
    \bar{d}_m) \in K$. If $\bar{d}_1,\ldots,
    \bar{d}_m \in \mn{adom}(\prod_{i\in[n]}I_i)$,
    then $R(h'(\bar{d}_1),\ldots,h'(\bar{d}_m)) = R(h(\bar{d}_1),\ldots,h(\bar{d}_m)) \in J \subseteq J'$. Otherwise,
    there is some $\bar{d}_i$ that contains a value from $\mn{adom}(J')$. Let $k$
    be smallest such that $\bar{d}_i$ contains such a value in its $k$-th component. Now let $d_1^k,\dots,d_m^k$ be the $k$-th components of 
    $\bar{d}_1,\dots,\bar{d}_m$. By definition of products, $R(\bar{d}_1,\ldots,
    \bar{d}_m) \in K$ implies that $R(d_1^k,\dots,d_m^k) \in I'_i$.
    But $I'_i$ is the disjoint union of $I_i$ and $J'$, and thus $d^k_i
    \in \mn{adom}(J')$ implies that $d_1^k,\dots,d_m^k \in \mn{adom}(J')$
    and thus we have $R(d_1^k,\dots,d_m^k) \in J'$.
    It remains to observe that, by definition of $h'$,  
    $R(h'(\bar{d}_1),\ldots,h'(\bar{d}_m)) = R(d_1^k,\dots,d_m^k)$.
  
    \smallskip
    
    \noindent ``$\Leftarrow$''. Assume that $(K^*, \bar{c}^*) \to (J', \bar{M})$, where
    $(K, \bar{c}) = \prod S_M$, for every non-empty maximally guarded
    $M\subseteq \mn{adom}(J')$. Take any such $M$ and a homomorphism $h$ that
    witnesses $(K^*, \bar{c}^*) \to (I, \bar{M})$.
    Let $h'$ denote the restriction of $h$ to $\mn{adom}(\prod_{i\in[n]}I_i)$.
    Then clearly $h'$ is a homomorphism from $\prod_{i\in[n]}I_i$ to $J'$. Since
    every value from $\prod_{i\in[n]}I_i$ occurs in some fact using a relation
    symbol different from $R_a$, $h'$ cannot map any value from
    $\prod_{i\in[n]}I_i$ to $a$, the fresh value. Thus $h'$ is also a
    homomorphism from $\prod_{i\in[n]}I_i$ to $J$, as required. This concludes
    the proof for \GTGD.
    
    \medskip

    For $\Lmc \in \{\FGTGD, \FONETGD, \TGD\}$, we follow the structure of the proof for \GTGD and establish analogous results.
    Instead of Theorem~\ref{thm:mtc:GTGDfitting}, we invoke the following:
    \begin{itemize}
        \item Theorem~\ref{thm:mtc:FGTGDfitting}, when $\Lmc = \FGTGD$;
        \item Theorem~\ref{thm:mtc:FONETGDfitting}, using values from $\mn{adom}(J')$ rather than maximally guarded sets, when $\Lmc = \FONETGD$;
        \item Theorem~\ref{thm:mtc:TGDfitting}, using unrestricted sets rather than maximally guarded sets, when $\Lmc = \TGD$.
    \end{itemize}

    With respect to the non-totality of $\bar M$, for \FGTGD we also consider maximally guarded sets, as for \GTGD. For \FONETGD and \TGD, note that for any non-empty $M\subseteq \mn{adom}(J')$, whether $M$ is a singleton or unrestricted, $\bar M$ is non-total in $J'$, following the same argument as in the proof for \GTGD. The non-totality of values is implied by the non-totality of $\bar M$, where $M$ is a singleton.
    
    Regarding Condition~1 of the respective characterizations for fitting \Lmc-TGD, consider the sets defined in Condition~1 for each $\Lmc$. Note that for \FGTGD, even though maximally guarded sets are used, as in the case of \GTGD, the definition of the sets $S_M$ uses homomorphisms from $(J',\bar{M})$, as opposed to $(J'\vert_M,\bar{M})$. However, this has no influence on the argument. Thus, from here on, we can follow the exact same proof as for \GTGD, since in both cases maximally guarded sets are considered. For $\Lmc\in \{\FONETGD, \GTGD\}$, by the same reasoning as in the proof for \GTGD, we observe:
    \begin{itemize}
        \item $S_M = \{(I'_i, \bar{M}) \mid i\in[n]\}$ holds for unrestricted $M$,
        \item $S_a = \{(I'_i, a) \mid i\in[n]\}$ holds for values $a$.
    \end{itemize}
    This confirms the non-emptiness of the sets defined in Condition~1 of each respective characterization.
    
    Thus, for each \Lmc, the reduction depends solely on Condition~2 of the theorem characterizing fitting \Lmc-TGD existence. This allows us to state a claim, analogous to the one in the proof for \GTGD: $\prod_{i\in[n]}I_i \to J$ if and only if
    \begin{itemize}
        \item $(K^*, \bar{c}^*) \to (J', \bar{M})$, where $(K, \bar{c}) = \prod S_M$, for unrestricted $M$, when $\Lmc = \TGD$,
        \item $\prod S_a
        \to (J', a)$, for $a\in\mn{adom}(J')$, when $\Lmc = \FONETGD$.
    \end{itemize}
    Observe that the proof of the claim for \GTGD does not fundamentally depend on $M$ being maximally guarded. Thus, the respective claim for \TGD can be shown in the exact same way, assuming $M$ is unrestricted. For $\Lmc = \FONETGD$, values take the place of maximally guarded sets. This does not alter the overall argument but simplifies the proof slightly, as $\prod S_a$ is considered directly, rather than its diversification. This finishes the proof of Point~1 of the theorem.

    \medskip We now turn to the proof of Point~2 of Theorem~\ref{thm:fitting-ex-tgd-complexity}, starting with the \coNExpTime-hardness. As announced, we reduce the PHP.
    Recall that the PHP is already \NExpTime-hard over instances over a single binary relation symbol~$E$~\cite{CateD15}, so we assume such instances from now on. Let $I_1,\ldots,I_n,J$ be an input to the PHP. We assume without loss of generality that $I_1,\ldots,I_n,J$ have pairwise disjoint active domains and that $c$ is a fresh value (that does not occur in any of the domains). We construct a fitting instance $(\Psf,\Nsf)$ as follows:
    \begin{itemize}
        \item $\Psf=\{P_1,P_2\}$ where $P_1=J\cup \{B(c)\}$ and $P_2=I_1\cup\{A(c),B(c)\}$;
        \item $\Nsf=\{I_1',\ldots,I_n'\}$ where $I_i'$ is obtained from $I_i$ by adding $A(a)$ for every $a\in\mn{adom}(I_i)$ and $B(c)$.
    \end{itemize}
    The following claim establishes correctness of the reduction.

    \medskip\noindent\textbf{Claim.} $\prod_i I_i\to J$ iff $(\Psf,\Nsf)$ has no fitting \FTGD.

    \medskip \noindent\textbf{Proof of the Claim.} For ``$\Rightarrow$'' suppose $\prod_i I_i\to J$ and let $h$ be a witness for this. Let $R_1,\ldots,R_n$ be relation symbols and $\bar a_1,\ldots,\bar a_n$ be tuples such that $\bar a_i\in\mn{adom}(\prod \Nsf)^{\mn{ar}(R_i)}$ and $R_i(\bar a_i[i])\notin I_i'$ for $i\in[n]$. We have to show that there is a $P\in\Psf$ and a homomorphism $h$ from $\prod\Nsf$ to $P$ such that $R_j(h(\bar a_j))\notin P$ for some $j\in [n]$. We distinguish cases on the relations symbol $R_1$:
    \begin{itemize}
    
        \item If $R_1=A$, then $\bar a_i[1]=c$ for all $i\in[n]$ and $h$ extended with $h(c,\ldots,c)=c$ is the required homomorphism to $P_1$;
        
        \item If $R_1=B$, then $a_i[1]\neq c$ for all $i\in[n]$ and the map $g:\mn{adom}(\prod \Nsf)\to \mn{adom}(P_2)$ defined by 
        \[g(b_1,\ldots,b_n)=\begin{cases} b_1 & \text{if $c\notin \{b_1,\ldots,b_n\}$} \\ c & \text{otherwise}\end{cases}\]
        is the required homomorphism to $P_2$;
        
        \item If $R_1=E$, then the map $g$ from the previous point
        is the required homomorphism to $P_2$;
          
    \end{itemize}
    For the other direction ``$\Leftarrow$'', suppose for every relation symbol $R$ and tuple $\bar a\in\mn{adom}(\prod \Nsf)^{\mn{ar}(R)}$ with $R(\bar a[i])\notin I_i'$ for all $i\in [n]$, there is a $P\in\Psf$ and a homomorphism $h$ from $\prod\Nsf$ to $P$ such that $R(h(\bar a))\notin P$ for some $j\in [n]$. Consider relation symbol $A$ and tuple $\bar a=c,\ldots,c$. Since $B(\bar a)\in\prod\Nsf$,
    the only potential target of $\bar a$ by such homomorphism $h$ is the point $c$ in $P_1$. But then $h$ has to map $\prod_i I_i$ into $J$, hence $\prod_i I_i\to J$ as required.\hfill$\dashv$ 

    \medskip
\medskip
It remains to prove the DP-hardness result. We do this
by reduction from 3-colorability/non-homomorphism,
that is, we are given a triple 
$(G,I,J)$ with $G=(V,E)$ an undirected graph and $I$ and $J$ instances over a single binary relation $E$, and we want to decide whether $G$ is 3-colorable
and $I \not\rightarrow J$. It is easy 
to prove that this problem is DP-complete.

Let $(G,I,J)$ be a triple of the
described form. 
Further let $G=(V,E)$ and $V=\{v_1,\dots,v_n\}$. We use the following relation symbols:
\begin{itemize}

    \item unary relation symbols $V_1,\dots,V_n$ and $C_1,\dots,C_n$ (for identifying the vertices of $G$ and colors for each vertex in $G$);

    \item a relation symbol $W$ of arity~4
    (for \emph{well-colored}), representing possible colorings of pairs of nodes in $G$;

    \item a relation symbol $R$ of arity~$2n$ for choosing a 3-coloring of $G$;

    \item a binary relation symbol $E$ for edges of $I$.
    
\end{itemize}
 We introduce the following examples:
\begin{itemize}

\item one negative example $N$ that contains the following facts:
\begin{itemize}

    \item $V_i(v_i)$ for $1 \leq i \leq n$;

    \item $C_i(r_i), C_i(g_i), C_i(b_i)$ for $1 \leq i \leq n$;

    \item for $1 \leq i < j \leq n$ and all
    $c,d \in \{r,g,b\}$, the fact
    $W(v_i,c_i,v_j,d_j)$ if $\{v_i,v_j\} \notin E$ or $c \neq d$;

   \item all facts from the instance $I$.
    
\end{itemize}

\item a positive example $P_0$ which ensures that if we choose a non-$R$-tuple $\bar a$ in $N$ that is of the wrong `type', then we find a homomorphism
from $N$ to $P_0$ such that $R(h(\bar a)) \notin P_0$. It contains the following
facts:
\begin{itemize}

    \item $V_1(v_1),\dots,V_n(v_n)$ and  $C_1(c_1),\dots,C_n(c_n)$;

    \item all facts $W(v_i,c_i,v_j,c_j)$ with $1 \leq i < j \leq n$;
 
     \item $R(v_1,c_1,\dots,v_n,c_n)$;

     \item $E(\bot,\bot)$.
   
\end{itemize}

\item a positive example $P_1$ 
that contains the
following facts:
\begin{itemize}

    \item $V_j(u_i)$ and $C_j(u_i)$ for $1 \leq j \leq n$ and all $i \in \{1,2\}$;

    \item $C_j(\bot_i)$ for $1 \leq j \leq n$ and $i \in \{1,2\}$;

    \item  $W(a_1,b_1,a_2,b_2)$ for all $a_1,a_2 \in \{u_1,u_2\}$ and $b_1,b_2 \in D$, where $D=\{u_1,u_2,\bot_1, \bot_2 \}$, such that
    \begin{enumerate}
        \item[(i)] $a_1=a_2$  or
        $(a_1,a_2)=(u_1,u_2)$;
        \item[(ii)] $a_i=b_i$ or $(a_i,b_i)=(u_1,\bot_1)$ or 
        $(a_i,b_i)=(u_2,\bot_2)$ for all $i \in \{1,2\}$;
        \item[(iii)] $(a_1,b_1,a_2,b_2) \neq (u_1,u_1,u_2,u_2)$.
    \end{enumerate}

      \item $R(\bar a)$ for all
    $\bar a = (a_1,b_1,\dots,a_n,b_n) \in D^{2n}$ such that
    \begin{enumerate}

        \item[(iv)] $a_i=a_{i+1}$  or
        $(a_i,a_{i+1})=(u_1,u_2)$ for $1 \leq i < n$;

        \item [(v)] $a_i=b_i$ or $(a_i,b_i)=(u_1,\bot_1)$ or
        $(a_i,b_i)=(u_2,\bot_2)$;

        \item[(vi)] $\{(u_1,u_1),(u_2,u_2)\} \not\subseteq \{(a_1,b_1),\dots,(a_n,b_n)\}$;

    \end{enumerate}

     \item $E(\bot,\bot)$.

\end{itemize}

  \item a positive example
  $P_2$ that contains the following facts:
  \begin{itemize}

      \item all facts on the relation symbols $V_1,\dots,V_n,C_1,\dots,C_n,W$ that use the single value $\bot$;

      \item all facts from the instance $J$.
      
  \end{itemize}

  \item a positive example $P_3$ that contains the following facts:
  \begin{itemize}

\item all facts on the relation symbols $V_1,\dots,V_n,C_1,\dots,C_n,W,R$ that use the single value $\bot$;

      \item all facts from $I$.
      
  \end{itemize}

\end{itemize}
It is not hard to verify that the size of
our examples is only polynomial in the size of $G$. By Theorem~\ref{thm:mtc:FTGDfitting},
it suffices to show the following.
\\[2mm]
{\bf Claim 1.} $G$ is 3-colorable and $I \not\rightarrow J$ if and only if there is a relation symbol $X$ of some
arity $k$ and an
$\bar a \in \mn{adom}(N)^{k}$ with $X(\bar a) \notin N$ such that for all homomorphisms 
$h$ from $N$ to a positive example $P$, we have $X(h(\bar a)) \in P$.
\\[2mm]
We first argue that the only interesting choice for the relation symbol $X$ is $R$, because for all $X \in \{V_1,\dots,V_n,C_1,\dots,C_n,W,E \}$
and all
$\bar a \in \mn{adom}(N)^{k}$ with $X(\bar a) \notin N$, there is a homomorphism $h$ from $N$ to a positive example $P$ with $X(h(\bar a)) \notin P$. Whenever
$X \in \{V_1,\dots,V_n,C_1,\dots,C_n\}$,
we may in fact choose $P=P_0$
and the homomorphism $h$ with
\begin{itemize}

    \item $h(v_i)=v_i$ for $1 \leq i \leq n$;

    \item $h(r_i)=h(g_i)=h(b_i)=c_i$ for $1 \leq i \leq n$;

    \item $h(a)=\bot$ for all
    $a \in \mn{adom}(I)$.
    
\end{itemize}
If $X=W$ and $\bar a$ is not of the form $(v_k,c,v_\ell,d)$ with
$c \in \{r_k,g_k,b_k\}$ and
$d \in \{r_\ell,g_\ell,b_\ell\}$, then 
we may again use $P=P_0$ and the
homomorphism $h$ just given.
Thus the remaining case is that $X=W$ and $\bar a$ is  of the form $(v_k,c,v_\ell,d)$ with $c \in \{r_k,g_k,b_k\}$ and
$d \in \{r_\ell,g_\ell,b_\ell\}$. We then
choose $P=P_1$ and the following
homomorphism $h$:
\begin{itemize}
    \item $h(v_i)= u_1$ for $1 \leq i \leq \ell$;
       \item $h(r_i)=h(g_i)=h(b_i) = u_1$ for $1 \leq i \leq k$;
    \item $h(c_k)=u_1$ and $h(d)=\bot_1$ for all $d \in \{r_k,g_k,b_k \} \setminus \{ c_k \}$;
        \item $h(r_i)=h(g_i)=h(b_i) = u_1$ for $k < i \leq \ell$;
    \item $h(v_i)= u_2$ for $\ell \leq i \leq n$;
        \item $h(c_\ell)=\bot_2$ and $h(d)=u_2$ for all $d \in \{r_\ell,g_\ell,b_\ell \} \setminus \{ c_\ell \}$;
    \item $h(r_i)=h(g_i)=h(b_i) =u_2$ for $\ell < i \leq n$;

        \item $h(a)=\bot$ for all
    $a \in \mn{adom}(I)$.

\end{itemize}
Finally, if $X=E$, then we may choose $P=P_3$ and the following homomorphism:
\begin{itemize}
   
    \item $h$ maps all values $v_i,r_i,g_i,b_i$ to $\bot$;

    \item $h$ is the identity on all values from $I$.
    
\end{itemize}
By what was said above, to show correctness of the reduction it suffices to prove the following claim.
\\[2mm]
{\bf Claim 2.} $G$ is 3-colorable and $I \not\rightarrow J$ if and only if
there is an 
$\bar a \in \mn{adom}(N)^{2n}$ with $R(\bar a) \notin N$ such that for all homomorphisms 
$h$ from $N$ to a positive example $P$, we have $R(h(\bar a)) \in P$.
\\[2mm]
\smallskip\noindent\textbf{Proof of the claim.}
``$\Rightarrow$''. Assume that $G$ is 3-colorable with $\chi:V \rightarrow \{r,g,b\}$  a proper 3-coloring, and that $I \not\rightarrow J$. We  show that
Point~1 of Claim~2 is satisfied.
Set $\bar a = (v_1,\chi(v_1)_1,\dots,v_n,\chi(v_n)_n)$. We have $R(\bar a)\notin N$, simply because $N$ contains no $R$-tuples. Let $h$ be a homomorphism into a positive example $P$. We have to show that  $R(h(\bar a))\in P$. We distinguish cases:
\begin{itemize}

    \item $P=P_0$. Then because of the use of the $V_i$ and $C_i$ relations,
    we must have $h(\bar a)=(v_1,c_1,\dots,v_n,c_n)$. But then
    $R(h(\bar a))\in P_0$, as required.
    
    \item $P=P_1$. Let $h(\bar a)=(a_1,b_1,\dots,a_n,b_n)$. Because $\bar a$ is derived from a 
    proper 3-coloring and by definition 
    of $N$, we have $W(v_i,\chi(v_i)_i,v_j,\chi(v_j)_j) \in N$ for $1 \leq i < j \leq n$.
        Due to the definition of the $W$-relation in $P_1$, this implies that Properties~(iv)
    and~(v) from the definition of $P_1$
    are satisfied. It also implies
    that Property~(vi) is satisfied 
    since there is no fact $W(u_1,u_1,u_2,u_2) \in P_1$.
 But then 
    $R(h(\bar a))\in P_1$, as required

    \item $P=P_2$. Then $h$ witnesses that $I \rightarrow J$, in 
    contradiction to  $I \not\rightarrow J$.

\end{itemize}

\smallskip
\noindent
``$\Leftarrow$''. First assume that there is an 
$\bar a \in \mn{adom}(N)^{2n}$ 
with $R(\bar a) \notin N$ such that for all homomorphisms 
$h$ from $N$ to a positive example $P$, we have $R(h(\bar a)) \in P$. 

Considering the positive example $P=P_0$,
this means that $\bar a$ takes the form
$(v_1,c_1,\dots,v_n,c_n)$ where $c_i \in \{r_i,g_i,b_i\}$ for $1 \leq i \leq n$.
In fact, consider the homomorphism $h$
from $N$ to $P_0$ defined as follows:
\begin{itemize}
    \item $h(v_i)=v_i$ for $1 \leq i \leq n$;
    \item $h(r_i)=h(g_i)=h(b_i)=c_i$ for $1 \leq i \leq n$.
\end{itemize}
If $\bar a$ does not take the stated
form, then $R(h(\bar a)) \notin P_0$,
in contradiction to our choice of $\bar a$.

Note that $\bar a$ thus represents a 
coloring $\chi$ of $G$ defined by setting
$\chi(v_i)=r$ if $c_i=r_i$, $\chi(v_i)=g$ if $c_i=g_i$, and $\chi(v_i)=b$ if $c_i=b_i$.
Considering the positive example $P=P_1$, we now argue that $\chi$ is proper.

Assume to the contrary that $\chi$ is not proper. Then by definition of $N$ there 
are $k,\ell$ with $1 \leq k < \ell \leq n$ such that $W(v_k,c_k,v_\ell,c_\ell) \notin N$. As a consequence, the following map
$h$ is a homomorphism from $N$ to $P_1$:
\begin{itemize}
    \item $h(v_i)= u_1$ for $1 \leq i \leq \ell$;
       \item $h(r_i)=h(g_i)=h(b_i) = u_1$ for $1 \leq i \leq k$;
    \item $h(c_k)=u_1$ and $h(d)=\bot_1$ for all $d \in \{r_k,g_k,b_k \} \setminus \{ c_k \}$;
        \item $h(r_i)=h(g_i)=h(b_i) = u_1$ for $k < i \leq \ell$;
    \item $h(v_i)= u_2$ for $\ell \leq i \leq n$;
        \item $h(c_\ell)=\bot_2$ and $h(d)=u_2$ for all $d \in \{r_\ell,g_\ell,b_\ell \} \setminus \{ c_\ell \}$;
    \item $h(r_i)=h(g_i)=h(b_i) =u_2$ for $\ell < i \leq n$;
        \item $h(a)=\bot$ for all
    $a \in \mn{adom}(I)$.

\end{itemize}
But $R(h(\bar a)) \notin P_1$, contradicting the choice of $\bar a$.

Assume to the contrary of what is left to be shown that there is a homomorphism
$h$ from $I$ to $J$. Then there is clearly
also a homomorphism $h$ from $N$ to $P_2$.
So we have $R(h(\bar a)) \in P_2$. This, however, contradicts the fact that $P_2$
does not contain any $R$-facts.
\end{proof}

We recall a known result that we are going to use as a black box. 
\begin{theorem}\label{thm:cqfitting-succinctness}\cite{CateD15,cp/Willard10}
  Let $n\geq 1$. There are instances $I_1,\ldots,I_n,J$ such that
  \begin{enumerate}
  
    \item the size of the instances is bounded by $p(n)$, $p$ a polynomial;
    
    \item $\prod_i I_i \not\to J$, that is, there is a Boolean CQ $q$ with
    $\prod_i I_i\models q$ and $J \not \models q$;

    \item the smallest such CQ $q$ 
    has size $2^n$.
    
  \end{enumerate}
\end{theorem}

We can now show the claimed lower bounds based on Theorem~\ref{thm:cqfitting-succinctness} and a construction similar to the one in the complexity hardness proof in Theorem~\ref{thm:mtc:GTGDfitting}. 

\thmTGDfitsizelower*

\noindent\begin{proof}\
Let $n\geq 1$ and $I_1,\ldots,I_n,J$ be the instances that exist by Theorem~\ref{thm:cqfitting-succinctness}, and let $q_0$ be the Boolean CQ witnessing Point~2. We assume without loss of generality that the active domains of all these instances are pairwise disjoint. We construct $\Psf_n,\Nsf_n$ by taking:
\begin{itemize}
    \item $\Nsf_n = \{J\}$;
    \item $\Psf_n = \{I_1',\ldots,I_n'\}$ where $I_i'=I_i\cup J$, for all $i$.
\end{itemize}
Clearly, Point~1 is satisfied. For Point~2, observe that $\top\to q_0$ witnesses that $(\Psf_n,\Nsf_n)$ admit a fitting guarded and frontier-one TGD. For Point~3, suppose $\rho = p(\bar x)\to q(\bar x)$ is a smallest TGD fitting $(\Psf_n,\Nsf_n)$. We make a couple of observations.

Observe first that we can assume that $q(\bar x)$ is connected. Here, we call $q(\bar x)$ connected if the undirected graph $(\mn{var}(q),\{ \{x,y\}\mid x,y \text{ co-occur in some atom in }q\})$ is connected. In fact, since $J$ is a negative example for $\rho$, there is a tuple $\bar c$ such that there is a homomorphism $h$ from $p(\bar x)$ to $(J,\bar c)$ but not a homomorphism from $q(\bar x)$ to $(J,\bar c)$. Hence, we can select a connected component $q'(\bar x)$ of $q(\bar x)$ that does not have a homomorphism to $(J,\bar c)$. 

Suppose now that $\bar x$ is not the empty tuple. Since $J\subseteq I_i$ for all $i$, $h$ from the previous paragraph
is also a homomorphism from $p(\bar x)$ to all $(I'_i,\bar c)$. Since all $I_i'$ are positive examples for $\rho$, there is also a homomorphism from $q(\bar x)$ to $(I'_i,\bar c)$. As $q(\bar x)$ is connected, this is actually a homomorphism from $q(\bar x)$ to $(J,\bar c)$, a contradiction. We conclude that~$\bar x$ is the empty tuple and $q$ is in fact a Boolean CQ.

It remains to note that $q$ fits $I_1',\ldots,I_n'$ and $J$ that is, $I_i'\models q$, but $J\not\models q$. Since $q$ is connected and $J$ is contained in each $I_i'$, we conclude that $I_i\models q$ for all~$i$. By Point~3 of Theorem~\ref{thm:cqfitting-succinctness}, $q$ and thus $\rho$ is of size $2^n$.

To see that the same also holds for ontologies, recall that by Lemma~\ref{lem:oneneg} a smallest ontology fitting $(\Psf_n,\Nsf_n)$ contains only one TGD since there is only one negative example involved. 
\end{proof}

\thmfulltgdsize*

\noindent\begin{proof}\ 
  Let $n\geq 1$ and $I_1,\ldots,I_n,J$ be the instances that exist by Theorem~\ref{thm:cqfitting-succinctness}. We assume without loss of generality that the active domains of all these instances are pairwise disjoint. Let $A$ be a fresh unary relation symbol and let $I_i^A$ denote the extension of $I_i$ with all facts $A(d)$ for $d\in\text{adom}(I_i)$;

We construct a fitting instance $(\Psf_n,\Nsf_n)$ by taking: 
\begin{itemize}

    \item $\Psf_n= \{ J' \}$ for 
    $J' =J\cup\bigcup_i I_i^A$;

    \item $\Nsf_n=\{I_1,\ldots,I_n\}$.
    
\end{itemize}
Clearly, Point~1 of the theorem is satisfied. For Point~2, let $q$ be any Boolean CQ with $\prod_i I_i\models q$, but $J\not\models q$, which exists due to Point~2 of Theorem~\ref{thm:cqfitting-succinctness}. Let $q'(x)$ be the variant of $q$ in which one (any) of the quantified variables of $q$ is made an answer variable. Then $q'(x)\to A(x)$ is a fitting full TGD for $(\Psf_n,\Nsf_n)$. 

For Point~3, let $\rho=\varphi(\bar x,\bar y)\to \psi(\bar y)$ be a fitting full TGD. Since all $I_i$ are negative examples, there are, for $i\leq n$, homomorphisms $h_i$ from $q(\bar y) = \exists \bar{x} \, \varphi(\bar x,\bar y)$ to $I_i$ such that $I_i\not\models \psi(h_i(\bar y))$. Suppose first that $\psi$ does not mention $A$. But then such $h_i$ is also a homomorphism from $q(\bar y)$ to $I_i^A$ and by construction $I_i^A\not\models \psi(h_i(\bar y))$. Hence, $J'$ is not a positive example for $\rho$, a contradiction. Thus, $\psi(\bar y)$ contains an atom $A(z)$. Since none of the $I_i\in \Nsf$ contains an $A$-fact, we can actually assume that $\psi(\bar y)$ is a single fact $A(z)$, and that $\rho$ takes the shape $q(z)\to A(z)$. Now, this $q(z)$ cannot have a homomorphism to $J$ since otherwise $J'$ would not be a positive example for $\rho$. Consider the Boolean CQ $q'=\exists z\, q(z)$. By what was said above $q'$ has a homomorphism into every $I_i$, but not into $J$. By Point~3 of Theorem~\ref{thm:cqfitting-succinctness}, $q'$ is of size at least $2^n$.
\end{proof}

\end{document}